\documentclass{article}

\usepackage{amsmath}
\usepackage{amsthm}
\usepackage{amssymb}
\usepackage{multirow}
\usepackage{makecell}
\usepackage{pbox}
\usepackage{graphicx}
\usepackage[]{mdframed}
\usepackage{float}
\usepackage[dvipsnames]{xcolor}
\usepackage[numbers,sort&compress]{natbib}

\newtheorem{theorem}{Theorem}

\usepackage[preprint]{neurips_2024}

\usepackage[utf8]{inputenc} % allow utf-8 input
\usepackage[T1]{fontenc}    % use 8-bit T1 fonts
\usepackage{hyperref}       % hyperlinks
\usepackage{url}            % simple URL typesetting
\usepackage{booktabs}       % professional-quality tables
\usepackage{amsfonts}       % blackboard math symbols
\usepackage{nicefrac}       % compact symbols for 1/2, etc.
\usepackage{microtype}      % microtypography
\usepackage{xcolor}         % colors

\title{Transformer Normalisation Layers and the Independence of Semantic Subspaces}
\author{%
  Stephen Menary \\
  University of Manchester, UK \\
  \texttt{stephen.menary@manchester.ac.uk} \\
   \And
  Samuel Kaski \\
  $^{1}$ University of Manchester, UK \\
  $^{2}$ Aalto University, Finland \\
  \texttt{samuel.kaski@aalto.fi} \\
   \And
  Andr{\'e} Freitas \\
    $^{1}$ Department of Computer Science, University of Manchester, UK \\
    $^{2}$ Idiap Research Institute, Switzerland \\
    $^{3}$ National Biomarker Centre, CRUK-MI, University of Manchester, UK\\
  \texttt{andre.freitas@manchester.ac.uk} \\
}

\begin{document}

%%%
%%%  Title
%%%
\maketitle

%%%
%%%  Abstract
%%%

\begin{abstract}
Recent works have shown that transformers can solve contextual reasoning tasks by internally executing computational graphs called \textit{circuits}. Circuits often use attention to logically match information from subspaces of the representation, e.g. using position-in-sequence to identify the previous token. In this work, we consider a \textit{semantic subspace} to be any independent subspace of the latent representation that can fully determine an attention distribution. We show that \texttt{Pre-Norm}, the placement of normalisation layer used by state-of-the-art transformers, violates this ability unless the model learns a strict representation structure of orthogonal spheres. This is because it causes linear subspaces to interfere through their common normalisation factor. Theoretically, we analyse circuit stability by modelling this interference as random noise on the $L_2$-norms of the query/key/value vectors, predicting a phenomenon of \textit{circuit collapse} when sparse-attention shifts to a different token. Empirically, we investigate the sensitivity of real-world models trained for mathematical addition, observing a 1\% rate of circuit collapse when the norms are artificially perturbed by $\lesssim$10\%. We contrast \texttt{Pre-Norm} with \texttt{QKV-Norm}, which places normalisation \textit{after} the attention head's linear operators. Theoretically this relaxes the representational constraints. Empirically we observe comparable in-distribution but worse out-of-distribution performance. 
\end{abstract}

%%%
%%%  Section: Introduction
%%%
\section{Introduction}
\label{sec:intro}

%- Very difficult to study structure empirically, strength of theoretical approach is that we can also study principles axiomatically
%- Add theorems: double norm is ineffective; heads can perform binary match\&pass operations (attention polysemanticity); heads can perform different operations on sparsely separated features; heads can fallback to doing nothing, creating a category of ``other'' without structural constraints
%In contrast to other works, we do not seek human-interpretable explanations of circuits. Instead we identify the key structural components and seek to understand how these must behave. This allows us to decouple the understanding of logival struicture from human interpretability, which is a core difficulty of prior works

Transformer-based models \cite{DBLP:journals/corr/VaswaniSPUJGKP17} are commonplace in machine learning, providing state-of-the-art contextual reasoning in domains ranging from natural language \cite{bubeck2023sparks,touvron2023llama} to protein-folding \cite{AlphaFold3,doi:10.1126/science.abj8754,10.7554/eLife.82819} and theoretical physics \cite{cai2024transforming}. Recent interpretability work investigates the internal mechanisms that lead to specific model behaviours \cite{olsson2022context,elhage2021mathematical,wang2022interpretability,ferrando2024primer,meng2022locating,DBLP:journals/corr/abs-2002-12327,liu2023transformers,goldowskydill2023localizing}. This is important for predicting behaviour in new environments, enables practitioners to match the inductive bias of a model with the structure of its task, and informs the design of architectures that promote desirable behaviour.

Two such works discovered complete \textit{circuits} \cite{cammarata2020thread:} in trained transformers \cite{olsson2022context,elhage2021mathematical,wang2022interpretability}. These are computational graphs that dominate the model prediction when activated in a specialised context. They perform a type of \textit{algorithmic reasoning} by internally executing a sequence of logical operations, using attention to pass information between memory buffers that begin as token embeddings and become increasingly abstract. Furthermore, a number of attention heads have been identified as performing logical operations (see \cite{ferrando2024primer} section 5). To understand transformer behaviour, an important goal is to understand how logical attention heads operate, and their generality beyond the simple cases that facilitate interpretability.

One key observation is that \textbf{the attention distribution is sometimes fully-determined by an independent subspace of the representation} - for example, an attention layer can identify the previous token by accessing a subspace that encodes only position-in-sequence. Indeed, low-rank weight matrices can only access linear subspaces by construction. A second observation is that, like most deep architectures, transformers use normalisation layers to improve training stability. A leading choice is to place normalisation at the \textit{input} to each attention layer, which we call \texttt{Pre-Norm} \cite{DBLP:journals/corr/abs-2002-04745}. Some interpretability works ignore this layer because it has a linear-up-to-scale structure, absorbing the linear part into adjacent weights. In this work we argue that the layer is important, because \textbf{Pre-Norm causes independent linear subspaces to interfere through a common normalisation factor, preventing their separation by linear attention layers}. 

The purpose of this work is to ask: if the use of independent subspaces is generally important, what are the expected consequences of \texttt{Pre-Norm} for (i) the latent representation structure, and (ii) circuit stability? To answer this, we take an abstract approach that complements direct interpretability by considering general behaviour beyond the interpretable limit. Our contributions are:
\begin{enumerate}
	\item \textbf{Conceptual:} we identify interference between independent subspaces as a potential destabiliser of circuits caused by \texttt{Pre-Norm}. We suggest \textit{separability of latent subspaces} as a target for study, and show it is easily satisfied by the alternative \texttt{QKV-Norm}. This differs from \texttt{Pre-Norm} by placing the normalisation layer after the linear operators. It is similar to \texttt{QK-Norm}, for which sparse evidence currently exists \cite{DBLP:journals/corr/abs-2010-04245,wortsman2023smallscale,dehghani2023scaling}.
	\item \textbf{Theoretical:} we formalise a \textit{semantic subspace} as any independent subspace of the latent representation that can fully determine the attention distribution. We show that \texttt{Pre-Norm} can only achieve this when semantic subspaces are spherical and mutually orthogonal. By contrast, \texttt{QKV-Norm} requires only that subspaces be linearly independent, matching the \texttt{No-Norm} case in this sense. We study the stability of attention to subspace interference, predicting a potentially problematic phenomenon of \textit{circuit collapse} when a sparse-attention distribution changes which embedding it attends to. 
	%\item \textbf{Experimental:} we measure the sensitivity of trained models to simulated interference in a numerical addition task. We find that per-token accuracy degrades by $\gtrsim 10\%$ when vector norms are perturbed by $\mathcal{O}(1\%)$. Supporting our theory, we find that (i) \texttt{Pre-Norm} models induce a narrower distribution of embedding norms than \texttt{QKV-Norm}, and (ii) the circuit collapse phenomenon occurs in practice when norms are perturbed by $\mathcal{O}(5-10\%)$.
	\item \textbf{Experimental:} we measure the sensitivity of trained models to simulated interference in a numerical addition task. Constraining our predictions, we find that (i) \texttt{Pre-Norm} models induce a narrower distribution of embedding $L_2$-norms than \texttt{QKV-Norm}, (ii) we bound the spread of $L_2$-norms to $\pm20\%$ with 90\% coverage, and (iii) the circuit collapse phenomenon occurs at a rate of 1\% when norms are perturbed by $\mathcal{O}(10\%)$.
\end{enumerate}

\section{The idea}
\label{sec: idea}

\noindent
\textbf{Independent subspaces are observed in real-world transformer circuits}

Before providing a formal definition in section~\ref{sec: theory: residual structure}, we explain what we mean by a \textit{semantic subspace} of the latent representation. To emphasise that this is observed in real-world models, we use a known example: the \textit{induction circuit} \cite{olsson2022context,elhage2021mathematical}. This two-layer circuit emerges in next-token-prediction models and implements a simple contextual reasoning algorithm called \textit{prefix-matching}. %\footnote{\textit{In-context learning} is when the answer to the prediction task does not exist in training, and it can only be solved by parsing the observed provided at inference-time.}

Consider text to be a sequence of tokens\footnote{In this example, we tokenise per-word to help with visualisation.}, and our task is to predict the next token at every point. The induction circuit solves this by copying a previous example from the context window: e.g. if the input includes the phrase ``Harry Potter'' and the last observed word was ``Harry'', the induction circuit will predict that ``Potter'' comes next. This solves the task even if the combination ``Harry Potter'' never occurred in the training data.
 %Importantly, the model may have nave seen the combination ``Harry Potter'' at training time, and so \textit{the only} way to solve this task is by consulting the context.

To achieve this, we initially create an embedding for each token, encoding it's \textit{position} and \textit{type}. Attention layers then \textit{copy information between embeddings in a directed way}, using two components that determine (i) \textit{which} embeddings to extract information from, and (ii) \textit{what} to extract. Remarkably, the model learns to implement logical gates that we will call ``match\&pass'', internally composing the algorithm:

\includegraphics[page=2, width=\textwidth, clip=True, trim=0 17cm 0 0cm]{figures/Paper_diagrams.pdf}

Each match\&pass step operates only on an independent subspace of information, which we will call a \textit{semantic subspace}. In this example, there are four semantic subspaces corresponding to \textit{position}, \textit{type}, \textit{prev-type}, and \textit{pred-suffix}. We observe that the latent embeddings can contain various information, and it is instructive to think of them as memory buffers rather than tokens. The principle of \textit{composing logical operations that act on latent semantic subspaces} is also observed in the more complex example of indirect-object identification in GPT2-Small \cite{wang2022interpretability}.

%\clearpage
%\textbf{Linear-attention requires linearly-independent semantic subspaces}

%t $P_\alpha x_\alpha = x_\alpha$ and $P_\alpha x_{\beta\neq\alpha} = 0$, then $P_\alpha x = P_\alpha \sum_\beta x_\beta = x_\alpha$.
%\begin{equation}
%	P_\alpha x ~=~ P_\alpha \sum_\beta x_\beta ~=~ \sum_\beta P_\alpha x_\beta ~=~ P_\alpha x_\alpha ~+~ \sum_{\beta\neq\alpha} P_\alpha x_\beta ~=~ P_\alpha x_\alpha ~+~ 0 ~=~ x_\alpha
%\end{equation}
%can be used to extract an individual semantic for use in e.g. the match\&pass operation.

\vspace{0.2cm}
\noindent
\textbf{The problem with Pre-Norm}

We express the latent embeddings as $x = \sum_\alpha x_\alpha$ where $x_\alpha$ encodes the value of concept $\alpha$. This is important, because linear-attention layers extract information from $x$ using linear operators (section~\ref{sec: formulation}), and can only isolate $x_\alpha$ if each subspace $\{x_\alpha~|~\alpha\}$ is \textit{linearly independent}. In other words, there must always exist a linear projection operator $P_\alpha$ such that $P_\alpha x = x_\alpha$.

Most transformers use either \texttt{RMSNorm} \cite{NEURIPS2019_1e8a1942} or \texttt{LayerNorm} \cite{DBLP:journals/corr/BaKH16} for their internal normalisation layers. Geometrically, \texttt{RMSNorm} projects a vector $z \in \mathbb{R}^N$ onto the unit-sphere $S^{N-1}$ according to
\begin{equation}
	z ~\rightarrow~ \frac{z}{|z|} ~~~~~~~~~~~~~~~~~~~\mathrm{where}~|z| ~\triangleq~ \sqrt[+]{\sum_{i=1}^N z_i^2} ~~\mathrm{is~the}~L_2\mathrm{-norm.}
\end{equation}
\texttt{LayerNorm} is similar, projecting onto the sphere $S^{N-2}$ defined perpendicular to the direction $1^N$. This does not affect our analysis, and we focus on \texttt{RMSNorm} for simplicity. Normalisation layers sometimes also include gain and/or bias parameters, applying a stretch-and-translate to the sphere. \texttt{Pre-Norm} \cite{DBLP:journals/corr/abs-2002-04745} normalises the latent embeddings at the \textit{input} to every attention layer. Consider the example $x = x_\mathrm{pos} + x_\mathrm{type} + x_\mathrm{prev-type}$. %This is transformed to
%\begin{equation}
%\frac{x}{|x|} ~=~ \frac{x_\mathrm{pos} ~+~ x_\mathrm{type} ~+~ x_\mathrm{prev-type}}{|x_\mathrm{pos} ~+~ x_\mathrm{type} ~+~ x_\mathrm{prev-type}|}
%\end{equation}
Applying \texttt{Pre-Norm}, we find $P_\mathrm{pos}  x = x_\mathrm{pos}$ is replaced by
\begin{equation}
P_\mathrm{pos}  \frac{x}{|x|} ~=~  \frac{P_\mathrm{pos} x}{|x|} ~=~ \frac{x_\mathrm{pos}}{|x_\mathrm{pos} ~+~ x_\mathrm{type} ~+~ x_\mathrm{prev-type}|}
\end{equation}
\textbf{Therefore it is impossible for a linear-attention layer to extract $x_\mathrm{pos}$ without interference from $x_\mathrm{type}$ and $x_\mathrm{prev-type}$, unless $|x_\mathrm{pos} + x_\mathrm{type} + x_\mathrm{prev-type}|$ is a constant}. In general, we have $P_\alpha \frac{x}{|x|} = \frac{x_\alpha}{|\sum_\beta x_\beta|}$, and semantic subspaces are entangled unless $|\sum_\alpha x_\alpha|$ is constant. This is only possible if $|x_\alpha|^2 = const_\alpha ~\forall~ \alpha$, i.e. every subspace is a sphere, and $x_\alpha^Tx_\beta = 0 ~\forall~ x_\alpha,x_{\beta \neq \alpha}$, i.e. all spheres are orthogonal (to maintain independence). This has several possible implications:
\begin{enumerate}
    \item It is a restrictive structure that must be learned during training, with unknown difficulty. Finite steps of gradient descent may separate the model from the manifold of acceptable representations, hindering the learning of circuit components that require semantic separation, like match\&pass, especially when training with large learning rates.
    \item The constraint $|x_\alpha|^2 = const_\alpha$ removes a degree of freedom for every $\alpha$, reducing the information capacity of the embedding space. For example, an embedding on $\mathbb{R}^5$ could have the two-subspace structure $\mathcal{S}^2 \bigoplus \mathcal{S}^1$ but not $\mathcal{S}^2 \bigoplus \mathcal{S}^2$.
	\item We hypothesise that the structure may be violated by (i) a tradeoff with other representational effects, (ii) imperfect model training, or (iii) encountering unexpected semantic combinations at inference-time when generalising out-of-distribution. These would cause semantic subspaces to interfere through their common normalisation factor, manifesting as noise on the $L_2$-norms of the \{query,~key,~value\} vectors.
	\item It is a structure that we can search for empirically.
\end{enumerate}

\vspace{0.2cm}
\noindent
\textbf{A possible solution: QKV-Norm}

A natural fix could be to apply the normalisation layer \textit{after} the linear operators. In practice this means that we normalise the \{query,~key,~value\} vectors, called \texttt{QKV-Norm} and defined in section~\ref{sec: formulation}.

\vspace{0.2cm}
\noindent
\textbf{Paper strategy}

Our work is based on three key observations: (i) semantic subspaces are observed in known circuits, (ii) they contribute to the model behaviour, and (iii) \texttt{Pre-Norm} requires them to follow a strict latent embedding structure or else interfere through the $L_2$-norms of the \{query,~key,~value\} vectors.

However, it is difficult to demonstrate specific examples of subspace interference. Firstly, a fully-converged model should learn to manage interference for in-distribution examples. Instead, we expect it to concern (i) training stability, (ii) model inductive bias, and (iii) out-of-distribution behaviour. Secondly, circuit explainability is difficult, only being achieved in simple cases. In general we expect circuits to become complicated, contain steps that are harder to interpret than match\&pass, and exploit non-interpretable latent subspaces. Difficulty is further increased by polysemanticity \cite{elhage2022superposition}, the ability for heads and features to change behaviour according to context.

In this work, we take an abstract approach instead. We formally define latent semantic separability, then investigate the theoretical consequences for \texttt{Pre-Norm} architectures if this behaviour is important generally. This allows us to make testable predictions about representation structure and model stability without needing to fully reverse-engineer a network or explain subspaces in human terms. We then place some data-driven limits on the effect size. Nonetheless, direct observation remains important, and we hope that future works can confirm or falsify the importance of the proposed representation structure and interference effect.

%%
%%  Section:  Related Works
%%
\section{Related Works}
\label{sec: related works}

Our work is motivated by transformer circuit discovery \cite{olsson2022context,elhage2021mathematical,wang2022interpretability,stolfo2023mechanistic,GoldowskyDill2023LocalizingMB,Ferrando2024InformationFR} and formation \cite{singh2023the,singh2024needs}. See \cite{ferrando2024primer} for a recent review of interpretability for language decoder models, with a list of known logical operations implemented by attention heads. This builds upon works in BERTology \cite{DBLP:journals/corr/abs-2002-12327,Devlin2019BERTPO}. We study normalisation, for which several formulations have been proposed \cite{DBLP:journals/corr/abs-2002-04745,nguyen-salazar-2019-transformers,nguyen-chiang-2018-improving,shleifer2021normformer,NEURIPS2019_2f4fe03d}. Our \texttt{QKV-Norm} variant is similar to \texttt{QK-Norm}, which is studied by \cite{DBLP:journals/corr/abs-2010-04245,wortsman2023smallscale,dehghani2023scaling} for asymptotic performance and training stability at large learning rates. These are motivated by logit-regularisation, whereas we are motivated by representational inductive bias and stability to latent semantic interference. %Our work therefore contributes an additional motivation for studying the performance of these normalisation methods.

We highlight other works that study transformer normalisation through its geometric interpretation as a projection onto a sphere. \cite{kobayashi-etal-2021-incorporating} investigated the role of normalisation in mixing the attention output with the residual stream in \texttt{Post-Norm} models, but does not consider \texttt{Pre-Norm}. \cite{brody2023expressivity} studies the computational abilities of \texttt{Pre-LayerNorm} architectures, in particular demonstrating that projection onto a sphere ensures that all keys reside on their own convex hull, preventing them from becoming ``unselectable''. \cite{Molina2023TravelingWA} interprets the latent embeddings of \texttt{Pre-Norm} models as a trajectory on a sphere. These works do not consider the interference of semantic subspaces. \cite{Dong2021AttentionIN} and the contemporary work \cite{wu2024role} study the role of \texttt{LayerNorm} in the related phenomenon of embedding rank collapse.

We highlight the contemporary work of \cite{wang2024understanding}, who also study multi-step contextual reasoning in transformers using matching operations over independent subspaces, for both \texttt{Pre-Norm} and \texttt{Post-Norm}. This builds upon \cite{boixadsera2024when}, who study the learning of abstract symbolic reasoning in transformers, and works that manipulate the flow of information to promote algorithmic reasoning, e.g. \cite{csordas2022the}.

We are not aware of previous works that study the impact of \texttt{Pre-Norm}'s spherical geometry on the structure of latent subspaces. However, many works consider linear subspaces, described in the following paragraph. These results are directly applicable to the \texttt{No-Norm} and \texttt{QKV-Norm} methods in this work, although \texttt{QKV-Norm} applies a subsequent spherical projection. \cite{Lamb2021TransformersWC} design subspace separability into their model by decoupling the normalisation layers for different mechanisms.

Works on vector embeddings \cite{word2vec1,NIPS2013_9aa42b31,Pennington2014GloVeGV} and the \textit{linear representation hypothesis} \cite{liguistic_regularities,Park2023TheLR,jiang2024origins} study the emergence of linear subspaces that encode separable concepts in embedding-unembedding models, using both interpretation and intervention techniques. Many works search for linear subspaces/directions in a transformer representation (e.g. linear probes \cite{semantic_subspace_probing,belinkov-2022-probing}) or search for faithful causal abstractions (e.g. \cite{Geiger2023FindingAB}), with a survey provided in \cite{ferrando2024primer} sections 3-4. %Some works focus on human interpretability in the input and output embeddings, and whilst we emphasise non-human-interpretable latent concepts and circuit stability, both share the common feature of linear independence. 
We also highlight works that study the use of features in linear superposition \cite{superposition1,elhage2022superposition}. This allows a model to store more features than it has dimensions, at the cost of interference in their linear projections. %, which we note is a different type of interference to the one induced by \texttt{Pre-Norm}.

The terminology of \textit{semantic subspaces} is used more generally, e.g. \cite{semantic_subspace_probing,5596640,Coenen2019VisualizingAM}. We consider a definition that does not require humans to define the separable concepts, only that abstract latent features remain independent in an attention layer. We also highlight works that study subspaces of static (model input) and contextual (latent or model output) embeddings in transformers, e.g. \cite{Hewitt2019ASP,Coenen2019VisualizingAM,ethayarajh-2019-contextual,song2024uncovering,muppet2022,hernandez2024linearity,chi-etal-2020-finding,cai2021isotropy,Hernandez2021TheLL} (review in \cite{DBLP:journals/corr/abs-2002-12327}). These are relevant because they also decompose embeddings into a combination of abstract subspaces, capturing different semantic and syntactic structures in a natural language setting. These may be used as semantic subspaces in our work. We highlight \cite{song2024uncovering} which studies interference between positional and contextual components using a decomposition similar to ours, and also experiments using a next-token addition task.

%%
%%  Section:  Formulation
%%
\section{Formulation}
\label{sec: formulation}

%We provide a concise summary of our formulation, with further details in appendix~\ref{}.

%We use similar terminology to \cite{elhage2021mathematical}. 

Consider the \texttt{No-Norm} case. Let $\mathcal{X}$ be an unordered set of message receiving tokens, and $\mathcal{Y}$ the message senders. Let $x \in \mathbb{R}^{N_x}$ be the $N_x$-dimensional representation of an element in $\mathcal{X}$, and $y_t \in \mathbb{R}^{N_y}$ be the $t^\mathrm{th}$ element in $\mathcal{Y}$, with $1 \leq t \leq T$. For self-attention we have $\mathcal{X}=\mathcal{Y}$. Let $W_Q\in\mathbb{R}^{N_{qkv}\times N_x}$ and $W_K\in\mathbb{R}^{N_{qkv}\times N_y}$ be the query and key weight matrices, with associated vectors $q=W_Qx\in\mathbb{R}^{N_{qkv}}$ and $k_t=W_Ky_t\in\mathbb{R}^{N_{qkv}}$ on an $N_{qkv}$-dimensional latent space. We do not include biases in $\{q,~k_t\}$ because they contribute terms that are nullified by the \texttt{softmax}, or are reproduced by constant directions in $x$ (Theorem~\ref{theorem: decomposition}).  We define dot-product attention \textit{scores} as:
\begin{equation}
    w_t ~=~ q^Tk_t ~=~ x^T W_Q^T W_K y_t ~=~ x^T W_{QK} y_t
\end{equation}
where $W_{QK}\triangleq W_Q^TW_K \in \mathbb{R}^{N_x\times N_y}$ is a matrix with $Rank(W_{QK}) \leq \min(N_x,N_y,N_{qkv})$. This is the maximum span of the attended subspace in $\{x,y_t\}$. The attention \textit{weights} are
\begin{equation}
    a_t ~=~ \texttt{softmax}\left(w_t\right) ~=~ \frac{e^{w_t}}{\sum_{t'} e^{w_{t'}}} ~~~~~~.
\end{equation}
Let $v_t=W_Vy_t\in\mathbb{R}^{N_x}$ be the value vectors with $W_V\in\mathbb{R}^{N_{qkv}\times N_y}$. We do not include biases in $v_t$ because they carry no dependence on the attended token. Each token emits the message $m_t = W_Ov_t \equiv W_OW_Vy_t \triangleq W_{OV}y_t$ where $W_O = \mathbb{R}^{N_x\times N_{qkv}}$ is the output-matrix. Each attention-head updates $x$ by adding the attention-weighted convex combination of messages, $x \rightarrow x + \Delta x$ with $\Delta x = \sum_{t} a_t m_t$. We usually run $H$ attention-heads in parallel, giving the total update:
\begin{equation}
     x ~\rightarrow~ x ~ + ~\sum_{h=1}^{H} \sum_{t=1}^{T} a_t^{(h)} m_t^{(h)} ~~~~~~~~~~~~~~~~~~ \text{Multi-head~attention}
\end{equation}
with unique weights $\{W_Q^{(h)},W_K^{(h)},W_V^{(h)},W_O^{(h)}\}$ for each head index $h$.

%Each attention head modifies the token representation $x$ via the linear addition of information, called \textit{annotation}. We often view $x$ as a buffer of information called the \textit{residual stream}. Independent circuits can compose operations by channelling information through the residual streams of different tokens. 

We now introduce normalisation layers. Let $z\in\mathbb{R}^{N_z}$ be any $N_z$-dimensional vector, then $\texttt{N}(z;\alpha_z):\mathbb{R}^{N_z}\rightarrow\mathbb{R}^{N_z}$ is a normalisation function with parameters $\alpha_z$. We consider two such functions:
% \begin{equation}
% \begin{split}
%     \texttt{LayerNorm}\left(z;~\alpha_z\right)  
%         ~&\equiv~ \mathrm{LN}\left(z;~\alpha_z\right) 
%         ~=~  \frac{1}{\sigma(z)} \mathrm{diag}\left(\alpha_z\right)P_\perp z ~=~ \frac{\sqrt{N_z}}{|z_\perp|} \mathrm{diag}\left(\alpha_z\right) z_\perp   \\
%     \texttt{RMSNorm}\left(z;~\alpha_z\right) 
%         ~&\equiv~ \mathrm{VN}\left(z;~\alpha_z\right) 
%         ~=~  \frac{\sqrt{N_z}}{|z|} \mathrm{diag}\left(\alpha_z\right) z \\
% \end{split}
% \end{equation}
\begin{equation}
    \texttt{RMSNorm}\left(z;~\alpha_z\right) =  \frac{\sqrt{N_z}}{|z|} \mathrm{diag}\left(\alpha_z\right) z 
%~~\cite{NEURIPS2019_1e8a1942}
~~~~~~~~~~~~~
    \texttt{LayerNorm}\left(z;~\alpha_z\right) = \frac{\sqrt{N_z}}{|z_\perp|} \mathrm{diag}\left(\alpha_z\right) z_\perp 
%~~\cite{DBLP:journals/corr/BaKH16}
\end{equation}
\cite{NEURIPS2019_1e8a1942,DBLP:journals/corr/BaKH16} where $P_\perp \triangleq \mathrm{diag}\left(1^{N_z}\right) - 1^{N_z}{1^{N_z}}^T$ is a linear operator that subtracts the mean of $z$ from every component, $1^{N_z}$ is vector of ones, and $z_\perp \triangleq P_\perp z$ is the component of $z$ perpendicular to $1^{N_z}$.

The \texttt{Pre-Norm} strategy means applying normalisation to the inputs $\{x,y_t\}$. The \texttt{QKV-Norm} strategy means applying normalisation to the vectors $\{q,k_t,v_t\}$. We then have three cases:

%\vspace{0.3cm}
\noindent
\begin{tabular}{rr|c|c|c}
  ~  &  ~  &  $w_t$  &  $v_t$  &  Norm params  \\
\hline
                & \texttt{No-Norm}    &  $x^T ~W_{QK}~ y_t$  &  $W_V~y_t$ & - \\
     (baseline) & \texttt{Pre-Norm}   &  $\texttt{N}\left(x;\alpha_x\right)^T ~W_{QK}~ \texttt{N}\left(y_t;\alpha^K_y\right)$  & $W_V~\texttt{N}\left(y_t;\alpha^V_y\right)$ & $\{\alpha_x,~\alpha^K_y,~\alpha^V_y\}$ \\

     (alternate)     & \texttt{QKV-Norm}  &  $\texttt{N}\left(W_Q x;\alpha_q\right)^T \texttt{N}\left(W_K y_t;\alpha_k\right)$  & $\texttt{N}\left(W_V y_t;\alpha_v\right)$ & $\{\alpha_q,~\alpha_k,~\alpha_v\}$ \\
\end{tabular}

We note that several of these degrees of freedom are redundant and could be combined, e.g. $\alpha_q$ and $\alpha_k$. We do not consider these variations (i) because they are not relevant for the results of this paper, and (ii) to standardise the number of training parameters.

%%
%%  Section:  Theory: residual structure
%%
%\clearpage
\section{Theory: representation structure required for independent subspaces}
\label{sec: theory: residual structure}

Let $\mathbb{S}^N \equiv \mathbb{R}^N$ be an $N$-dimensional latent representation of $\mathcal{X}$ or $\mathcal{Y}$.

\begin{mdframed}[backgroundcolor=red!5]
\textbf{[Definition] ~ Semantic subspace:} any independent $N_\alpha$-dimensional subspace $\mathbb{S}_\alpha^{N_\alpha} \subset \mathbb{S}^N$ for which every element may be uniquely identified by some parameters $\theta_\alpha$, such that it is possible for the attention scores $w_t$ to be fully specified by $\theta_\alpha$. \textbf{Semantic separability:} ability for parallel heads to be fully specified by different semantic subspaces.
%\textbf{[Corollary]} ~ Using linear-attention, the residual representation may be written as the tensor sum $\mathbb{S} = \bigoplus_\alpha \mathbb{S}_\alpha$ where $\{\alpha\}$ includes the special subspaces \texttt{other} (a subspace that carries degrees-of-freedom but is non-attended) and \texttt{null} (the attention null-space), and $\mathbb{S}_\alpha$ has $N_\alpha$-dimensions such that $N=\sum_\alpha N_\alpha$.
\end{mdframed}

Let $\{\alpha\}$ be the set of indivisible semantic subspaces. This can be seen as a \textit{co-ordinate system} for the attendable embedding space. Semantic separability requires that each co-ordinate $\alpha$ be independently measurable by an attention head. Let $\mathbb{S}^N$ contain $N_s$ indivisible semantic subspaces $1\leq\alpha\leq N_s$. Then $\mathbb{S}^N=\prod_\alpha \mathbb{S}^{N_\alpha}_\alpha \bigoplus\mathbb{S}_\mathrm{null}$ such that $\sum_\alpha N_\alpha \leq N$ satisfies semantic separability, where $\prod_\alpha,\bigoplus$ are Cartesian products and $\mathbb{S}_\mathrm{null}$ is a separable space of non-attended information. 

The following theorems derive the representation structure required for semantic separability:

\begin{mdframed}[backgroundcolor=blue!5]
\underline{\textbf{Semantically separable representation structures}}
~~~~~~~~~~~~~~~~~~~~~~~~~~~~~~~~\textit{[Proofs in appendix~\ref{appendix: proofs}]}

\begin{theorem}
    \texttt{No-Norm}: If two heads with finite non-zero temperature attend to different semantic subspaces, the subspaces must be linearly independent $\mathbb{S}^{N_\alpha}_\alpha \equiv \mathbb{R}^{N_\alpha}$. Corollary: $W_{QK}$ is a low-rank matrix with (left and right) null-spaces that span all non-attended information.
\label{theorem: structure: no-norm}
\end{theorem}

\begin{theorem}
    \texttt{Pre-Norm}: Semantic subspaces must be represented as orthogonal spheres $\mathbb{S}^{N_\alpha}\equiv\mathcal{S}^{N_\alpha-1}$ defined using the $L_2$-norm. Corollary: if either orthogonality or constant-norm are violated, semantic subspaces interfere through a multiplicative factor on $w_t$.
\label{theorem: structure: pre-norm}
\end{theorem}

\begin{theorem}
    \texttt{QKV-Norm}: Semantic subspaces must be linearly independent.
\label{theorem: structure: qkv-norm}
\end{theorem}
\end{mdframed}

We note that every linear subspace $\mathbb{R}^{N_\alpha}$ has $N_\alpha$ continuous degrees of freedom, whilst $\mathcal{S}^{N_\alpha-1}$ has only $N_\alpha-1$, the other being removed by the fixed-norm constraint. The subspace $\mathcal{S}^0$ is allowed and may be seen as a binary variable with values $\pm const_\alpha$, and the total representation can store $N_s$ such variables. For \texttt{QKV-Norm}, we note that the residual subspace $\mathbb{R}^{N_\alpha}$ only contributes $N_\alpha-1$ continuous degrees of freedom to the attention calculation, because we apply the projection $\mathbb{R}^{N_\alpha} \rightarrow \mathcal{S}^{N_\alpha-1}$ after extracting the subspace. Table~\ref{table:residual structures main} provides a summary.

\vspace{0.2cm}
\noindent
\textbf{Structure of messages}

We note the special case of compositional annotation, in which a layer creates a semantic subspace that is extracted by a later layer. This is used by circuits including the \textit{induction circuit} \cite{elhage2021mathematical} described in section~\ref{sec: idea}. By normalising the inputs, \texttt{Pre-Norm} induces a \textit{spheroid} message structure close to the \textit{sphere} required for separability in later layers. This may facilitate compositional annotation, aiding in circuit-formation. Message structures are summarised in Table~\ref{table:value structures main}.

\begin{table}[h]
\centering
\begin{tabular}{rlll}
    Strategy                 &  $\mathbb{S}^N$  &  Representation structure   &  Attendable d.o.f.  \\
    \hline
    \texttt{No-Norm}         &  $\prod_\alpha \mathbb{R}^{N_\alpha}$     &  Linearly independent subspaces  &  $N$            \\
    \texttt{Pre-LayerNorm}   &  $\prod_\alpha \mathcal{S}^{N_\alpha-1}$  &  Orthogonal spheres $\perp 1^N$  &  $N - N_s - 1$  \\
    \texttt{Pre-RMSNorm}  &  $\prod_\alpha \mathcal{S}^{N_\alpha-1}$  &  Orthogonal spheres              &  $N - N_s$      \\
    \texttt{QKV-Norm}     &  $\prod_\alpha \mathbb{R}^{N_\alpha}$     &  Linearly independent subspaces  &  $N - N_s$      \\
\end{tabular}
\vspace{0.1cm}
\caption{Representation structure required for semantic separability; \textit{d.o.f.} means \textit{degrees of freedom}.}
\label{table:residual structures main}
\vspace{0.1cm}
\begin{tabular}{rlll}
    Strategy                     &  $m_t$  &  Structure of $m_t$  &  Compositional annotation if  \\
    \hline
    \texttt{No-Norm}    &   $W_{OV} y_t$                       &  Linear      &   $m_t$ on independent subspace  \\
    \texttt{Pre-Norm}   &   $W_{OV} \texttt{N}(y_t;\alpha_v)$  &  Spheroid    &   $m_t$ on orthogonal sphere     \\
    \texttt{QKV-Norm}   &   $W_O\texttt{N}(W_Vy_t;\alpha_v)$   &  Spheroid    &   $m_t$ on independent subspace  \\
\end{tabular}
\vspace{0.1cm}
\caption{Summary of message structures induced by different placements of normalisation layer.}
\label{table:value structures main}
\end{table}

%%
%%  Section:  Theory: circuit stability
%%
%\clearpage
\section{Theory: stability to subspace interference}
\label{sec: theory: circuit stability}

We now investigate the impact of interfering subspaces. Consider the almost-separable limit, modelling interference as a random infinitesimal perturbation of the vectors $\{q,k_t,m_t\}$. Let $\epsilon$-symbols denote perturbations such that $\epsilon^{\Delta x(q)} \rightarrow \frac{\partial \Delta x}{\partial q} \epsilon^q$ for $\epsilon^q \rightarrow 0$ is the change of $\Delta x$ induced by $\epsilon^q$. %We then study specific normalisation strategies by considering the structure of perturbation they induce. Specifically, we expect \texttt{Pre-Normalisation} to entangle semantics via a multiplicative factor, therefore $\epsilon^q \propto q$ etc.
We consider (i) the sparse limit, in which the attention is concentrated entirely on a single embedding, and (ii) the isotropic limit, in which it is distributed evenly among embeddings. We are particularly interested in the sparse case, since this highly directed flow of information is used by match\&pass, although semantic separation can also be used by non-sparse heads. %We also study the opposite limit of isotropic attention, which attends to all tokens equally.

\begin{mdframed}[backgroundcolor=red!5]
\textbf{[Definition] ~ Sparse attention:} the low-temperature limit $a_t \approx \delta_{tt^*}$ and $\Delta x = m_{t^*}$, where $\delta$ is the Kronecker delta. This occurs when there is a large difference between the top two scores: $t^* = \mathrm{argmax}_t w_t$ and $w_{t^*} - \max_{t\neq t^*} w_t \gg 1$. \textbf{Isotropic attention:} the high-temperature limit $a_t = \frac{1}{T}$ and $\Delta x = \langle m_t\rangle_t$. This occurs when $w_t$ is constant, requiring $q=0$ or constant $k_t$.
\end{mdframed}

\begin{mdframed}[backgroundcolor=blue!5]
\underline{\textbf{Stability of attention updates to perturbations on q, k, v}}
~~~~~~~~~~~~~~~~~~~\textit{[Proofs in appendix~\ref{appendix: proofs}]}

\begin{theorem}
    Consider independent infinitesimal perturbations on queries $\epsilon^q \in \mathbb{R}^{N_{qkv}}$, keys $\epsilon^k_t \in \mathbb{R}^{N_{qkv}}$, and messages $\epsilon^m_t \in \mathbb{R}^{N_{qkv}}$. These propagate onto $\Delta x = \sum_{t}a_tm_t$ as
    \begin{align}
        \epsilon^{\Delta x(q)} ~~&\xrightarrow[\epsilon^q\rightarrow0]{\mathrm{~~~~perturb~q~~~~}}~~ \mathop{\mathbb{E}}_{a_t} \Big[ m_t {\tilde k}_t^T \Big] \epsilon^q ~~~~~~~~~~~~~~~~~  {\tilde k}_t ~\triangleq~ k_t ~- \mathop{\mathbb{E}}_{a_t} \Big[ k_t \Big] \\
        \epsilon^{\Delta x(k)} ~~&\xrightarrow[\epsilon^k_t\rightarrow0]{\mathrm{~~~~perturb~k~~~~}}~~ \mathop{\mathbb{E}}_{a_t} \Big[ {\tilde m}_t {\epsilon^k_t}^T \Big] q ~~~~~~~~~~~~~~~~~  {\tilde m}_t ~\triangleq~ m_t ~- \mathop{\mathbb{E}}_{a_t} \Big[ m_t \Big] \\
        \epsilon^{\Delta x(m)} ~~&\xrightarrow[\epsilon^m_t\rightarrow0]{\mathrm{~~~~perturb~m~~~~}}~~ \mathop{\mathbb{E}}_{a_t} \Big[ \epsilon^m_t \Big]
    \end{align}
    where ${\tilde z}_t$ is the value of $z_t$ measured from the attention-weighted centroid $\mathbb{E}_{a_t}[z_t] = \sum_t a_t z_t$.
\label{theorem: stability: general}
\end{theorem}

\begin{theorem}
    For sparse attention:
    \begin{equation}
        \epsilon^{\Delta x(q)} \xrightarrow[\epsilon^q\rightarrow0]{\mathrm{~~perturb~q~~}} 0   ~~~~~~~~~~
        \epsilon^{\Delta x(k)} \xrightarrow[\epsilon^k_t\rightarrow0]{\mathrm{~~perturb~k~~}} 0   ~~~~~~~~~~
        \epsilon^{\Delta x(m)} \xrightarrow[\epsilon^m_t\rightarrow0]{\mathrm{~~perturb~m~~}} \epsilon^m_{t^*}
    \end{equation}
    i.e. the message is stable with respect to small interference in the queries and keys. Interference in the selected value is linearly transferred onto the message.
\label{theorem: stability: sparse}
\end{theorem}

\begin{theorem}
    For isotropic attention:
    \begin{equation}
        \epsilon^{\Delta x(q)} \xrightarrow[\epsilon^q\rightarrow0]{\mathrm{perturb~q}} \langle m_t {\tilde k}_t^T \rangle_t \epsilon^q ~~~~~~~~
        %\epsilon^{\Delta x(q)} \xrightarrow[\epsilon^q\rightarrow0]{\mathrm{~perturb~q~}} 0   ~~~~~~~~~
        \epsilon^{\Delta x(k)} \xrightarrow[\epsilon^k_t\rightarrow0]{\mathrm{perturb~k}} \langle {\tilde m}_t {\epsilon^k_t}^T \rangle_t ~q   ~~~~~~~~
        \epsilon^{\Delta x(m)} \xrightarrow[\epsilon^m_t\rightarrow0]{\mathrm{perturb~m}} \langle \epsilon^m_t \rangle_t
    \end{equation}
    % \begin{align}
    %     \epsilon^{\Delta x(q)} ~&\xrightarrow[\epsilon^q\rightarrow0]{\mathrm{~~perturb~q~~}}~ 0 \\
    %     \epsilon^{\Delta x(k)} ~&\xrightarrow[\epsilon^k_t\rightarrow0]{\mathrm{~~perturb~k~~}}~ \langle {\tilde v}_t {\epsilon^k_t}^T \rangle_t ~q  \\
    %     \epsilon^{\Delta x(v)} ~&\xrightarrow[\epsilon^v_t\rightarrow0]{\mathrm{~~perturb~v~~}}~ \langle \epsilon^v_t \rangle_t
    % \end{alig
    N.B. isotropy requires $k_t=const$ or $q=0$. Lemma 1: the update is stable to noisy $q$ when $k_t=const$, or when $m_t \perp k_t$ (c.f. keys and messages from independent subspaces). Lemma 2: the update is stable to noisy $k_t$ when $q=0$, or when $m_t \perp \epsilon_t^k$. Lemma 3: the update is stable to noisy $m_t$ when $\langle \epsilon^m_t \rangle_t=0$. Other cases propagate linearly.
\label{theorem: stability: isotropic}
\end{theorem}
\end{mdframed}

The stability of sparse attention is because \texttt{softmax} becomes an \texttt{argmax} for low-temperature heads, which is only sensitive to the order of $w_t$. However, this introduces a different vulnerability when perturbations cause the order of $w_t$ to change, as the attention distribution undergoes a phase transition to select a different token. We call this \textit{circuit collapse}. For example, the induction circuit collapses when the operation \textit{attend to the previous token} attends to any other token because of interference.

\begin{mdframed}[backgroundcolor=red!5]
\textbf{[Definition] ~ Circuit collapse:} spontaneous phase transition in which a sparse attention distribution selects a different token due to noise on $\{q, k_t\}$. Let $\epsilon^w_t = k_t^T\epsilon^q + q^T\epsilon^k_t + \mathcal{O}({\epsilon^q}^T\epsilon^k_t)$ be perturbations on $w_t$ that result from $\epsilon^q$ and $\epsilon^k_t$. Circuit collapse occurs when there exists a $t \neq t^*$ for which $w_{t^*} - w_t < \epsilon^w_t - \epsilon^w_{t^*}$.
\end{mdframed}

We now study the $L_2$-norm interference that we expect to be induced by \texttt{Pre-Norm} when semantic separability is violated. This is characterised by perturbations that are parallel to their corresponding vector. Theorem~\ref{theorem: multiplicative stability: sparse} shows the conditions under which we expect circuit collapse to occur.

\begin{mdframed}[backgroundcolor=blue!5]
\underline{\textbf{Stability of attention updates to scaling of q, k, v}}
~~~~~~~~~~~~~~~~~~~~~~~~~~~~~~~~~~~~\textit{[Proofs in appendix~\ref{appendix: proofs}]}

\begin{theorem}
    Sensitivity of sparse attention to multiplicative perturbations $\epsilon^q = \kappa^q q$ and $\epsilon^k = \kappa^k_t k_t$ with $\kappa^q,\kappa^k_t\ll1$. Circuit collapse occurs when $\exists~ t \neq t^*$ for which:
    \begin{equation}
        \frac{w_{t^*}}{w_t} ~\begin{cases} ~<~ \lambda_w & \mathrm{if}~ w_t \left(1 + \kappa^q + \kappa^k_{t^*}\right) > 0 \\
        ~>~ \lambda_w & \mathrm{otherwise} \\ \end{cases}
        ~~~~~~~~~~~~~ \lambda_w ~\triangleq~ \frac{1 + \kappa^q + \kappa^k_t}{1 + \kappa^q + \kappa^k_{t^*}}
    \end{equation}
    where temperature cancels in the fraction. \textbf{Attention is fully stable above the critical transition point $\lambda_w$} (c.f. $w_t \left(1 + \kappa^q + \kappa^k_{t^*}\right) > 0$). We see that query perturbations alone are insufficient, as they result in $\lambda_w=1$. Lemma: consider the special case when all keys have similar length $k_t \approx const$, the attended token has $\theta_{t^*}\approx0$, the keys are far-from-orthogonal s.t. $\theta_t \ll 1$, and $\kappa^q\approx0$. Using $w_t \triangleq |q| |k_t| \cos\theta_t$, circuit collapse occurs when $\exists~ t \neq t^*$ for which:
    \begin{equation}
            \frac{1}{2}\theta_t^2 ~\lesssim~ \kappa^k_t - \kappa^k_{t^*}   ~~~~~~~~~~~ \mathrm{if}~ w_t \left(1  + \kappa^k_{t^*}\right) > 0 ~\text{, otherwise reverse}
    \label{eq: sparse circuit collapse result}
    \end{equation}
    i.e. stability requires either well-separated keys s.t. $\theta_t \gg 0$, or small perturbations $\kappa_t-\kappa^*_t \ll 1$.
\label{theorem: multiplicative stability: sparse}
\end{theorem}

\begin{theorem}
    Sensitivity of isotropic attention to multiplicative perturbations. Say $\epsilon^k = \kappa^k_t k_t$ with $\kappa^k_t\ll1$ where $\{\kappa_t\}$ have comparable amplitudes. Then
    \begin{equation}
        \epsilon^{\Delta x(k)} %~\xrightarrow[\epsilon^k_t\rightarrow0]{\mathrm{~~perturb~k~~}}%~ w~ \langle {\tilde v}_t \kappa^k_t \rangle_t 
        ~\approx~
        \begin{cases}
        0 ~&~ \text{if~$\kappa_t$~independent~of~${\tilde m}_t$,~by~symmetry} \\
        0 ~&~ \text{if~$\kappa_t\equiv\kappa$~for~constant~$\kappa$} \\
        0 ~&~ \text{if~$q=0$} \\
        w \langle {\tilde m}_t \kappa^k_t \rangle_t  ~&~ \text{otherwise}
        \end{cases}
    \end{equation}
\label{theorem: multiplicative stability: isotropic}
\end{theorem}
\end{mdframed}

%In summary, sparse attention is unstable with respect to a spontaneous phase transition induced when interference changes the ordering of $w_t$, particularly when keys are densely spaced. To protect against this, keys must either have a highly orthogonal substructure, or be highly separable from other semantics s.t. interference is negligible

%In summary, isotropic attention distributions are stable with respect to small interference on query vectors, are sensitive to interference on the key vectors only when it correlates strongly with $v_t$, and linearly transport interference on value vectors - but with a stabilising effect in the mean-field limit.

%%
%%  Section: Experimental results
%%
\section{Experimental results}
\label{sec: experimental results}

We now use experiments to empirically probe (i) the real-world embedding structure, and (ii) the sensitivity to artificial noise on the \{query, key, value\} $L_2$-norms. Whilst this does not directly observe real-world interference, it constrains the effect importance.

We consider a base-10 integer-addition task with a question-answer structure, and train for next-token prediction. We use a decoder architecture, common for state-of-the-art language models, with $10$ layers, per-character tokenisation, and begin \texttt{[} and end \texttt{]} tokens. In the output, we mask \texttt{*} tokens that precede the answer. For example, the first training sequence has input \textcolor{Maroon}{\texttt{[453+16+17-N846=1332}} and output \textcolor{Maroon}{\texttt{***************1332]}}. We compare two models that use \texttt{Pre-Norm} and \texttt{QKV-Norm} respectively. Appendices~\ref{appendix: experimental setup}-\ref{app: model variations} provide a full experimental setup and supplementary plots. %Code and log files are provided as a \texttt{zip} in supplementary materials. 
In this section we make all plots using an in-distribution test set that is expected to have some overlap with the training set, bounded at $\ll 20\%$.

We choose this task because it emphasises contextual reasoning in a small-scale setting, is configurable for complexity, and allows us to define meaningful out-of-distribution test sets. The \texttt{Pre-Norm} (\texttt{QKV-Norm}) model achieves an in-distribution per-token accuracy of $91.4\%$ ($91.0\%$), dropping to $87.5\%$ ($82.5\%$) when generalising out-of-distribution to intermediate complexity, and $66.7\%$ ($46.8\%$) for increased complexity. Statistical uncertainties are below $0.1\%$. The in-distribution performance is comparable, but \texttt{QKV-Norm} generalises worse in this task, implying it has learned less task-appropriate solutions.  Appendix~\ref{appendix: Pre vs QKV norm} shows additional comparisons suggesting that the \texttt{Pre-Norm} and \texttt{QKV-Norm} models behave differently, supporting the observations of \cite{DBLP:journals/corr/abs-2010-04245,wortsman2023smallscale,dehghani2023scaling}.

%\clearpage
\textbf{Embedding structure}

Our theory predicts that \texttt{Pre-Norm} attention is stable with respect to information in non-attended subspaces if all input embeddings have similar $L_2$-norms, whereas \texttt{QKV-Norm} imposes no norm constraint. We seek to experimentally bound the degree to which this structure is learned in practice.

\begin{figure}[h]
    \centering
    \includegraphics[width=\textwidth]{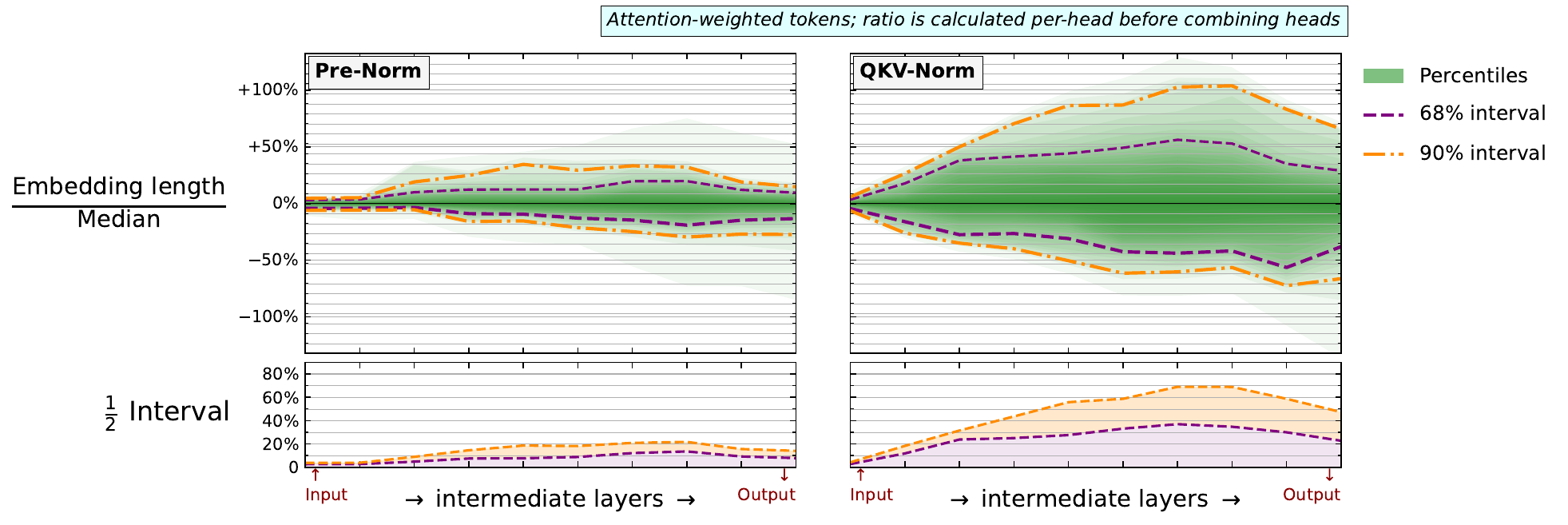}
    \caption{Spread of embedding $L_2$-norms experienced by attention heads at increasing model depth, excluding the \texttt{[} token. For \texttt{Pre-Norm}, 90\% of the spread is observed within an interval of $\pm20\%$. Supplementary Figure~\ref{fig: embedding L2 norms prenorm head} shows the distributions used to make this plot. Supplementary Figures~\ref{fig: embedding spread: model variations MID}-\ref{fig: embedding spread: model variations END} replicate the analysis for two model variations.}
    \label{fig: embedding spread}
\end{figure}

We do this by plotting \textit{the spread of norms with respect to their median}. A confounding effect is that the norms may differ for (i) embeddings attended to by different heads, (ii) the same head acting in different contexts, and (iii) embeddings that are never attended. We therefore measure the ratio \textit{per-head}, and weight each embedding by its assigned attention. We remove the begin-sequence token from consideration. Figure~\ref{fig: embedding spread} shows the resulting spread for all attention layers. On the LHS, we see that 90\% of the distribution is contained within an interval of $\pm20\%$ when using \texttt{Pre-Norm}. On the RHS, we see that \texttt{QKV-Norm} allows a much wider spread. This is consistent with our theory, and experimentally bounds the representation effect on \texttt{Pre-Norm} to $\lesssim 20\%$ in this model. Supplementary Figures~\ref{fig: embedding spread: model variations MID}-\ref{fig: embedding spread: model variations END} show consistent results for two model variations, although we note that \texttt{Pre-Norm} and \texttt{QKV-Norm} are more comparable for the variation labelled \texttt{Alternate}.

\textbf{Model stability with simulated interference}

\begin{figure}[t]
    \centering
    \includegraphics[page=3, width=0.9\textwidth, clip=True, trim=0 8.cm 0 1cm]{figures/Paper_diagrams.pdf}
    \caption{\textbf{Left:} evolution of per-token accuracy as we increase noise on the $L_2$-norms of $\{q,k_t,m_t\}$. A $\gtrsim 10\%$ drop in performance is observed when $1\%$ noise is applied to all layers. \textbf{Right:} applying noise only to $\{q,k_t\}$, we see that non-sparse attention drives the drop at small noise, whereas the sparse case is stable. This is consistent with Theorems~\ref{theorem: stability: sparse}-\ref{theorem: stability: isotropic}, but this interpretation is confounded by the relative importance of non-sparse distributions caused by frequency and depth-dependence.}
    \label{fig: noise sensitivity}
    \vspace{0.1cm}
    \includegraphics[width=0.9\textwidth, clip=True, trim=0 0 0 0]{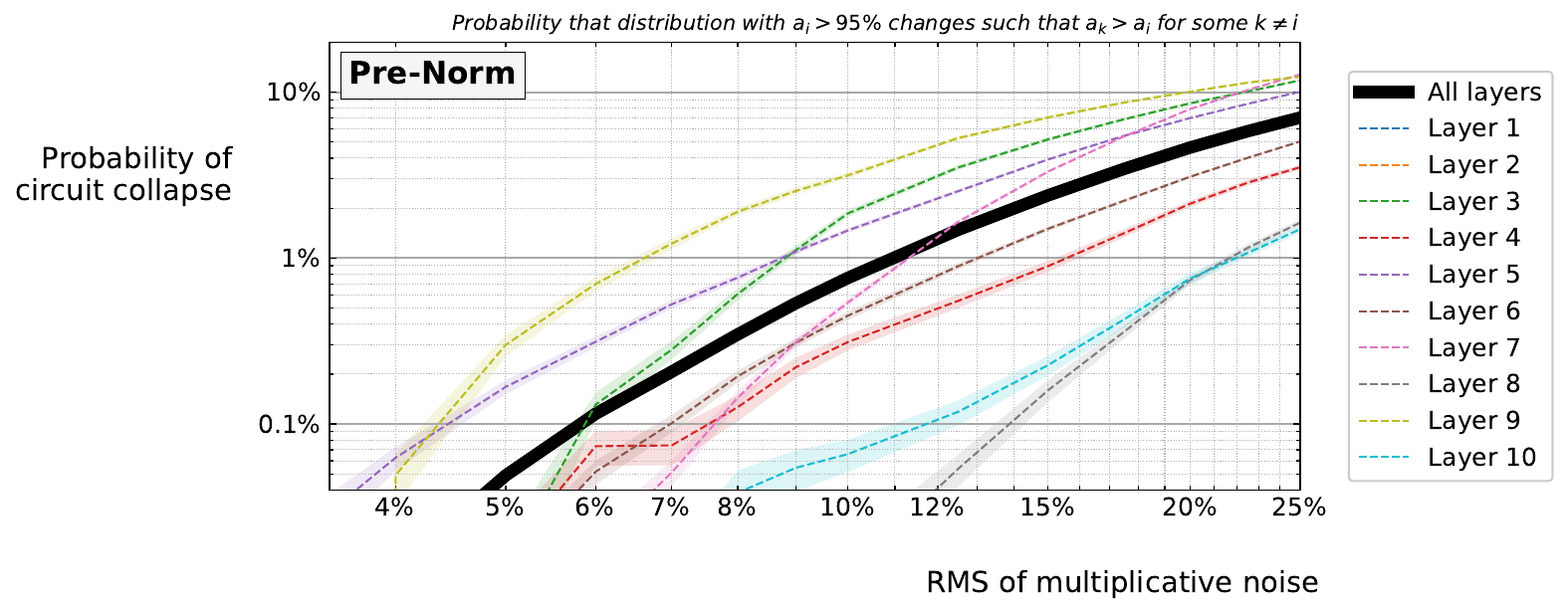}
    \caption{Probability of \textit{circuit collapse} vs increasing noise. This observes the effect predicted in Section~\ref{sec: theory: circuit stability}, and measures that $1\%$ of sparse distributions collapse at a noise level of $11\%$.}
    \label{fig: circuit collapse}
    %\vspace*{-7mm}
    \vspace*{-5mm}
\end{figure}

Section~\ref{sec: theory: circuit stability} theoretically modelled the semantic interference induced by \texttt{Pre-Norm} as a random perturbation on the norms of $\{q,k_t,m_t\}$. To estimate the real-world sensitivity to such an effect, we artificially introduce uncorrelated uniform noise onto these norms inside our trained \texttt{Pre-Norm} model. Even though Gaussian noise is expected in the large-$N_s$ limit, we use uniform noise to avoid outliers. Figure~\ref{fig: noise sensitivity} shows the evolution of in-distribution per-token accuracy with increasing RMS. On the LHS, we see that performance falls by $\gtrsim10\%$ at only a $1\%$ noise level. We also show the trend excluding the end-sequence token, which contributes a significant fraction of the metric. On the RHS, we introduce $\{q,k_t\}$ noise only to sparse heads (when $\max_t a_t \geq 95\%$) and non-sparse heads (when $\max_t a_t < 70\%$). We see that the model is stable with respect to $\%$-scale noise on sparse-attention, and this regime is dominated by the non-sparse case. 

Figure~\ref{fig: noise sensitivity} (right) is consistent with the stability predictions of Theorems~\ref{theorem: stability: sparse}-\ref{theorem: stability: isotropic}. However, it may also be explained if non-sparse distributions are simply more important to the model. This could be caused by non-sparse distributions being more common, as well as depth-dependence. This is because artificial noise is applied to all layers during the forward pass, therefore later layers are perturbed by both the noise component \textit{and} the shifting of their inputs due to previous layers, which is expected to compound with depth. We are interested in capturing this effect, however it may increase the importance of early layers. See Figure~\ref{fig: attention map prenorm} for a visualisation of the observed attention maps.
%It may be instructive to analyse Figure~\ref{fig: noise sensitivity}(left and right) on a per-layer basis.

\vspace{0.2cm}
\textbf{Circuit collapse}

Figure~\ref{fig: circuit collapse} shows the probability that our artificial noise causes the circuit collapse phenomenon as defined in section~\ref{sec: theory: circuit stability}. In this experiment, we add noise to every layer independently. This prevents the confounding effect of shifting inputs due to noise in previous layers. We observe that $1\%$ of sparse attention distributions collapse when they experience noise at a level of $11\%$. This reduces to $7.5\%$ and $5.5\%$ for the two model variations shown in Appendix~\ref{app: model variations}.

\section{Summary \& Outlook}
\label{sec: summary}

We have presented the idea that transformer \texttt{Pre-Norm} can cause interference between independent subspaces of the latent embeddings, a feature used by some real-world transformer circuits. Theoretically, we found this can only be avoided when using an embedding structure of \textit{orthogonal spheres}. By contrast, the \texttt{QKV-Norm} architecture requires only linearly independent subspaces. We predict that sparse attention is stable with respect to interference, until a certain threshold of noise is reached, at which point it undergoes a phase transition called \textit{circuit collapse}. 

Empirically, we observe that the $L_2$-norms of attended embeddings are contained within a spread of $\pm20\%$ for \texttt{Pre-Norm} (with 90\% coverage), whilst \texttt{QKV-Norm} creates a wider spread. We simulate interference by introducing artificial noise onto the $L_2$-norms of $\{q, k_t, v_t\}$ in our trained \texttt{Pre-Norm} model, observing that $1\%$ of sparse distributions collapse at a noise level of $11\%$. We observe that per-token accuracy degrades by $\mathcal{O}(10\%)$ when norms are simultaneously perturbed by noise of 1\% in all layers, but is stable to \%-scale noise in only sparse distributions. This may be attributed to either the predicted stability of sparse attention, or to a difference in the importance of sparse vs non-sparse heads induced by frequency and depth-dependence. More work is needed to disentangle these.

This work contributes a theoretical hypothesis of model behaviour, and empirically constrains the effect size without full model reverse-engineering. We have made predictions on representation structure, interference, and circuit collapse that practitioners may search for in their own models.

\section{Limitations}
\label{sec: limitations}

We have not directly observed subspace independence or interference, and further work is required to establish their importance in real-world models. Experimentally, we simulate interference as being independent and similar in amplitude across heads and layers, however it is possible that it is correlated and depth-dependent. Whilst our stability experiments demonstrate that the model is more stable with respect to noise in sparse than non-sparse distributions, we have not shown whether this is due to the inherent stability of the attention distribution predicted by our theory, or the relative importance of sparse vs non-sparse distributions to the model. We show experimental results for a small model on a targeted task (with model variations in Appendix~\ref{app: model variations}); further work is needed to study the behaviour of larger models and different corpora.

%It is not shown whether the drop in model performance is driven by perturbations in early layers, where we expect  
%stability experiments do not disentangle the contributions of simulated interference and compounding errors, in which the perturbation of early layers also affects the inputs to later ones. 

%%
%%  Acknowledgements
%%
\begin{ack}
This work is supported by UKRI Turing AI World-Leading Researcher Fellowship (EP/W002973/1). This work was partially funded by the Swiss National Science Foundation (SNSF) project NeuMath (\href{https://data.snf.ch/grants/grant/204617}{200021\_204617}), by the EPSRC grant EP/T026995/1, ``EnnCore: End-to-End Conceptual Guarding of Neural Architectures'' under Security for all in an AI enabled society, by the CRUK National Biomarker Centre, and supported by the Manchester Experimental Cancer Medicine Centre and the NIHR Manchester Biomedical Research Centre.
\end{ack}

%%
%%  References
%%
%\clearpage
\bibliographystyle{unsrt} % We choose the "plain" reference style
\bibliography{bibliography}

\begin{thebibliography}{10}

\bibitem{DBLP:journals/corr/VaswaniSPUJGKP17}
Vaswani et~al.
\newblock Attention is all you need.
\newblock In {\em Advances in Neural Information Processing Systems},
  volume~30. Curran Associates, Inc., 2017.

\bibitem{bubeck2023sparks}
Bubeck et~al.
\newblock Sparks of artificial general intelligence: Early experiments with
  gpt-4, 2023.

\bibitem{touvron2023llama}
Touvron et~al.
\newblock Llama: Open and efficient foundation language models, 2023.

\bibitem{AlphaFold3}
Abramson et~al.
\newblock Accurate structure prediction of biomolecular interactions with
  alphafold 3.
\newblock {\em Nature}, 2024.

\bibitem{doi:10.1126/science.abj8754}
Baek et~al.
\newblock Accurate prediction of protein structures and interactions using a
  three-track neural network.
\newblock {\em Science}, 373(6557):871--876, 2021.

\bibitem{10.7554/eLife.82819}
Chandra et~al.
\newblock Transformer-based deep learning for predicting protein properties in
  the life sciences.
\newblock {\em eLife}, 12:e82819, jan 2023.

\bibitem{cai2024transforming}
Cai et~al.
\newblock Transforming the bootstrap: Using transformers to compute scattering
  amplitudes in planar n = 4 super yang-mills theory, 2024.

\bibitem{olsson2022context}
Olsson et~al.
\newblock In-context learning and induction heads.
\newblock {\em Transformer Circuits Thread}, 2022.
\newblock
  https://transformer-circuits.pub/2022/in-context-learning-and-induction-heads/index.html.

\bibitem{elhage2021mathematical}
Elhage et~al.
\newblock A mathematical framework for transformer circuits.
\newblock {\em Transformer Circuits Thread}, 2021.
\newblock https://transformer-circuits.pub/2021/framework/index.html.

\bibitem{wang2022interpretability}
Wang et~al.
\newblock Interpretability in the wild: a circuit for indirect object
  identification in {GPT}-2 small.
\newblock In {\em The Eleventh International Conference on Learning
  Representations}, 2023.

\bibitem{ferrando2024primer}
Ferrando et~al.
\newblock A primer on the inner workings of transformer-based language models,
  2024.

\bibitem{meng2022locating}
Meng et~al.
\newblock Locating and editing factual associations in {GPT}.
\newblock In Alice~H. Oh, Alekh Agarwal, Danielle Belgrave, and Kyunghyun Cho,
  editors, {\em Advances in Neural Information Processing Systems}, 2022.

\bibitem{DBLP:journals/corr/abs-2002-12327}
Rogers et~al.
\newblock A primer in bertology: What we know about how bert works.
\newblock {\em Transactions of the Association for Computational Linguistics},
  8:842--866, 12 2020.

\bibitem{liu2023transformers}
Liu et~al.
\newblock Transformers learn shortcuts to automata.
\newblock In {\em The Eleventh International Conference on Learning
  Representations}, 2023.

\bibitem{goldowskydill2023localizing}
Goldowsky-Dill et~al.
\newblock Localizing model behavior with path patching, 2023.

\bibitem{cammarata2020thread:}
Cammarata et~al.
\newblock Thread: Circuits.
\newblock {\em Distill}, 2020.
\newblock https://distill.pub/2020/circuits.

\bibitem{DBLP:journals/corr/abs-2002-04745}
Xiong et~al.
\newblock On layer normalization in the transformer architecture.
\newblock In {\em Proceedings of the 37th International Conference on Machine
  Learning}, ICML'20. JMLR.org, 2020.

\bibitem{DBLP:journals/corr/abs-2010-04245}
Henry et~al.
\newblock Query-key normalization for transformers.
\newblock {\em CoRR}, abs/2010.04245, 2020.

\bibitem{wortsman2023smallscale}
Wortsman et~al.
\newblock Small-scale proxies for large-scale transformer training
  instabilities.
\newblock In {\em The Twelfth International Conference on Learning
  Representations}, 2024.

\bibitem{dehghani2023scaling}
Dehghani et~al.
\newblock Scaling vision transformers to 22 billion parameters.
\newblock In {\em Proceedings of the 40th International Conference on Machine
  Learning}, volume 202 of {\em Proceedings of Machine Learning Research},
  pages 7480--7512. PMLR, 23--29 Jul 2023.

\bibitem{NEURIPS2019_1e8a1942}
Biao Zhang and Rico Sennrich.
\newblock Root mean square layer normalization.
\newblock In Wallach et~al, editor, {\em Advances in Neural Information
  Processing Systems}, volume~32. Curran Associates, Inc., 2019.

\bibitem{DBLP:journals/corr/BaKH16}
Ba~et~al.
\newblock Layer normalization.
\newblock {\em CoRR}, abs/1607.06450, 2016.

\bibitem{elhage2022superposition}
Elhage et~al.
\newblock Toy models of superposition.
\newblock {\em Transformer Circuits Thread}, 2022.

\bibitem{stolfo2023mechanistic}
Stolfo et~al.
\newblock A mechanistic interpretation of arithmetic reasoning in language
  models using causal mediation analysis, 2023.

\bibitem{GoldowskyDill2023LocalizingMB}
Goldowsky-Dill et~al.
\newblock Localizing model behavior with path patching.
\newblock {\em ArXiv}, abs/2304.05969, 2023.

\bibitem{Ferrando2024InformationFR}
Javier Ferrando and Elena Voita.
\newblock Information flow routes: Automatically interpreting language models
  at scale.
\newblock {\em ArXiv}, abs/2403.00824, 2024.

\bibitem{singh2023the}
Singh et~al.
\newblock The transient nature of emergent in-context learning in transformers.
\newblock In {\em Thirty-seventh Conference on Neural Information Processing
  Systems}, 2023.

\bibitem{singh2024needs}
Singh et~al.
\newblock What needs to go right for an induction head? a mechanistic study of
  in-context learning circuits and their formation, 2024.

\bibitem{Devlin2019BERTPO}
Devlin et~al.
\newblock Bert: Pre-training of deep bidirectional transformers for language
  understanding.
\newblock In {\em North American Chapter of the Association for Computational
  Linguistics}, 2019.

\bibitem{nguyen-salazar-2019-transformers}
Nguyen et~al.
\newblock Transformers without tears: Improving the normalization of
  self-attention.
\newblock In {\em Proceedings of the 16th International Conference on Spoken
  Language Translation}, Hong Kong, November 2-3 2019. Association for
  Computational Linguistics.

\bibitem{nguyen-chiang-2018-improving}
Toan Nguyen and David Chiang.
\newblock Improving lexical choice in neural machine translation.
\newblock In {\em Proceedings of the 2018 Conference of the North {A}merican
  Chapter of the Association for Computational Linguistics: Human Language
  Technologies, Volume 1 (Long Papers)}, pages 334--343, New Orleans,
  Louisiana, June 2018. Association for Computational Linguistics.

\bibitem{shleifer2021normformer}
Shleifer et~al.
\newblock Normformer: Improved transformer pretraining with extra
  normalization, 2021.

\bibitem{NEURIPS2019_2f4fe03d}
Xu~et~al.
\newblock Understanding and improving layer normalization.
\newblock In H.~Wallach, H.~Larochelle, A.~Beygelzimer, F.~d\textquotesingle
  Alch\'{e}-Buc, E.~Fox, and R.~Garnett, editors, {\em Advances in Neural
  Information Processing Systems}, volume~32. Curran Associates, Inc., 2019.

\bibitem{kobayashi-etal-2021-incorporating}
Kobayashi et~al.
\newblock {I}ncorporating {R}esidual and {N}ormalization {L}ayers into
  {A}nalysis of {M}asked {L}anguage {M}odels.
\newblock In {\em Proceedings of the 2021 Conference on Empirical Methods in
  Natural Language Processing}, pages 4547--4568, Online and Punta Cana,
  Dominican Republic, November 2021. Association for Computational Linguistics.

\bibitem{brody2023expressivity}
Brody et~al.
\newblock On the expressivity role of layernorm in transformers’ attention.
\newblock pages 14211--14221, 01 2023.

\bibitem{Molina2023TravelingWA}
Raul Molina.
\newblock Traveling words: A geometric interpretation of transformers.
\newblock {\em ArXiv}, abs/2309.07315, 2023.

\bibitem{Dong2021AttentionIN}
Dong et~al.
\newblock Attention is not all you need: Pure attention loses rank doubly
  exponentially with depth.
\newblock {\em PMLR}, 139, 2021.

\bibitem{wu2024role}
Wu~et~al.
\newblock On the role of attention masks and layernorm in transformers, 2024.

\bibitem{wang2024understanding}
Wang et~al.
\newblock Towards understanding how transformer perform multi-step reasoning
  with matching operation, 2024.

\bibitem{boixadsera2024when}
Boix-Adser{\`a} et~al.
\newblock When can transformers reason with abstract symbols?
\newblock In {\em The Twelfth International Conference on Learning
  Representations}, 2024.

\bibitem{csordas2022the}
Csord{\'a} et~al.
\newblock The neural data router: Adaptive control flow in transformers
  improves systematic generalization.
\newblock In {\em International Conference on Learning Representations}, 2022.

\bibitem{Lamb2021TransformersWC}
Lambd et~al.
\newblock Transformers with competitive ensembles of independent mechanisms.
\newblock {\em ArXiv}, abs/2103.00336, 2021.

\bibitem{word2vec1}
Mikolov et~al.
\newblock Efficient estimation of word representations in vector space.
\newblock pages 1--12, 01 2013.

\bibitem{NIPS2013_9aa42b31}
Mikolov et~al.
\newblock Distributed representations of words and phrases and their
  compositionality.
\newblock In C.J. Burges, L.~Bottou, M.~Welling, Z.~Ghahramani, and K.Q.
  Weinberger, editors, {\em Advances in Neural Information Processing Systems},
  volume~26. Curran Associates, Inc., 2013.

\bibitem{Pennington2014GloVeGV}
Jeffrey Pennington, Richard Socher, and Christopher~D. Manning.
\newblock Glove: Global vectors for word representation.
\newblock In {\em Conference on Empirical Methods in Natural Language
  Processing}, 2014.

\bibitem{liguistic_regularities}
Mikolov et~al.
\newblock Linguistic regularities in continuous space word representations.
\newblock {\em Proceedings of NAACL-HLT}, pages 746--751, 01 2013.

\bibitem{Park2023TheLR}
Park et~al.
\newblock The linear representation hypothesis and the geometry of large
  language models.
\newblock {\em ArXiv}, abs/2311.03658, 2023.

\bibitem{jiang2024origins}
Jiang et~al.
\newblock On the origins of linear representations in large language models,
  2024.

\bibitem{semantic_subspace_probing}
Dmitry Nikolaev and Sebastian Padó.
\newblock Investigating semantic subspaces of transformer sentence embeddings
  through linear structural probing.
\newblock 10 2023.

\bibitem{belinkov-2022-probing}
Yonatan Belinkov.
\newblock Probing classifiers: Promises, shortcomings, and advances.
\newblock {\em Computational Linguistics}, 48(1):207--219, March 2022.

\bibitem{Geiger2023FindingAB}
Geiger et~al.
\newblock Finding alignments between interpretable causal variables and
  distributed neural representations.
\newblock {\em ArXiv}, abs/2303.02536, 2023.

\bibitem{superposition1}
Sanjeev et~al.
\newblock Linear algebraic structure of word senses, with applications to
  polysemy.
\newblock {\em Transactions of the Association for Computational Linguistics},
  6, 01 2016.

\bibitem{5596640}
Tripathi et~al.
\newblock Semantic subspace learning with conditional significance vectors.
\newblock In {\em The 2010 International Joint Conference on Neural Networks
  (IJCNN)}, pages 1--8, 2010.

\bibitem{Coenen2019VisualizingAM}
Coenen et~al.
\newblock Visualizing and measuring the geometry of bert.
\newblock In {\em Neural Information Processing Systems}, 2019.

\bibitem{Hewitt2019ASP}
Hewitt et~al.
\newblock A structural probe for finding syntax in word representations.
\newblock In {\em North American Chapter of the Association for Computational
  Linguistics}, 2019.

\bibitem{ethayarajh-2019-contextual}
Kawin Ethayarajh.
\newblock How contextual are contextualized word representations? {C}omparing
  the geometry of {BERT}, {ELM}o, and {GPT}-2 embeddings.
\newblock In {\em Proceedings of the 2019 Conference on Empirical Methods in
  Natural Language Processing and the 9th International Joint Conference on
  Natural Language Processing (EMNLP-IJCNLP)}, pages 55--65, Hong Kong, China,
  November 2019. Association for Computational Linguistics.

\bibitem{song2024uncovering}
Song et~al.
\newblock Uncovering hidden geometry in transformers via disentangling position
  and context, 2024.

\bibitem{muppet2022}
Mickus et~al.
\newblock How to dissect a muppet: The structure of transformer embedding
  spaces.
\newblock {\em Transactions of the Association for Computational Linguistics},
  10:981--996, 09 2022.

\bibitem{hernandez2024linearity}
Hernandez et~al.
\newblock Linearity of relation decoding in transformer language models.
\newblock In {\em The Twelfth International Conference on Learning
  Representations}, 2024.

\bibitem{chi-etal-2020-finding}
Chi et~al.
\newblock Finding universal grammatical relations in multilingual {BERT}.
\newblock In {\em Proceedings of the 58th Annual Meeting of the Association for
  Computational Linguistics}, pages 5564--5577, Online, July 2020. Association
  for Computational Linguistics.

\bibitem{cai2021isotropy}
Cai et~al.
\newblock Isotropy in the contextual embedding space: Clusters and manifolds.
\newblock In {\em International Conference on Learning Representations}, 2021.

\bibitem{Hernandez2021TheLL}
Evan Hernandez and Jacob Andreas.
\newblock The low-dimensional linear geometry of contextualized word
  representations.
\newblock In {\em Conference on Computational Natural Language Learning}, 2021.

\bibitem{tensorflow2015-whitepaper}
Mart\'{i}n~Abadi et~al.
\newblock {TensorFlow}: Large-scale machine learning on heterogeneous systems,
  2015.
\newblock Software available from tensorflow.org.

\bibitem{harris2020array}
Harris et~al.
\newblock Array programming with {NumPy}.
\newblock {\em Nature}, 585(7825):357--362, September 2020.

\bibitem{agarap2019deep}
Abien~Fred Agarap.
\newblock Deep learning using rectified linear units (relu), 2019.

\bibitem{pmlr-v9-glorot10a}
Xavier Glorot and Yoshua Bengio.
\newblock Understanding the difficulty of training deep feedforward neural
  networks.
\newblock In Yee~Whye Teh and Mike Titterington, editors, {\em Proceedings of
  the Thirteenth International Conference on Artificial Intelligence and
  Statistics}, volume~9 of {\em Proceedings of Machine Learning Research},
  pages 249--256, Chia Laguna Resort, Sardinia, Italy, 13--15 May 2010. PMLR.

\bibitem{chollet2015keras}
Fran\c{c}ois Chollet et~al.
\newblock Keras.
\newblock \url{https://keras.io}, 2015.

\bibitem{loshchilov2019decoupled}
Ilya Loshchilov and Frank Hutter.
\newblock Decoupled weight decay regularization, 2019.

\end{thebibliography}

%%%%%%%%%%%%%%%%%%%%%%%%%%%%%%%%%%%%%%%%%%%%%%%%%%%%%%%%%%%%

%\clearpage
%\input{checklist}

%%%%%%%%%%%%%%%%%%%%%%%%%%%%%%%%%%%%%%%%%%%%%%%%%%%%%%%%%%%%

\appendix

%%
%%  Appendix:  Experimental setup
%%
\clearpage
\section{Experimental setup}
\label{appendix: experimental setup}

\newcommand{\data}[1]{\textcolor{Maroon}{\texttt{{#1}}}}

\subsection{Data}

%The anonymised code used to produce these results is provided in the supplementary \texttt{zip} file. 
Due to the task nature, we do not require static datasets and so generate both train and test data on-the-fly. This alleviates storage and memory concerns for long training runs in which a static dataset would have to be large. Datasets are reproducible through configuration of the environment and global random seed, which is used to manually control the random seeds of \texttt{Python}, \texttt{TensorFlow} \cite{tensorflow2015-whitepaper}, and \texttt{NumPy} \cite{harris2020array}. This also reproduces the model initialisation. %Random seeds are provided for the models shown. %The \texttt{zip} file includes detailed log files of training runs to help debug the replication of experiments.

\subsection{Task specification}

We consider an integer addition task, where each character is a base-10 numeral \data{0}-\data{9}, mathematical operator \{\data{+}, \data{-}, \data{=}, \data{N}\}, or special character \{\data{[}, \data{]}, \data{*}\}. The \data{N} operator signifies that the following integer is \textit{negative}, and is used to avoid overloading notation with the \data{-} operator, which means \textit{minus}. The special characters are the begin-sequence token \data{[}, end-sequence token \data{]}, and mask character \data{*}. Input sequences in the same batch are right-padded with mask tokens to the same length, which do not contribute to the model. Characters that are masked in the output do not contribute to the evaluation metrics. We tokenise per-character so the model does not need to disambiguate different representations for identical patterns (e.g. if the number \textcolor{Maroon}{\texttt{112}} is tokenised as [11,2], and \textcolor{Maroon}{212} is tokenised as [2,12], then the pattern \textcolor{Maroon}{\texttt{12}} has a context-dependent representation). The token dictionary has a length of 17. 

For a decoder architecture, the model is a sequence-sequence transformer and each datapoint has a question-answer structure separated by the \data{=} character, e.g. the first datapoint is:
\begin{equation}
    \data{[453+16+17-N846=1332}  ~~~~\rightarrow~~~~ \data{***************1332]}
\end{equation}
The model must therefore predict the numerical outputs and the \data{]} token. For an encoder-decoder architecture, the encoder input is the \textit{question} and the decoder performs next-token prediction over the \textit{answer}, e.g.
\begin{equation}
    \data{[453+16+17-N846]}~\text{~(encoder)},~~\data{[1332}~\text{~(decoder)}  ~~~~\rightarrow~~~~ \data{1332]}~\text{~(decoder)}
\end{equation}

\begin{figure}[!]
\centering
\includegraphics[width=0.8\textwidth]{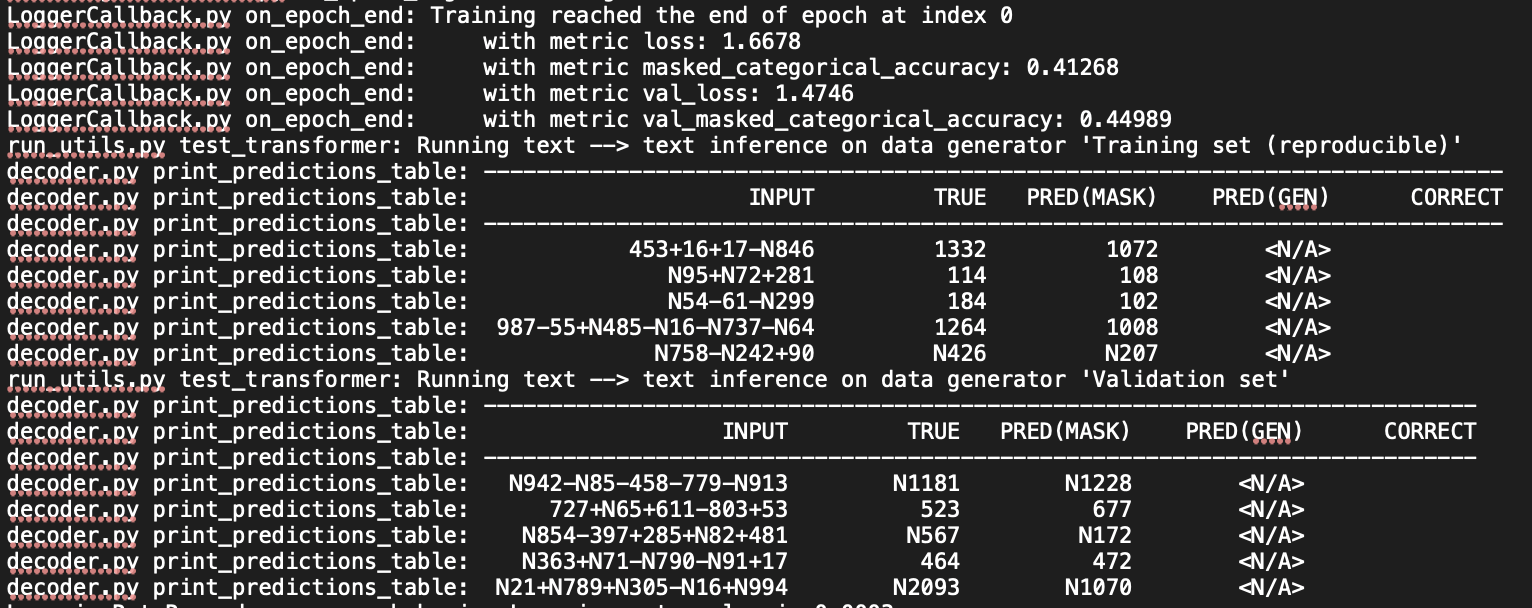}
\caption{\texttt{Baseline} \texttt{Pre-Norm} model predictions after 1 training epoch.}
\label{fig: screenshot model begin}
\vspace{0.5cm}
\includegraphics[width=0.8\textwidth]{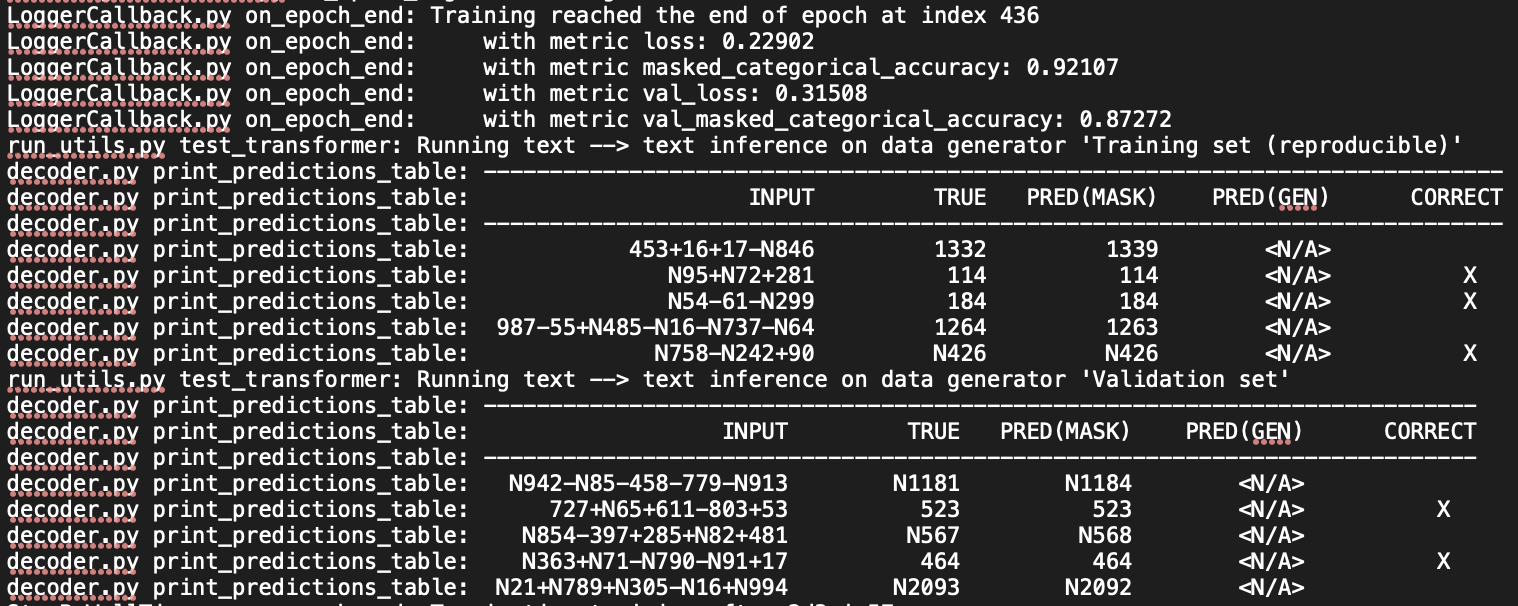}
\caption{\texttt{Baseline} \texttt{Pre-Norm} model predictions after training.}
\label{fig: screenshot model end}
\vspace{0.5cm}
\includegraphics[width=0.8\textwidth]{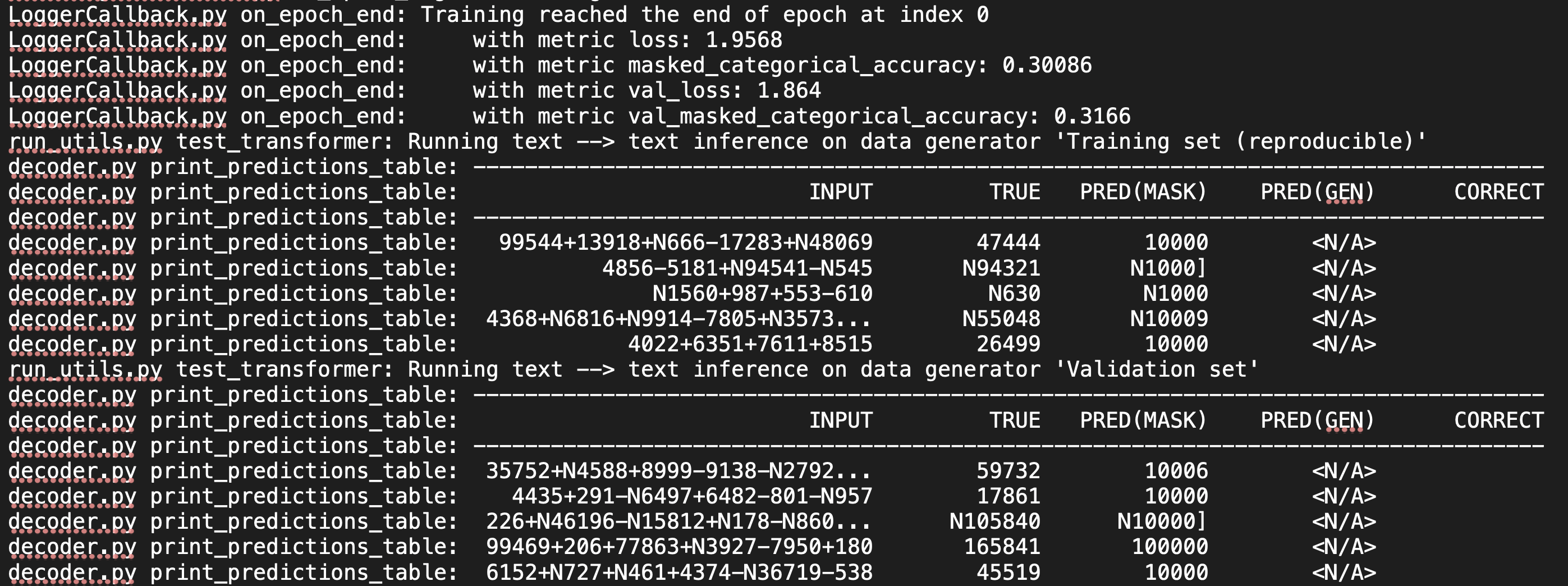}
\caption{\texttt{Large} \texttt{Pre-Norm} model predictions after 1 training epoch.}
\label{fig: screenshot large model begin}
\vspace{0.5cm}
\includegraphics[width=0.8\textwidth]{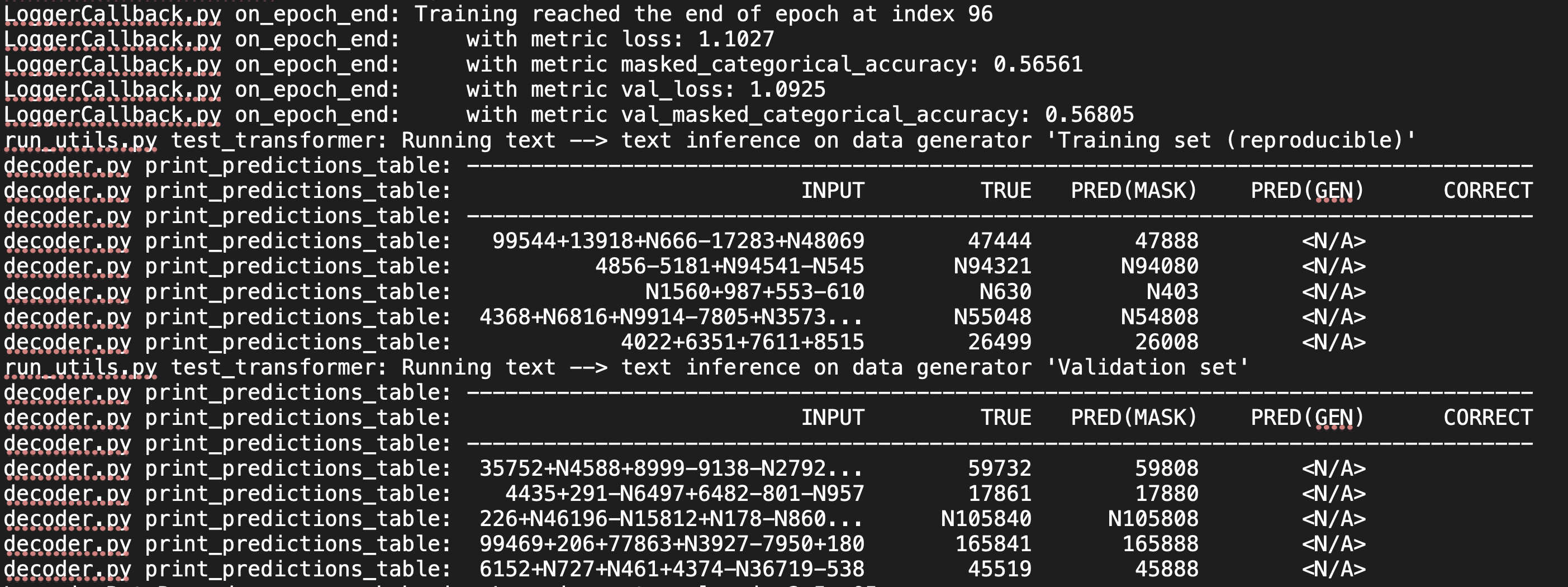}
\caption{\texttt{Large} \texttt{Pre-Norm} model predictions after training.}
\label{fig: screenshot large model end}
\end{figure}

To help visualise the task, Figure~\ref{fig: screenshot model begin} shows the predictions of the baseline \texttt{Pre-Norm} model after 1 epoch. Figure~\ref{fig: screenshot model end} shows the fully-trained model, to help visualise the attainable in-distribution performance. The final epoch per-token accuracy is logged as 92\%; the model sometimes correctly predicts all digits of the answer, otherwise it appears to be correct in the leading digits. Figures~\ref{fig: screenshot large model begin}-\ref{fig: screenshot large model end} repeat this for the \texttt{Large} model variation, which acts on a more complex task setting and achieves a lower per-token accuracy of 57\%. Once again, the correctly predicted tokens appear to be driven by the leading digits.

\subsection{Data-generation process}

One advantage of this task is the ability to modulate its complexity. Each dataset is defined by two hyperparameters:

\begin{tabular}{cll}
    Dataset parameter    &   Example   &   Description   \\
    \hline
    $N$   &   [3, 4, 6]    &   The allowed number of integers per-sequence  \\
    $L$   &   [2, 3]   &   The allowed number of digits per-integer  \\
\end{tabular}

Each datapoint is generated by uniformly sampling a value of $N$, then uniformly sampling a value of $L$ for each integer. This ensures that examples are not simply dominated by integers with the maximum number of digits. Each integer is uniformly sampled from all positive and negative integers with that length. Between each integer, an operator is uniformly sampled from the list $[+, -]$. For example, the datapoint 
\begin{equation}
    \data{[453+16+17-N846=1332}  ~~~~\rightarrow~~~~ \data{***************1332]}
\end{equation}
was generated by sampling a value of $N=4$ to determine that the sum contains four integers, then sampling four values of $L=[3,2,2,3]$ to determine their lengths, then sampling the numbers $[453, 16, 17, N846]$ and operators $[+, +, -]$. The inclusion of subtraction, addition of negative numbers, and double-negatives is intended to emphasise solutions that parse the \textit{context} of each digit within the sum.

\subsection{Train/test specifications}

Table~\ref{table: dataset specifications main} shows the $N$ and $L$ parameters used for the \texttt{Baseline} and \texttt{Alternate} experiments. We also show the number of datapoints, and the per-datapoint sampling probability. This is a range, with higher probabilites for the simpler sums. Table~\ref{table: dataset specifications large} shows the task specification for the \texttt{Large} model variation, which is trained on a more complex setting. We also perform a scan over model size and learning rate to compare the training stability of \texttt{Pre-Norm} and \texttt{QKV-Norm}. These experiments were performed using an earlier problem configuration shown in Table~\ref{table: dataset specifications scan}.

\begin{table}[h]
\centering
\begin{tabular}{rllll}
    Dataset    &   $N$   &   $L$   &   Num datapoints   &   Datapoint probability   \\
    \hline
    Train                     &   $[3, 4, 6]$   &   $[2, 3]$   &   110M, acc=90\% @ 40M   &   $2\times 10^{-9} $ to $5\times10^{-24}$   \\
    Validation     &   $[5]$         &   $[2, 3]$   &   6.4k   &   $1\times 10^{-14}$ to $1\times 10^{-19}$   \\
    \hline
    In-distribution         &   $[3, 4, 6]$   &   $[2, 3]$   &   128k   &   $2\times 10^{-9}$ to $5\times10^{-24}$   \\
    OOD (interpolation)     &   $[5]$         &   $[2, 3]$   &   128k   &   $1\times 10^{-14}$ to $1\times 10^{-19}$   \\
    OOD (extrapolation)     &   $[7, 8, 9]$   &   $[2, 3]$   &   128k   &   $7\times 10^{-21}$ to $1\times 10^{-35}$   \\
\end{tabular}
\vspace{0.3cm}
\caption{Dataset configurations used for \texttt{Baseline} and \texttt{Alternate} results.}
\label{table: dataset specifications main}
\vspace{0.5cm}
\begin{tabular}{rllll}
    Dataset    &   $N$   &   $L$   &   Num datapoints   &   Datapoint probability   \\
    \hline
    Train                     &   $[4, 5, 7, 8]$   &   $[3, 4, 5]$   &   25M   &   $4\times 10^{-17} $ to $3\times10^{-49}$   \\
    Validation     &   $[6]$         &   $[3, 4, 5]$   &   6.4k   &   $2\times 10^{-24}$ to $2\times 10^{-36}$   \\
    \hline
    In-distribution         &   $[4, 5, 7, 8]$   &   $[3, 4, 5]$   &   128k   &   $4\times 10^{-17} $ to $3\times10^{-49}$   \\
    OOD (interpolation)     &   $[6]$         &   $[3, 4, 5]$   &   128k   &   $2\times 10^{-24}$ to $2\times 10^{-36}$   \\
    OOD (extrapolation)     &   $[9, 10, 11]$   &   $[3, 4, 5]$   &   128k   &   $3\times 10^{-37}$ to $3\times 10^{-67}$   \\
\end{tabular}
\vspace{0.3cm}
\caption{Dataset configurations used for \texttt{Large} results.}
\label{table: dataset specifications large}
\vspace{0.5cm}
\begin{tabular}{rlll}
    Dataset    &   $N$   &   $L$   &   Datapoint probability   \\
    \hline
    Train set                     &   $[3, 4, 5, 7]$   &   $[3, 4, 5]$   &   $4\times 10^{-13}$ to $3\times 10^{-43}$   \\
    In-distribution         &   $[3, 4, 5, 7]$   &   $[3, 4, 5]$   &   $4\times 10^{-13}$ to $3\times 10^{-43}$   \\
\end{tabular}
\vspace{0.3cm}
\caption{Dataset configurations used for training stability results (Figure~\ref{fig: training stability scan}).}
\label{table: dataset specifications scan}
\end{table}

We halt training according to wall time, which leads to a range of observed dataset sizes. Model convergence may also occur much earlier. We therefore show an order-of-magnitude estimate for the number of observed datapoints, as well as the point at which the baseline model reaches 90\% per-token accuracy (this represents almost-convergence, which is logged at 92.1\%). 

Note that our data-generation strategy does not ensure that training examples are exclusive (there may be repetitions), nor that the in-distribution test set does not contain overlap with training examples. The final column is therefore important, because it demonstrates that the highest per-datapoint sampling probability is $2\times10^{-9}$, whilst the model converges with $\mathcal{O}(10^7)$ datapoints and observes $\mathcal{O}(10^8)$ in total. Since the datapoint probability is $2\times10^{-9}$ for the simplest configurations and $5\times10^{-25}$ for the most complicated, this ensures that the in-distribution evaluation metric is dominated by novel examples. The validation set is only used for visual inspection of model behaviour during training, as in Figures~\ref{fig: screenshot model begin}-\ref{fig: screenshot large model end}.

\subsection{Model specification (main experiments)}

We use a decoder architecture, meaning that the dot-product self-attention layers are causally masked such that token $t$ can only attend to tokens $\leq t$. The model has the following structure:
\begin{equation*}
\begin{matrix} 
    \texttt{Embedding + positional encoding} \\
    \downarrow \\
    N_{layer} \times 
    \begin{bmatrix} 
    \texttt{Attention block} \\ 
    \downarrow \\
    \texttt{Feed-forward block~(ReLU)} \\
    \end{bmatrix} \\
    \downarrow \\
    \texttt{Multi-layer perceptron~(ReLU)} \\
    \downarrow \\
    \texttt{Predicted~logits} \\
\end{matrix}
\end{equation*}

\vspace{0.2cm}
\textbf{Embedding + positional encoding} ~ We initialise each token embedding as $x = x_{type} + x_{pos}$, where $x_{type}$ is a token embedding with $N_{emb}$ elements, and $x_{pos}$ use cyclic positional encodings of the same form as the original transformer architecture \cite{DBLP:journals/corr/VaswaniSPUJGKP17}, with $N_{freq}$ frequencies initialised as a base $e$ log-series between periods of $3$ and $1k$ tokens. For each sequence, all position indices are simultaneously offset by a random integer between $0$ and $50$. This augmentation is designed to encourage the use of \textit{relative} positions rather than absolute. The frequencies are then left as trainable parameters. The positional encodings contribute the first $2N_{freq}$ components of $x_{pos}$, and the remaining are set to $0$. This configuration guarantees that the token embeddings and positional encodings can be made orthogonal in the first layer, and $x_{pos}$ have constant $L_2$-norm, consistent with our theoretical structure.

\vspace{0.2cm}
\textbf{Attention block} ~ $N_{layer}$ is the number of residual blocks of our model, where our baseline is $N_{layer}=10$. The update is as formulated in section~\ref{sec: formulation}, where $H$ is the number of parallel attention heads per layer. Since the embeddings have length $N_{emb}$, we must have $N_x=N_{emb}$, whilst the latent dimension $N_{qkv}$ is configurable. Either the \texttt{Pre-Norm} or \texttt{QKV-Norm} strategy is used, as configured.

\vspace{0.2cm}
\textbf{Feed-forward block} ~ The feed-forward blocks update embeddings using the function $x \rightarrow x + FF(\texttt{LayerNorm}(x))$, where $FF$ is a dense network with one hidden layer of size $N_{ff}$. The network uses a \texttt{ReLU} \cite{agarap2019deep} activation function on the intermediate layer, followed by a linear projection back onto embedding space. To maintain consistency with other models, we apply \texttt{LayerNorm} at the input to $FF$. Both \texttt{LayerNorm} and $FF$ use bias parameters.

\vspace{0.2cm}
\textbf{Multi-layer perceptron} ~ The final embeddings $x$ are mapped onto token logits $y$ using the function $y = MLP(\texttt{LayerNorm}(x))$, where $MLP$ is a multi-layer perceptron with two hidden layers of size $N_{MLP}$ and ReLU activation. %If configured, the activations are followed by \texttt{LayerNorm} normalisation. 
The final layer is a linear projection onto the space of logits, which has length 17. For the training stability scan in Figure~\ref{fig: training stability scan}, the MLP has three hidden layers instead.

\vspace{0.2cm}
\textbf{Hyperparameters} ~ Table~\ref{table: model spec main} shows the hyperparameters used to configure the networks of the main experiments. Table~\ref{table: model spec stability} show the hyperparameters used for the training stability analysis. This experiment also uses encoder-decoder models, following the same setup as the original transformer architecture \cite{DBLP:journals/corr/VaswaniSPUJGKP17} and with the layer configurations listed here.

\begin{table}[h]
\centering
\begin{tabular}{rlllllllll}
    Model   &   $N_{freq}$   &   $N_{emb}$   &   $N_{layer}$   &   $H$   &   $N_{qkv}$   &   $N_{ff}$   &   $N_{MLP}$   &   seed   \\
\hline
    \texttt{Baseline}        &   32   &   512    &   10    &   12   &   64   &   512   &   2$\times$512   &   100   \\
    \texttt{Alternative}   &   32   &   512    &   8     &   12   &   64   &   512   &   2$\times$512   &   100   \\
    \texttt{Large}   &   32   &   1024   &   12    &   16   &   64   &   512   &   2$\times$512   &   100   \\
\end{tabular}
\vspace{0.2cm}
\caption{Model hyperparameters for main experiments (i.e. other than training stability). \texttt{Baseline} is used for the main results presented in section~\ref{sec: experimental results} (short). \texttt{Alternate} and \texttt{Large} are presented in appendix~\ref{app: model variations} to show reproducibility of observations.}
\label{table: model spec main}
\vspace{0.2cm}
\begin{tabular}{rlllllllll}
    Model   &   $N_{freq}$   &   $N_{emb}$   &   $N_{layer}$   &   $H$   &   $N_{qkv}$   &   $N_{ff}$   &   $N_{MLP}$   &   seed   \\
\hline
    \texttt{All}        &   16   &   -   &   -    &   12   &   -   &   512   &   3$\times$512   &   1,2   \\
\end{tabular}
\vspace{0.2cm}
\caption{Model hyperparameters for training stability experiments. Empty parameters are varied per-model and displayed in Figure~\ref{fig: training stability scan}.}
\label{table: model spec stability}
\end{table}

\vspace{0.2cm}
\textbf{Loss} ~ The loss function is \texttt{categorical cross entropy}, calculated from the output logits.

\subsection{Model initialisation}

We use a custom initialisation strategy to give control over the initial state of the model. In particular, we use \texttt{Checkpoint} layers to ensure that the initial states are comparable between \texttt{Pre-Norm} and \texttt{QKV-Norm}. This ensures that any observed differences are driven by the normalisation function, rather than being confounded by the layer placement creating more/less favourable initial conditions. 

\texttt{Checkpoint} layers are calibrated on the first training batch immediately prior to training. They use this data to measure the standard deviation at that point, and calculate a scale factor that fixes the standard deviation to a pre-defined hyperparameter $\sigma$. All subsequent passes through the layer simply apply this scale factor. This ensures that the model is initialised with a standard deviation of $\sigma$ at that point.

We apply \texttt{Checkpoint} layers to the token embeddings $x_{type}$ ($\sigma_{type}=0.5$), and the initial embeddings $x$ ($\sigma_x=1.0$), ensuring they are relatively balanced and unit scale. In every attention layer, we apply \texttt{Checkpoint} layers to re-calibrate the possibly-Pre-normalised embeddings to $\sigma_x$ immediately before applying the $W_Q$, $W_K$, and $W_V$ operators. This counteracts the effect that transformer necessarily increases the embedding variance throughout the model at initialisation. We apply \texttt{Checkpoint} layers to $w_t$ in every attention layer, with constant $\sigma_w=0.1$. This controls the variance on the initial-state attention distribution. We apply \texttt{Checkpoint} layers to $\Delta x$ in every attention layer, with constant $\sigma_{\Delta x}=0.05$, calibrating it with respect to $x$. 

In the attention layer, we use uniform initialisation of the weight matrices $W_Q$, $W_K$, $W_V$, and $W_O$. The limits are configured to ensure that the initial state standard deviations on $w_t$ and $\Delta x$ are close to their target values. Defining $\sigma_{qk} \triangleq \sqrt[4]{\frac{\sigma_w}{N_{qkv}^3}}$, the limits are calculated as follows:

\begin{tabular}{rl}
    Weight   &   Limits   \\
    \hline
    $W_Q$   &   $\pm \sqrt{3} \sigma_{qk}$   \\
    $W_K$   &   $\pm \sqrt{3} \sigma_{qk}$   \\
    $W_V$   &   $\pm \sqrt\frac{3}{N_{qkv}}$    \\
    $W_O$   &   $\pm \sqrt{\frac{3}{H N_{qkv}}}$     \\
\end{tabular}

However, we note that this initialisation is superseded by the calibration of the \texttt{Checkpoint} layers for determining the initial state, and we include it only to promote numerical stability. All other feed-forward layers use \texttt{Glorot uniform} \cite{pmlr-v9-glorot10a} initialisation, as implemented in \texttt{Keras} \cite{chollet2015keras}. Normalisation gain parameters are initialised to $1$ and biases, where used, to $0$.

\subsection{Training algorithm}

\begin{figure}[h]
\includegraphics[width=\textwidth]{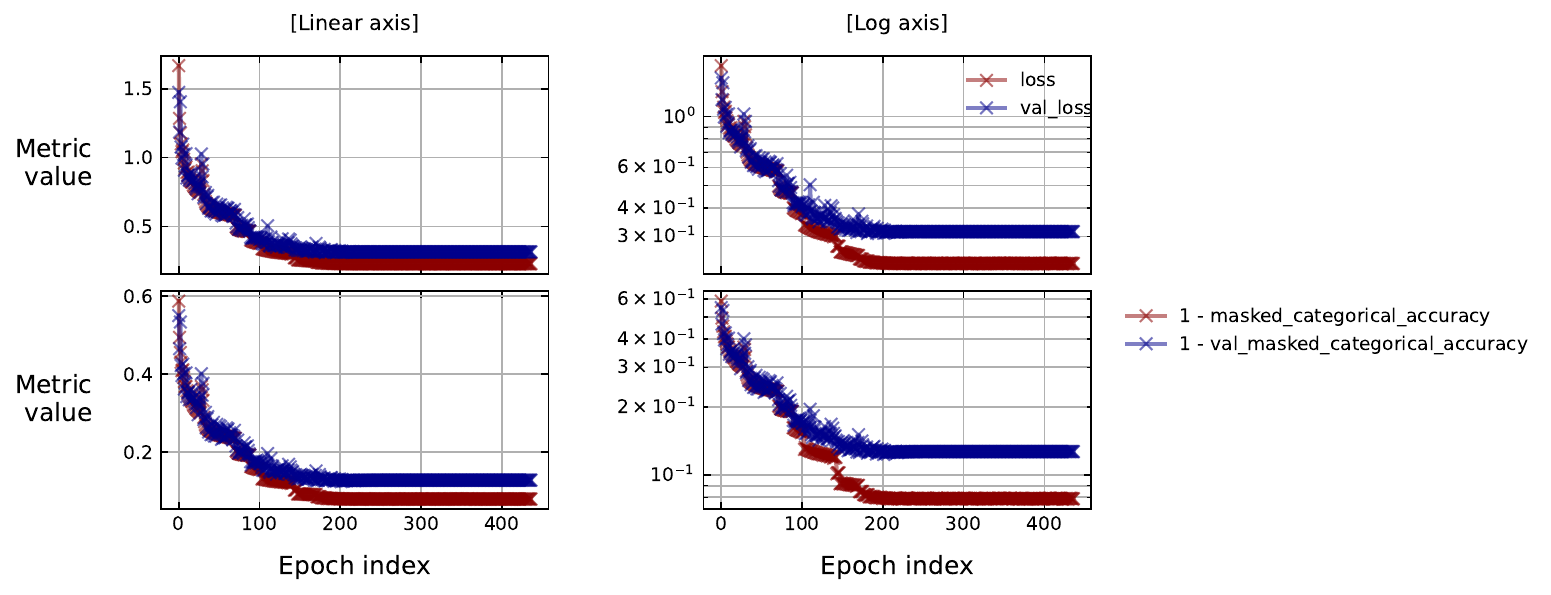}
\caption{Model training curves for the \texttt{Baseline} \texttt{Pre-Norm} configuration.}
\label{fig: model train curves}
\end{figure}

We train using the \texttt{AdamW} optimiser \cite{loshchilov2019decoupled} with learning rate $3\times10^{-4}$ and weight decay of $0.01$, with all other parameters following their default values  in \texttt{TensorFlow+Keras v2.15.0}. Each epoch consists of $2000$ batches of $128$ datapoints. For the main experiments and model variations, we use an \textit{adaptive learning rate decay} strategy. This means that the learning rate is multiplied by a factor of $0.5$ if the training loss does not improve for $3$ consecutive epochs. We find that this balances training speed with improved performance by using small learning rates later in training. Training is halted after two days of wall time, which we observe to allow model convergence, as shown in Figure~\ref{fig: model train curves}. For the model stability scan, training is run for $60$ hours, and learning rate is not allowed to decay (stability with respect to learning rate being one of the targets of study).

\subsection{Computational resources}

The main experiments are all performed on a single \texttt{Nvidia v100-SXM2-16GB (Volta)} GPU. The scan of models used for the stability analysis were trained on a batch cluster with a variety of compute nodes, using $8$ cores per training run. A representative compute node is \texttt{2×12-core Intel Xeon E5-2690 v3 @ 2.60GHz + 128GB RAM}.

\subsection{Environment details}

%Training logs provide a dump of all Python packages used in the environment. Along with the quoted random seeds (also in the log files) and supplementary code, this is sufficient for replicating our models. 
The main contributing package versions are as follows:

\begin{tabular}{rl}
    Package    &    Version    \\
    \hline
    Python       &   \texttt{3.11.5 (main, Sep 11 2023, 13:54:46) [GCC 11.2.0]}  \\
    TensorFlow   &   \texttt{2.15.0}  \\
    Keras        &   \texttt{2.15.0}  \\
    NumPy        &   \texttt{1.26.2}  \\
\end{tabular}

% \subsection{Run commands}

% Training runs are executed from the \texttt{remote\_training\_env/run} a command such as

% \texttt{python train\_model.py -{}-env <env.cfg> -{}-model <model\_base.cfg> <model.cfg> -{}-training <training.cfg> -{}-strategy PRE\_VECNORM -{}-tag RUN\_TAG -{}-duration 2days -{}-seed 100}

% along with the provided config files. Plotting is achieved by running the notebooks in the directory the \texttt{local\_analysis\_env/run}, pointing the appropriate config value to the trained model. Strategies are PRE\_VECNORM and HEAD\_VECNORM for \texttt{Pre-Norm} and \texttt{QKV-Norm} respectively.

%%
%%  Appendix:  Main experiments 
%%
\clearpage
\section{Extended main experiments}
\label{appendix: extended experiments}

This appendix provides an extended explanation of the experimental results in section~\ref{sec: experimental results}.

\subsection{Embedding structure}

\begin{figure}[h]
\includegraphics[width=\textwidth]{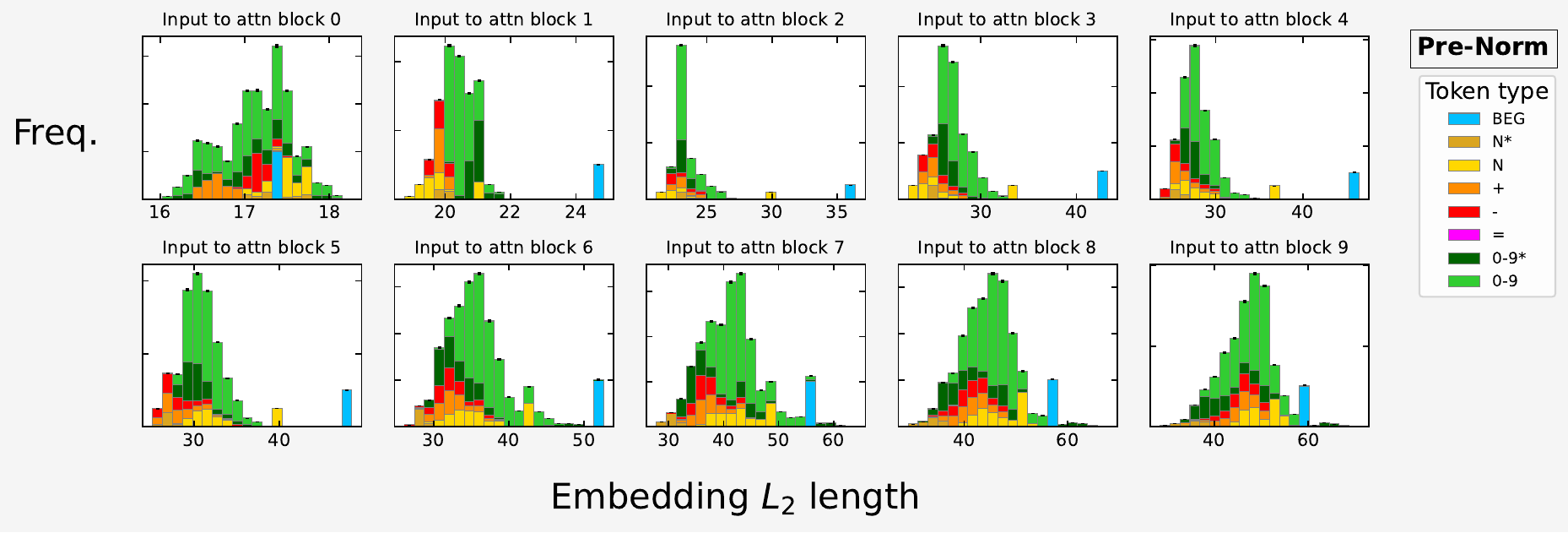}
\caption{Distribution of embedding $L_2$-norms at different model depths using the \texttt{Baseline Pre-Norm} model.}
\label{fig: embedding L2 norms prenorm}
\vspace{0.4cm}
\includegraphics[width=\textwidth]{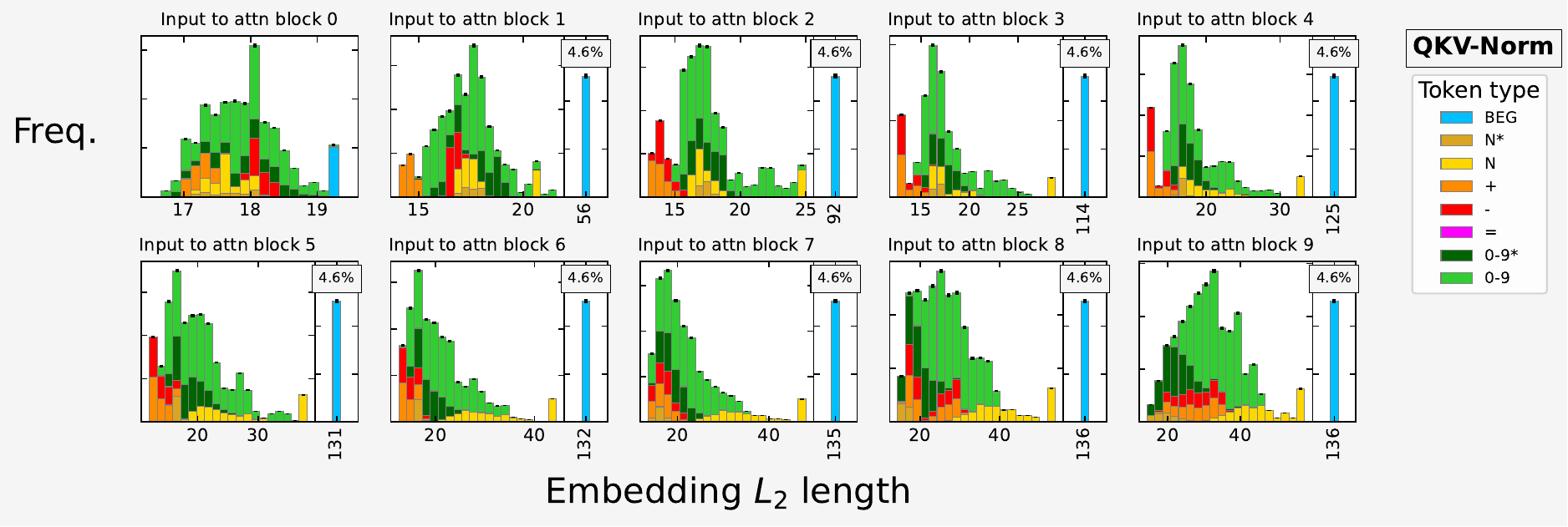}
\caption{Distribution of embedding $L_2$-norms at different model depths using the \texttt{Baseline QKV-Norm} model.}
\label{fig: embedding L2 norms qkvnorm}
\end{figure}

Figure~\ref{fig: embedding spread} presented the spread of embedding $L_2$-norms as a function of model depth. Let us now describe in detail how this plot was made. Figure~\ref{fig: embedding L2 norms prenorm} shows the distribution of embeddings at the input to every attention layer, for \texttt{Baseline} models trained using \texttt{Pre-Norm} (top) and \texttt{QKV-Norm} (bottom). Colours represent the initial token type corresponding to that embedding. Asterisks denote tokens in the \textit{answer}, with all labels denoting the \textit{question}.

We see that the begin-sequence token (\texttt{BEG}) is often separate from the distribution, which may be because it remains non-annotated and fulfils a qualitatively different role. We remove this from our estimates to avoid erroneously inflating the spread. Interestingly, \texttt{BEG} still tends to be close to the main bulk for \texttt{Pre-Norm}, but can be very far for \texttt{QKV-Norm}, and we have to use overflow panels to capture it. This is consistent with our hypothesis that \texttt{Pre-Norm} stability requires embeddings to have similar norms, whilst \texttt{QKV-Norm} does not require this.

\begin{figure}[!]
\includegraphics[width=\textwidth]{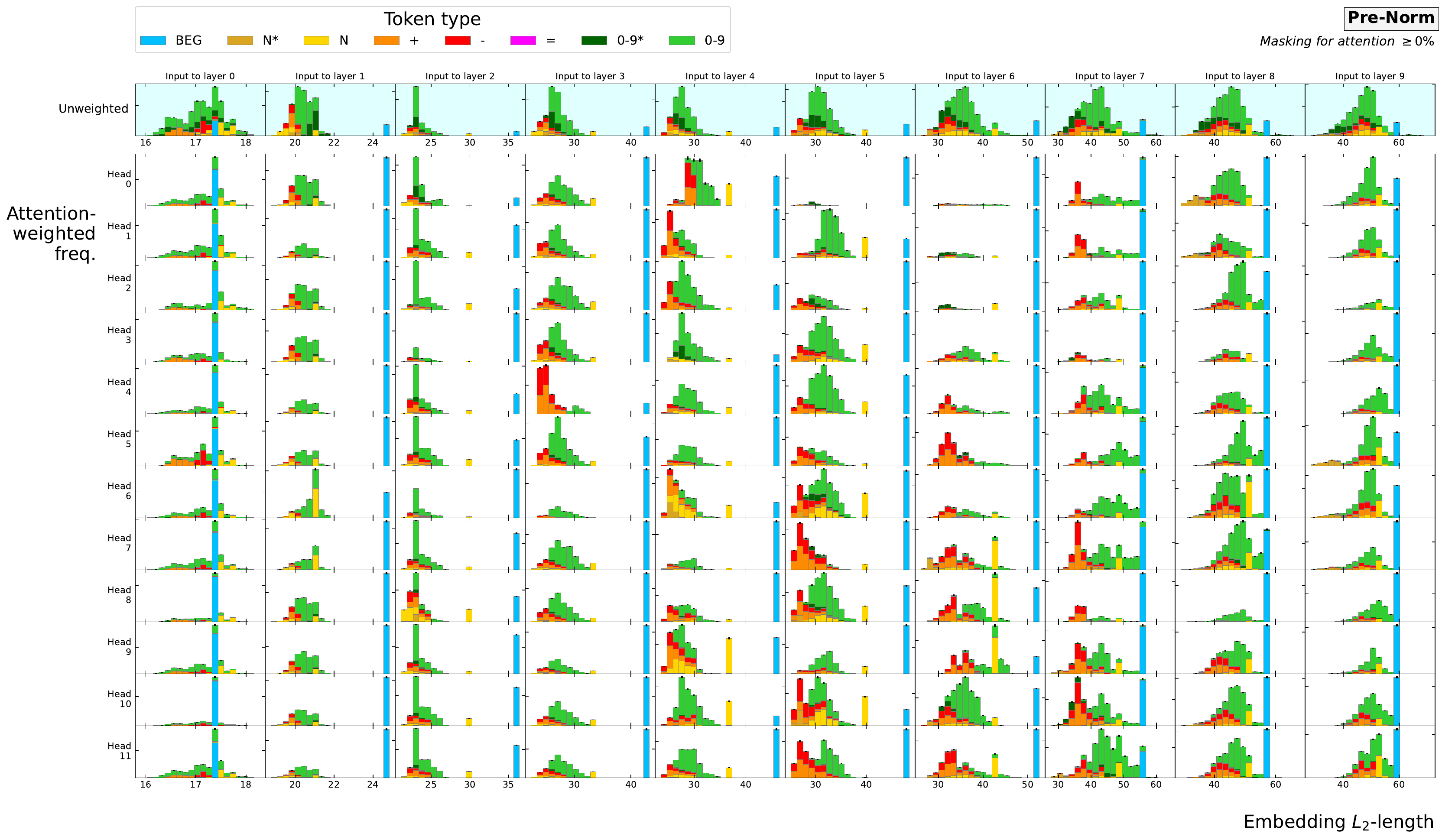}
\caption{Embedding distributions at different model depths using the \texttt{Baseline Pre-Norm} model. The categories \texttt{0-9} and \texttt{N} are separated into whether they occur in the question (light colour) or answer (dark colour, distinguished by $*$ label). These distributions are used to compute the LHS of Figure~\ref{fig: embedding spread} after removing the \texttt{BEG} tokens.}
\label{fig: embedding L2 norms prenorm head}
\vspace{0.6cm}
\includegraphics[width=\textwidth]{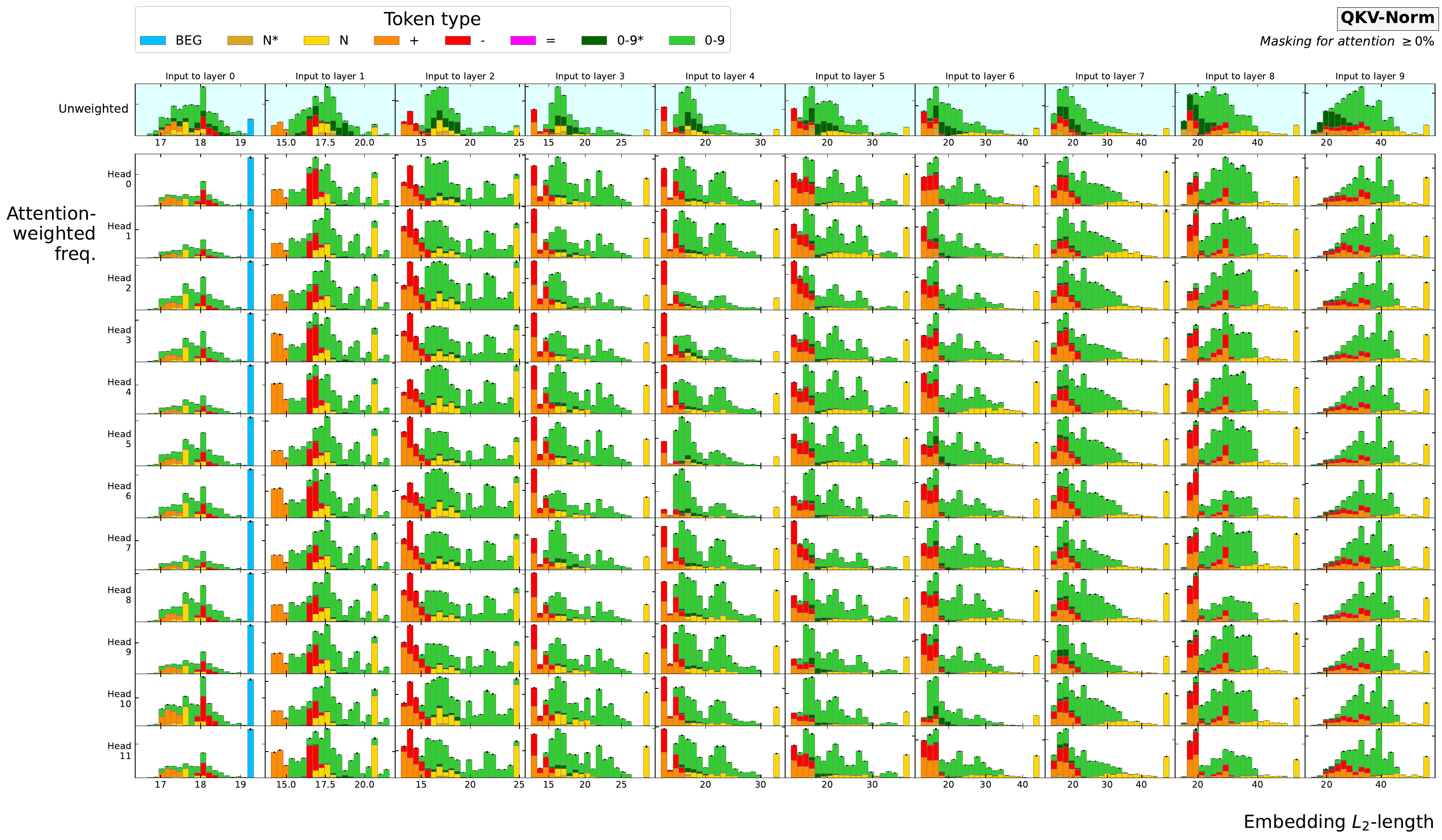}
\caption{Embedding distributions at different model depths using the \texttt{Baseline QKV-Norm} model. The categories \texttt{0-9} and \texttt{N} are separated into whether they occur in the question (light colour) or answer (dark colour, distinguished by $*$ label). Note that overflow panels are excluded from this plot for legibility. These distributions are used to compute the RHS of Figure~\ref{fig: embedding spread} after removing the \texttt{BEG} tokens.}
\label{fig: embedding L2 norms qkvnorm head}
\end{figure}

It is not sufficient to simply measure the spread of Figures~\ref{fig: embedding L2 norms prenorm} and \ref{fig: embedding L2 norms qkvnorm}, because an attention head may not be sensitive to all embeddings in the layer. The easiest way to account for this is to weight every embedding according to its assigned attention. Secondly, the distribution is expected to be narrow only on a per-head basis, and there is no reason why distinct heads cannot be centred around different medians. We therefore calculate the weighted distribution of embeddings on a per-head basis, as shown in Figures~\ref{fig: embedding L2 norms prenorm head}-\ref{fig: embedding L2 norms qkvnorm head}. 

\subsection{Circuit collapse}

\begin{figure}[h]
    \centering
    \includegraphics[width=\textwidth]{figures/collapse/circuit_collapse_probability_min_att_0.95_0.0_loose.pdf}
    \vspace{0.4cm}
    \includegraphics[width=\textwidth]{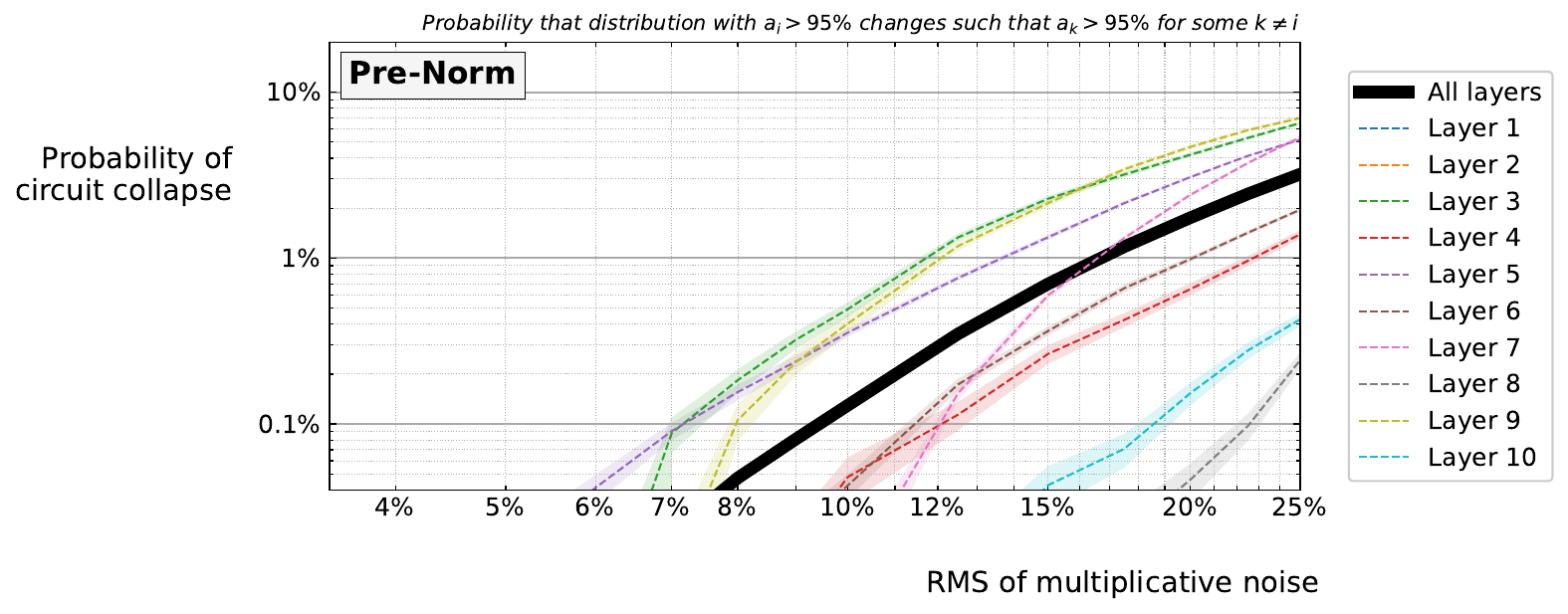}
    \caption{Probability of circuit collapse vs increasing noise. \textbf{Top:} using the baseline definition. This is a reproduction of Figure~\ref{fig: circuit collapse}. \textbf{Bottom:} requiring the attention distribution to remain sparse after switching to a different token.}
    \label{fig: circuit collapse sparse}
\end{figure}

Figure~\ref{fig: circuit collapse} shows the probability of circuit collapse. This is the probability that an attention distribution with no noise selects embedding $i$ with high probability $a_i \geq 95\%$, and when noise is added, it transitions such that some $k \neq i$ becomes the maximum attended embedding. This definition is chosen because it matches our theoretical results in section~\ref{sec: theory: circuit stability}. However, it does not require that the distribution remains sparse after the noise addition. Figure~\ref{fig: circuit collapse sparse} compares this baseline result (top) with an alternative definition (bottom), in which the second distribution must also be sparse, meaning $a_k\geq95\%$ for some $k \neq i$. We see that $1\%$ of sparse attention distributions collapse at a noise level of $11\%$ when using the original definition, delayed until $17\%$ when using the sparse definition. Therefore we observe that the sparse-to-sparse case does occur, but requires a higher noise level.

%%
%%  Appendix:  Reproducibility of results across model variations
%%
\clearpage
\section{Main experiments: results with different models}
\label{app: model variations}

In this appendix we reproduce the main experimental results using our model variations. 

\subsection{Embedding lengths}

\begin{figure}[h]
\includegraphics[width=\textwidth]{figures/baseline_model/Embedding_length_comparison_0.0.pdf}
\caption{Attention-weighted spread of embeddings at increasing model depth using the \texttt{Baseline} model and task configuration. This is a replication of Figure~\ref{fig: embedding spread}.}
\label{fig: embedding spread: model variations BEG}
\includegraphics[width=\textwidth]{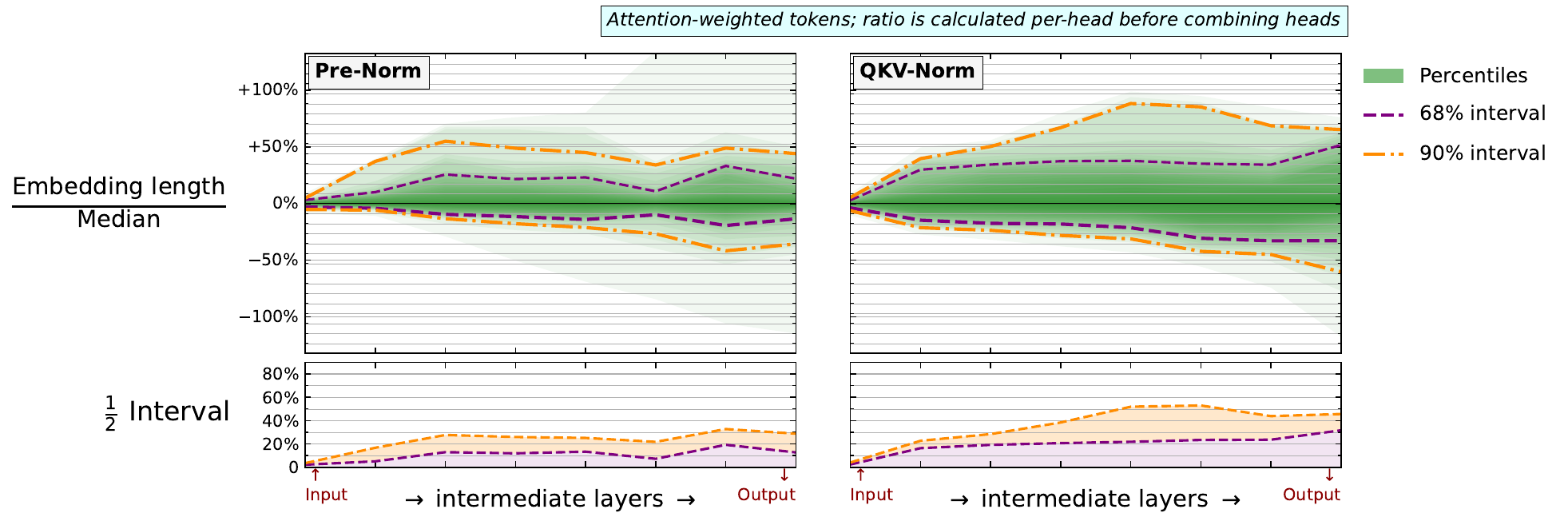}
\caption{Attention-weighted spread of embeddings at increasing model depth using the \texttt{Alternate} model and task configuration.}
\label{fig: embedding spread: model variations MID}
\includegraphics[width=\textwidth]{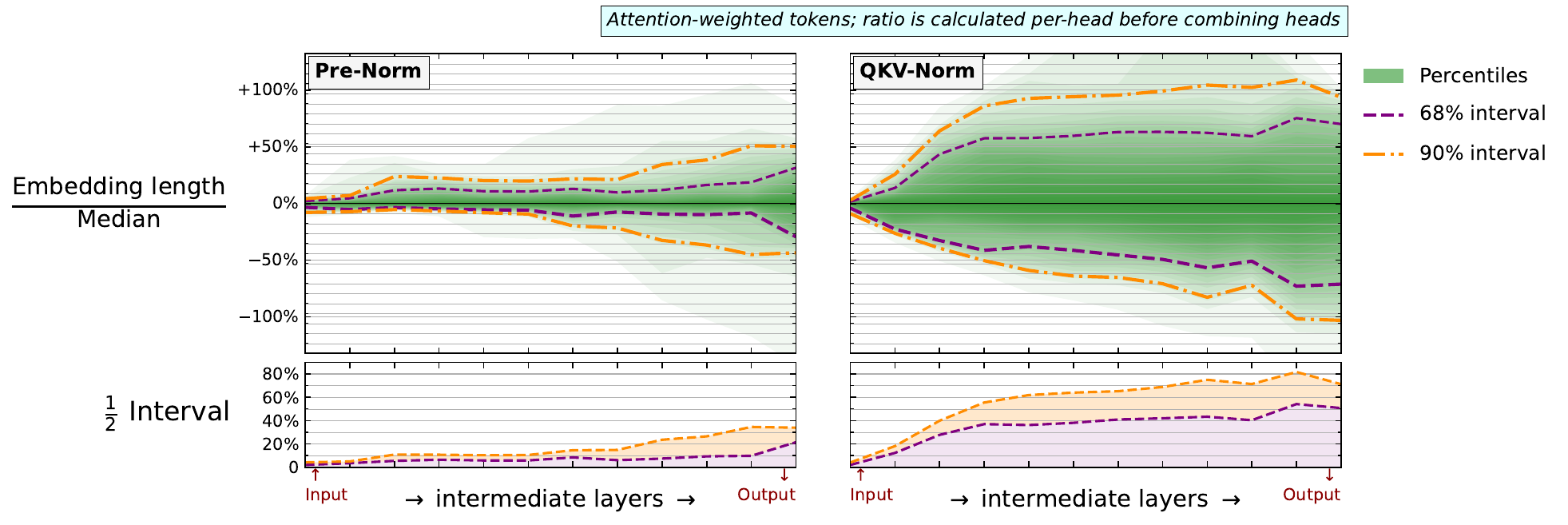}
\caption{Attention-weighted spread of embeddings at increasing model depth using the \texttt{Large} model and task configuration.}
\label{fig: embedding spread: model variations END}
\end{figure}

Figure~\ref{fig: embedding spread} shows the empirical results demonstrating the attention-weighted spread of embeddings. Figures~\ref{fig: embedding spread: model variations BEG}-\ref{fig: embedding spread: model variations END} show the results we obtain when we perform the same analysis using the \texttt{Alternate} and \texttt{Large} model variations. In all cases, we observe 90\% of embeddings within a spread of roughly $\pm30\%$ when using \texttt{Pre-Norm}. In all cases, the spread of embeddings for \texttt{QKV-Norm} is larger, although we note that the effect is smaller when using the \texttt{Alternate} variation.

\subsection{Model stability with simulated inference}

\begin{figure}[h]
\includegraphics[width=\textwidth]{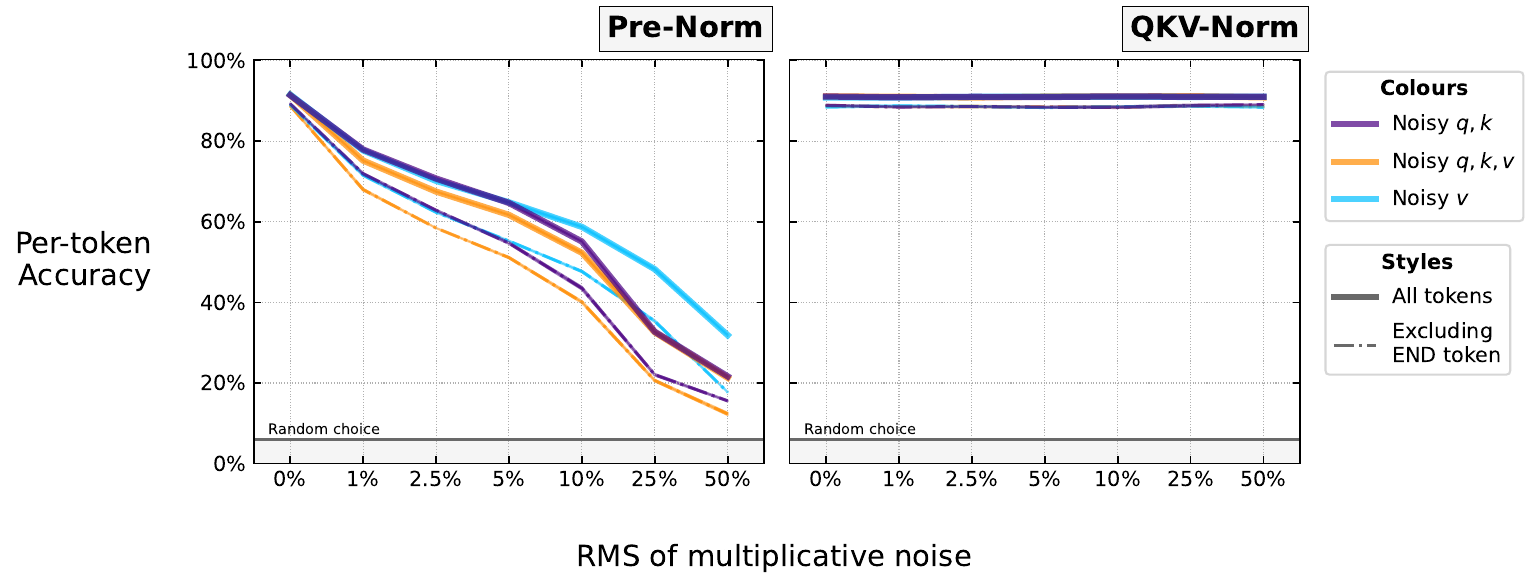}
\caption{Evolution of per-token accuracy as we increase noise on the $L_2$-norms of $\{q,k_t,m_t\}$ for the \texttt{Baseline} model and task configuration.}
\includegraphics[width=\textwidth]{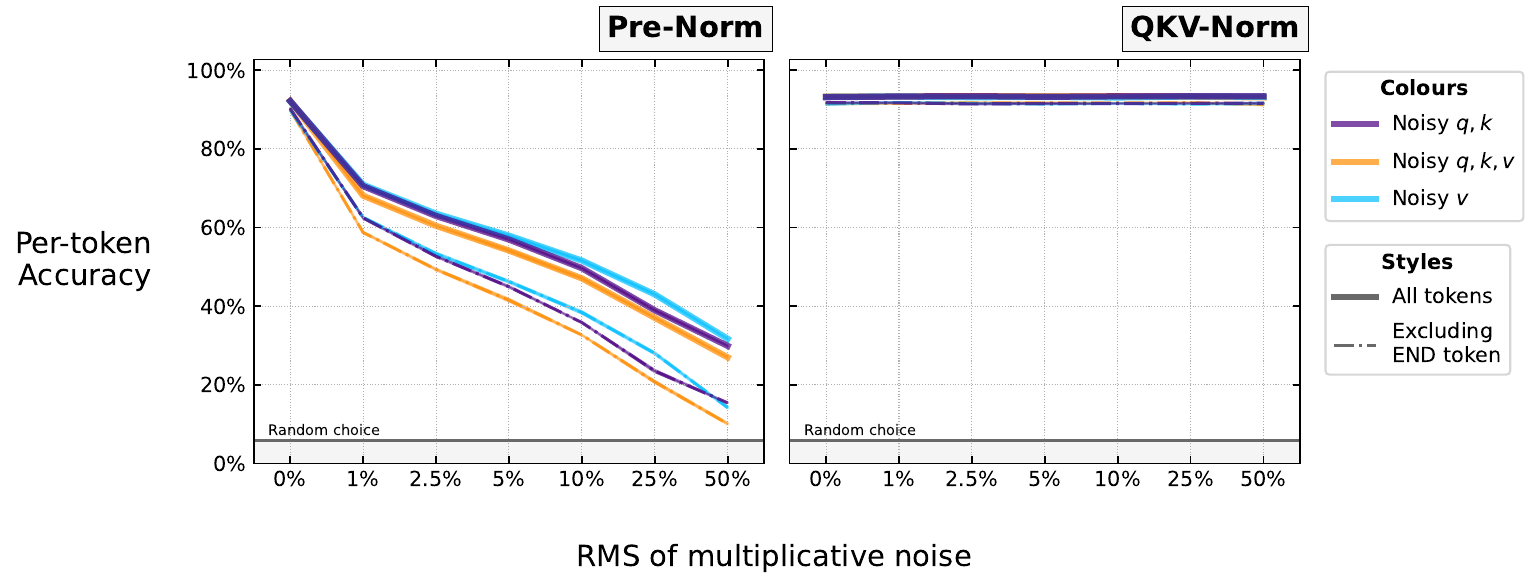}
\caption{Evolution of per-token accuracy as we increase noise on the $L_2$-norms of $\{q,k_t,m_t\}$ for the \texttt{Alternate} model and task configuration.}
\label{fig: noise sensitivity: model variations}
\end{figure}

Figure~\ref{fig: noise sensitivity}(left) shows the stability of the model predictions under simulated interference. Figure~\ref{fig: noise sensitivity: model variations} shows the results we obtain when we perform the same analysis using the \texttt{Alternate} model variation. The \texttt{Large} model was not run due to its high computational load. We find that the \texttt{Alternate} model has a larger effect size that \texttt{Baseline}, with a $\gtrsim20\%$ loss of per-token accuracy with only a $1\%$ noise effect. For completeness, we show \texttt{QKV-Norm} on the RHS. This is stable by construction, and only jitter due to finite sampling is observed. In these plots, we estimate the statistical uncertainty by evaluating over three datasets and calculating the standard error on the mean. This is plotted as a shaded band, but tends to be narrower than the line width.

\begin{figure}[h]
\includegraphics[width=\textwidth]{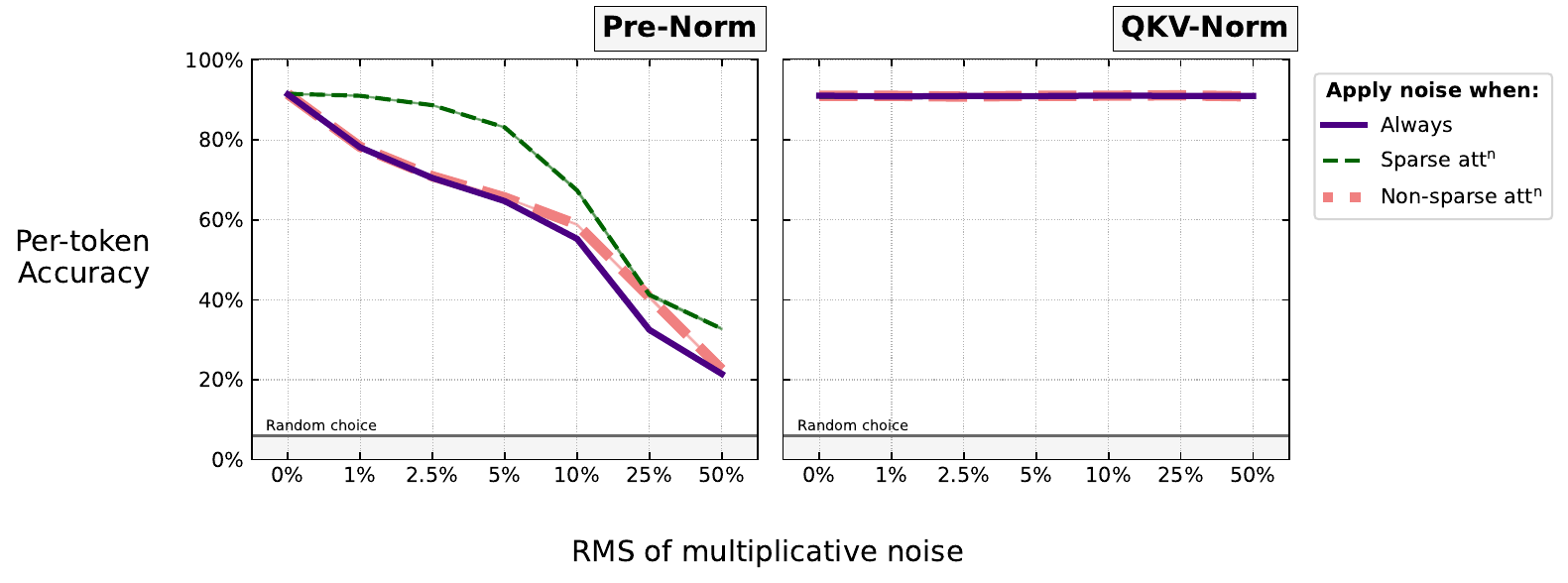}
\caption{Evolution of per-token accuracy as we increase noise on the $L_2$-norms of $\{q,k_t,m_t\}$ for the \texttt{Baseline} model and task configuration.}
\includegraphics[width=\textwidth]{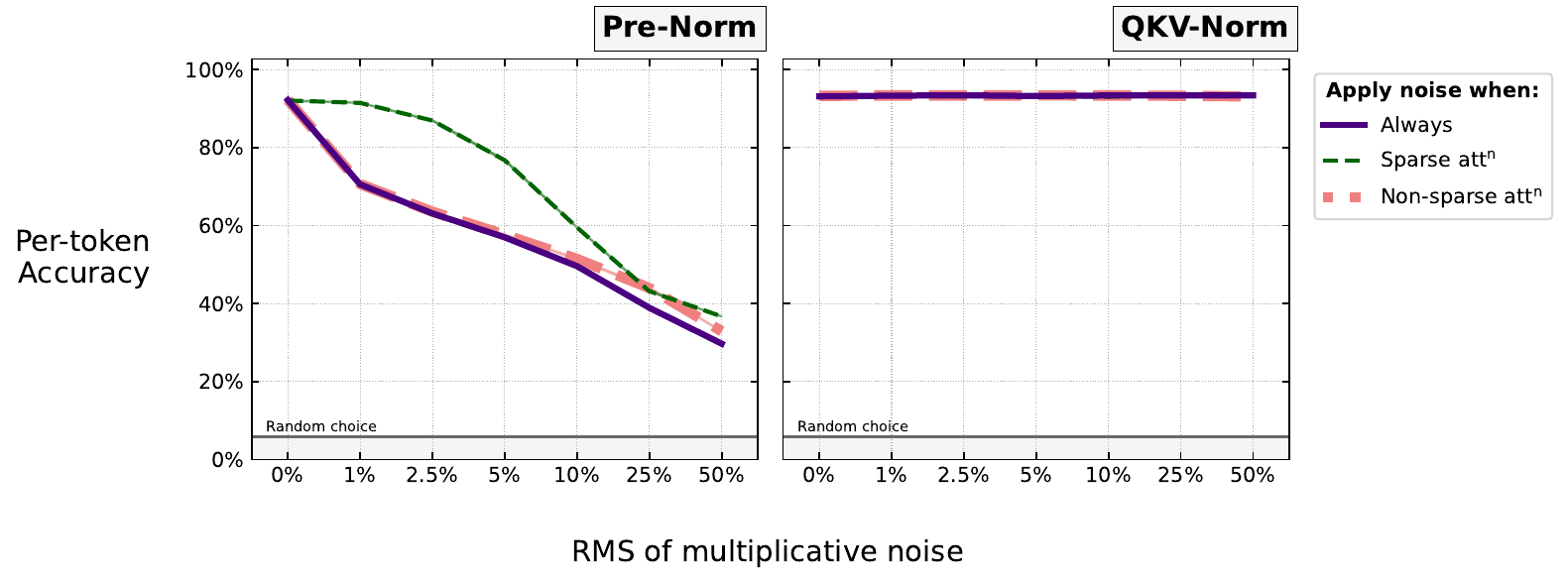}
\caption{Evolution of per-token accuracy as we increase noise on the $L_2$-norms of $\{q,k_t,m_t\}$ for the \texttt{Alternate} model and task configuration.}
\label{fig: noise sensitivity: model variations sparse}
\end{figure}

Figure~\ref{fig: noise sensitivity}(right) compares the stability when we only apply noise to sparse heads (defined as $\max_i a_i \geq 95\%$, thin dashed line) and non-sparse heads (defined as $\max_i a_i < 70\%$, thick dashed line). Figure~\ref{fig: noise sensitivity: model variations sparse} compares these results with the \texttt{Alternate} model variation. In both experiments, sparse-attention is stable under \%-level noise, and non-sparse distributions dominate this regime.

Note that later layers experience both the artificial noise injection as well as perturbation of their inputs due to the compounding of errors caused by noise in the previous layers.

\subsection{Circuit collapse}

\begin{figure}[h]
\includegraphics[width=\textwidth]{figures/collapse/circuit_collapse_probability_min_att_0.95_0.0_loose.pdf}
\caption{Probability of \textit{circuit collapse} vs increasing noise using the \texttt{Baseline} model and task configuration. This is a replication of Figure~\ref{fig: circuit collapse}.}
\includegraphics[width=\textwidth]{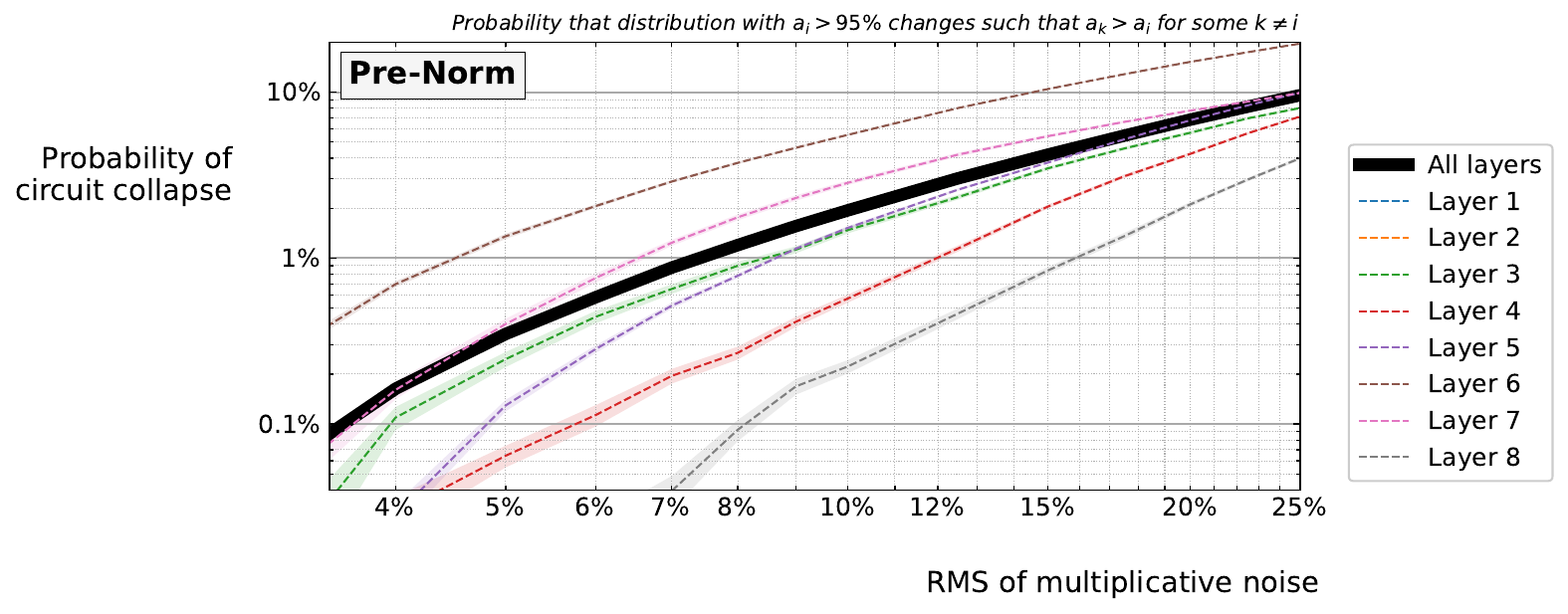}
\caption{Probability of \textit{circuit collapse} vs increasing noise using the \texttt{Alternate} model and task configuration.}
\includegraphics[width=\textwidth]{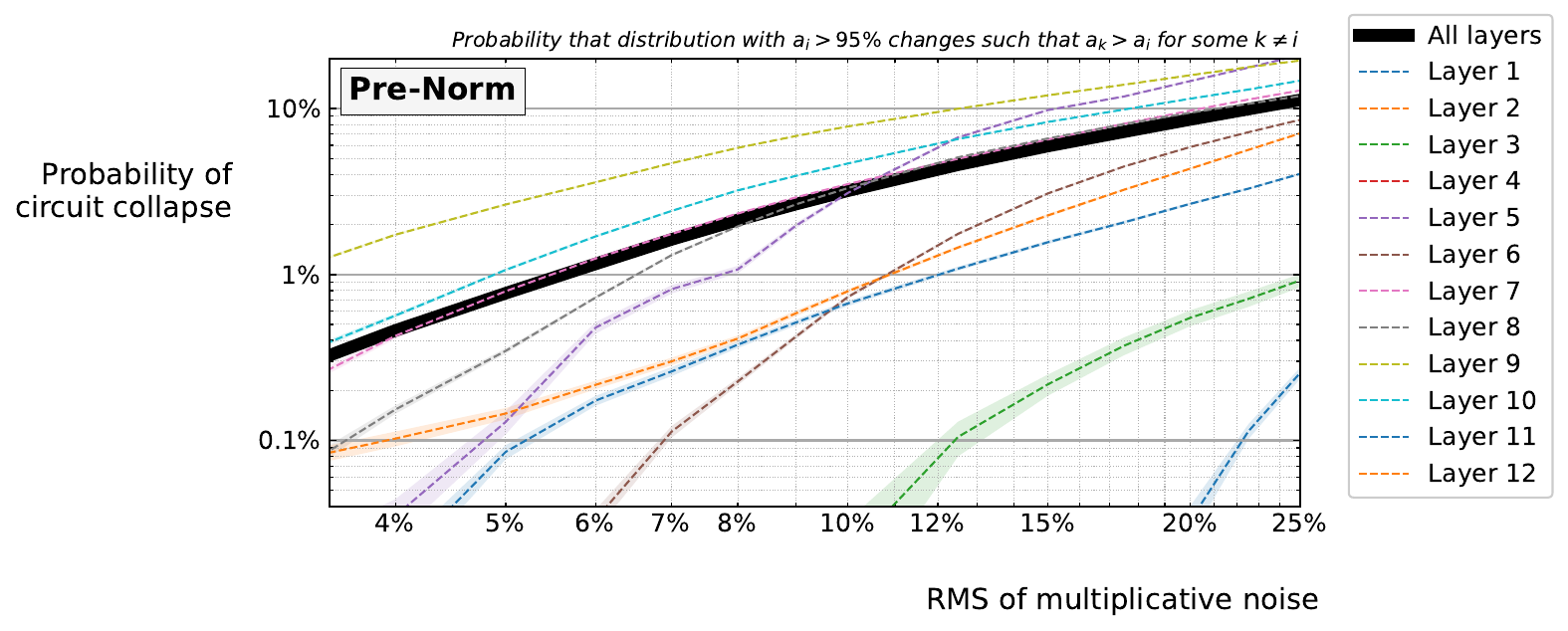}
\caption{Probability of \textit{circuit collapse} vs increasing noise using the \texttt{Large} model and task configuration.}
\label{fig: circuit collapse: model variations}
\end{figure}

Figure~\ref{fig: circuit collapse} shows the probability of circuit collapse. This is the probability that an attention distribution with no noise selects embedding $i$ with high probability $a_i \geq 95\%$, and when noise is added, it transitions such that some $k \neq i$ becomes the maximum attended embedding. Figure~\ref{fig: circuit collapse: model variations} shows the results we obtain when we perform the same analysis using the \texttt{Alternate} and \texttt{Large} model variations. In both cases, we observe the onset of circuit collapse at smaller noise levels. Whilst the \texttt{Baseline} model observed that 1\% of sparse attention heads collapsed with 11\% noise, this value is 7.5\% for \texttt{Alternate} and 5.5\% for \texttt{Large}.

%%
%%  Appendix:  Pre-Norm vs QKV-Norm
%%
\clearpage
\section{Additional comparisons between Pre-Norm and QKV-Norm}
\label{appendix: Pre vs QKV norm}

\subsection{Model performance}

Table~\ref{table: model performance} shows the per-token accuracy performance for the trained \texttt{Baseline} models. \texttt{Pre-Norm} and \texttt{QKV-Norm} have comparable in-distribution per-token accuracies of $91.4\%$ and $91.0\%$ respectively. However, performance drops to $87.5\%$ ($82.5\%$) for generalisation to intermediate task difficulty, and $66.7\%$ ($46.8\%$) for increased difficulty. The performance drop of \texttt{QKV-Norm} implies that it has learned a less generalisable solution. This re-enforces our motivation that architectural changes should be important for the inductive bias of a model.

\begin{table}[h]
\centering
\begin{tabular}{rccc}
Dataset               &    \texttt{Pre-Norm}    &    \texttt{QKV-Norm}   \\
\hline
In-distribution       &    $91.38\pm0.04\%$     &    $90.99\pm0.03\%$    \\
OOD (interpolation)   &    $87.46\pm0.04\%$     &    $82.54\pm0.04\%$    \\
OOD (extrapolation)   &    $66.65\pm0.05\%$     &    $46.76\pm0.05\%$    \\
\end{tabular}
\vspace{0.2cm}
\caption{Per-token accuracy for the \texttt{Baseline} models. Dataset configurations are shown in Table~\ref{table: dataset specifications main}.}
\label{table: model performance}
\end{table}

\subsection{Training stability}

\begin{figure}[h]
\includegraphics[width=\textwidth]{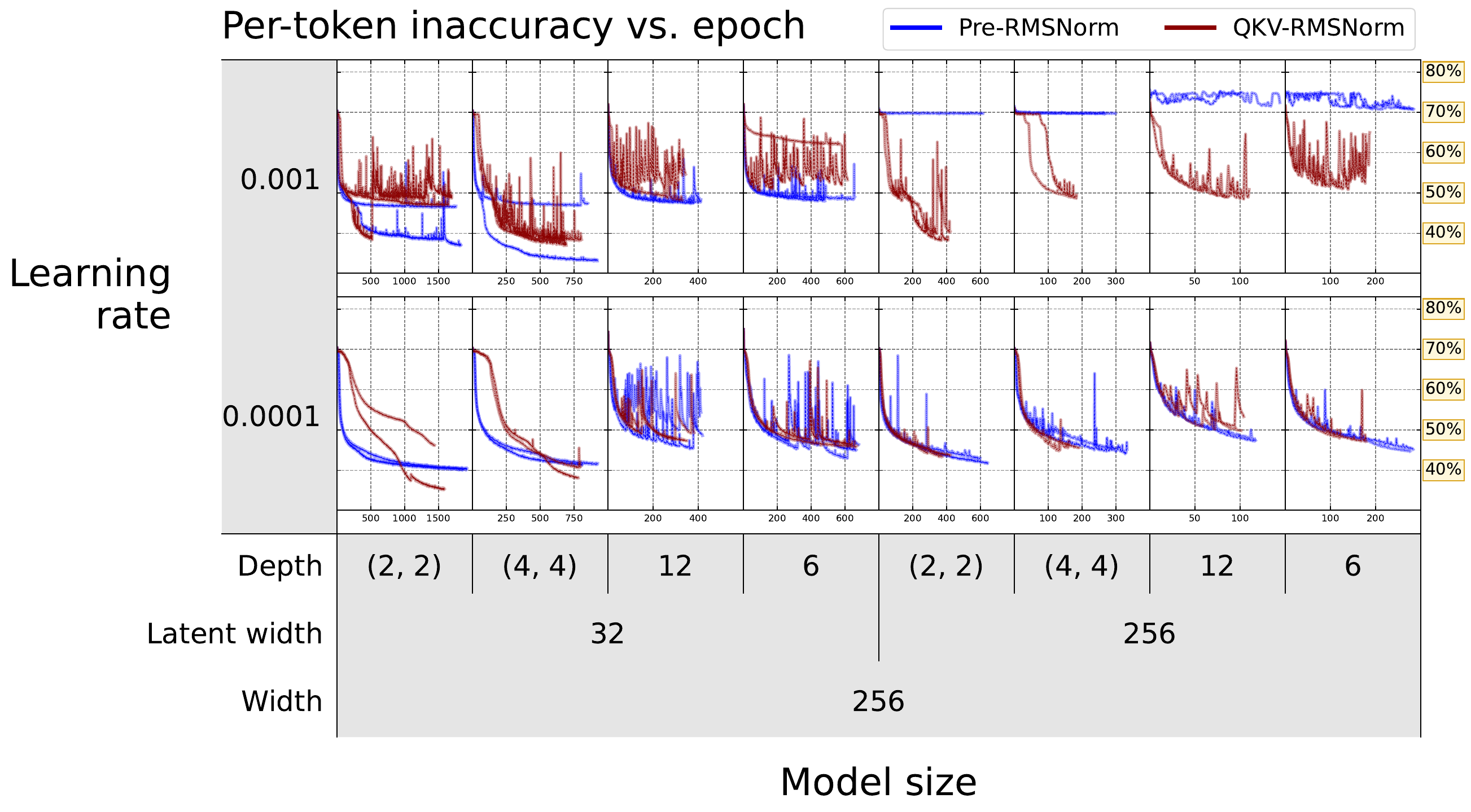}
\caption{Training curves when learning the task configuration shown in Table~\ref{table: model spec stability}.}
\label{fig: training stability scan}
\end{figure}

Changing the normalisation layer is expected to affect the training rate and stability. To investigate this, Figure~\ref{fig: training stability scan} shows the training curves for different model sizes and learning rates. The task is configured as presented in Table~\ref{table: model spec stability}. The \texttt{Depth} parameter is the number of layers, where brackets indicate the values for an encoder-decoder model. For example, $(2,2)$ means that we use $2$ encoder blocks and $2$ decoder blocks. Each decoder block has a self-attention and a cross-attention layer, and so the total model has $6$ attention layers. A single \texttt{Depth} value indicates a decoder architecture, with the number of layers shown. \texttt{Width} is the number of neurons per layer, and \texttt{Latent width} is the number of neurons on the space of $\{q,~k_t,v_t\}$ (called $N_{qkv}$ in section~\ref{sec: formulation}). Training curves on the top row use a learning rate of $0.001$, whilst the bottom row use a value of $0.0001$. In each panel, two training runs are shown, with different random seeds. \texttt{Pre-Norm} is shown in blue, and \texttt{QKV-Norm} in red.

We find that \texttt{Pre-Norm} training is unstable for large learning rates and model sizes, as shown by the flat blue curves in the top right hand panels. Similar stabilisation improvements at large learning rate is reported for \texttt{QK-Norm} in \cite{wortsman2023smallscale,dehghani2023scaling}, which applies layer normalisation to $\{q,~k_t\}$ but not $v_t$, as for \texttt{QKV-Norm}. However, we note that training large models with a smaller learning rate leads to improved model performance, as shown by the panels on the bottom right. Finally, we note that both methods typically train the model at similar rates, however small model training follows a very different trajectory, with \texttt{QKV-Norm} learning more slowly at the beginning of training (bottom left panels). There is also some visible evidence that small model training is actually \textit{less} effective when using \texttt{QKV-Norm} with a large learning rate (top left panels).

\subsection{Attention sparsity}

\begin{figure}[h]
\centering
\includegraphics[width=0.9\textwidth]{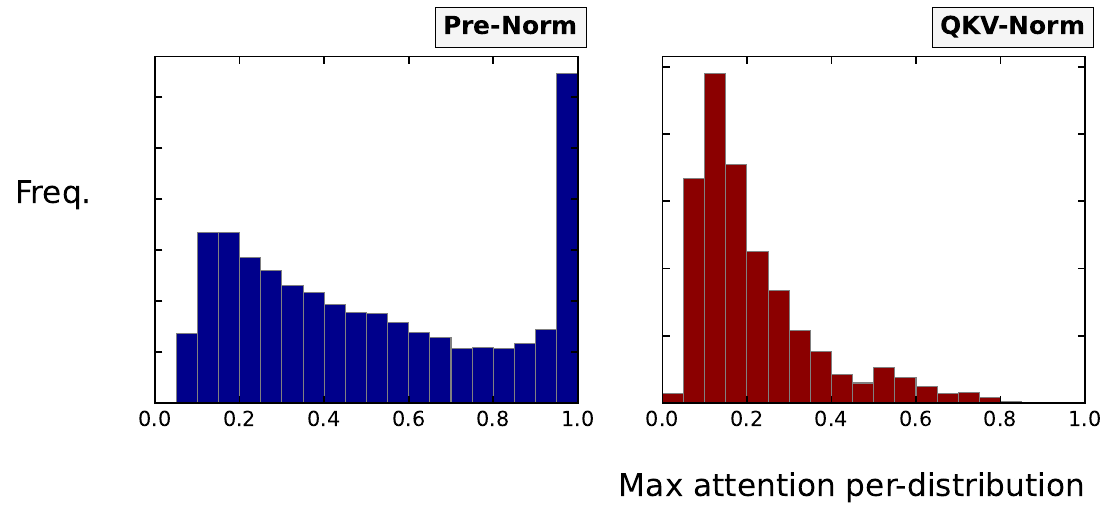}
\caption{Distribution of the maximum attention observed per-distribution, i.e. $\max_i a_i$, in the \texttt{Baseline} case. We observe that the \texttt{Pre-Norm} model often utilises sparse-attention, as seen by the peak at $1$. By contrast, \texttt{QKV-Norm} shows no such peak. Similar behaviour is reported for \texttt{QK-Norm} in \cite{DBLP:journals/corr/abs-2010-04245}.}
\label{fig: attention sparsity distribution}
\end{figure}

We find that our \texttt{Pre-Norm} models often exploit sparse-attention, whereas models trained with \texttt{QKV-Norm} do not. Similar behaviour is reported for \texttt{QK-Norm} in \cite{DBLP:journals/corr/abs-2010-04245}. For a systematic comparison, Figure~\ref{fig: attention sparsity distribution} shows a histogram of the maximum attention observed per-distribution (i.e. a histogram of $\max_i a_i$). When making this plot, we do not consider the first row of the attention matrix, in which the \texttt{[} token attends fully to itself.

\begin{figure}
\centering
\includegraphics[width=0.87\textwidth]{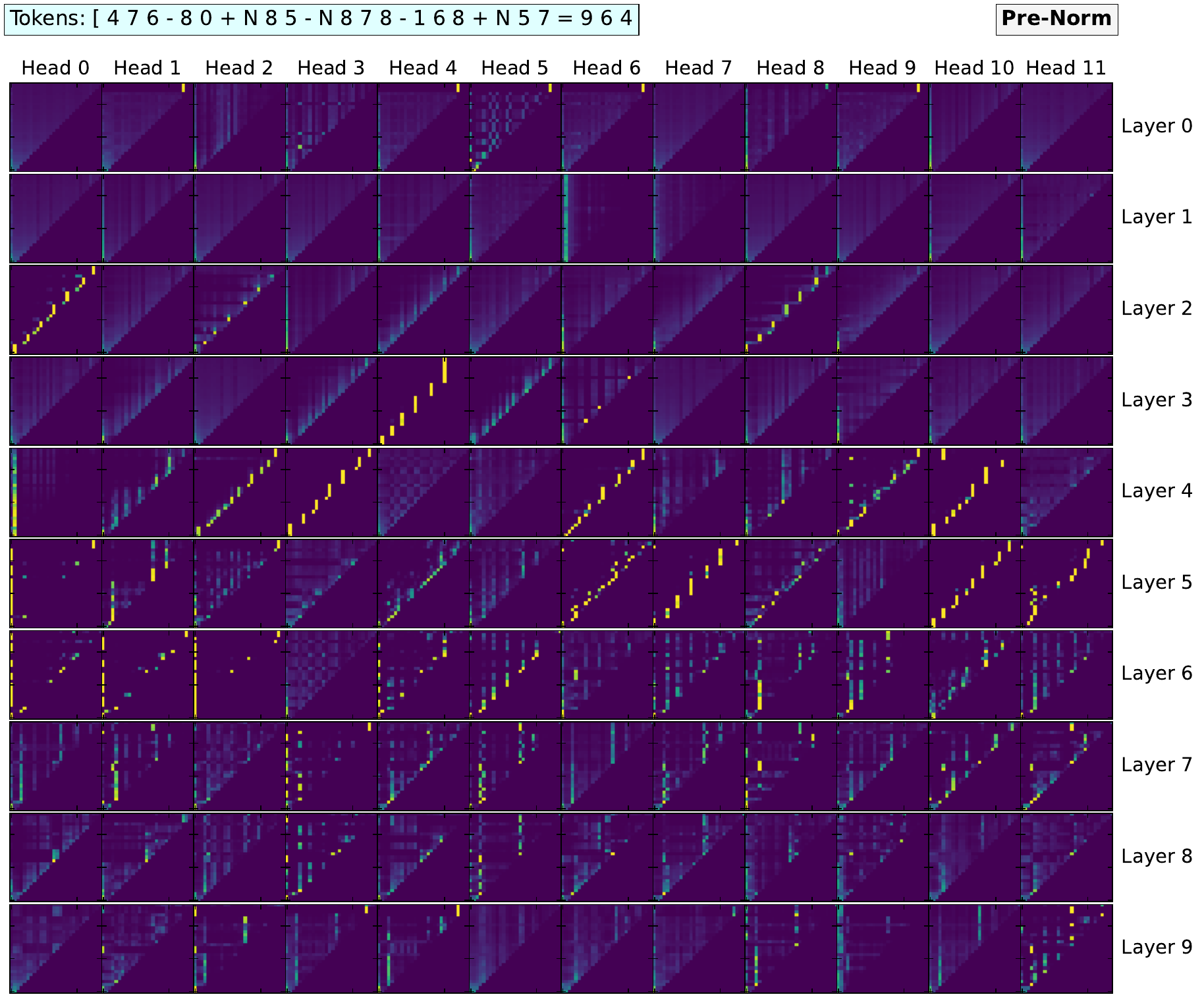}
\caption{Attention maps for a random in-distribution example using the \texttt{Baseline} \texttt{Pre-Norm} model. Several attention heads create sparse attention distributions.}
\label{fig: attention map prenorm}
\vspace{0.3cm}
\includegraphics[width=0.87\textwidth]{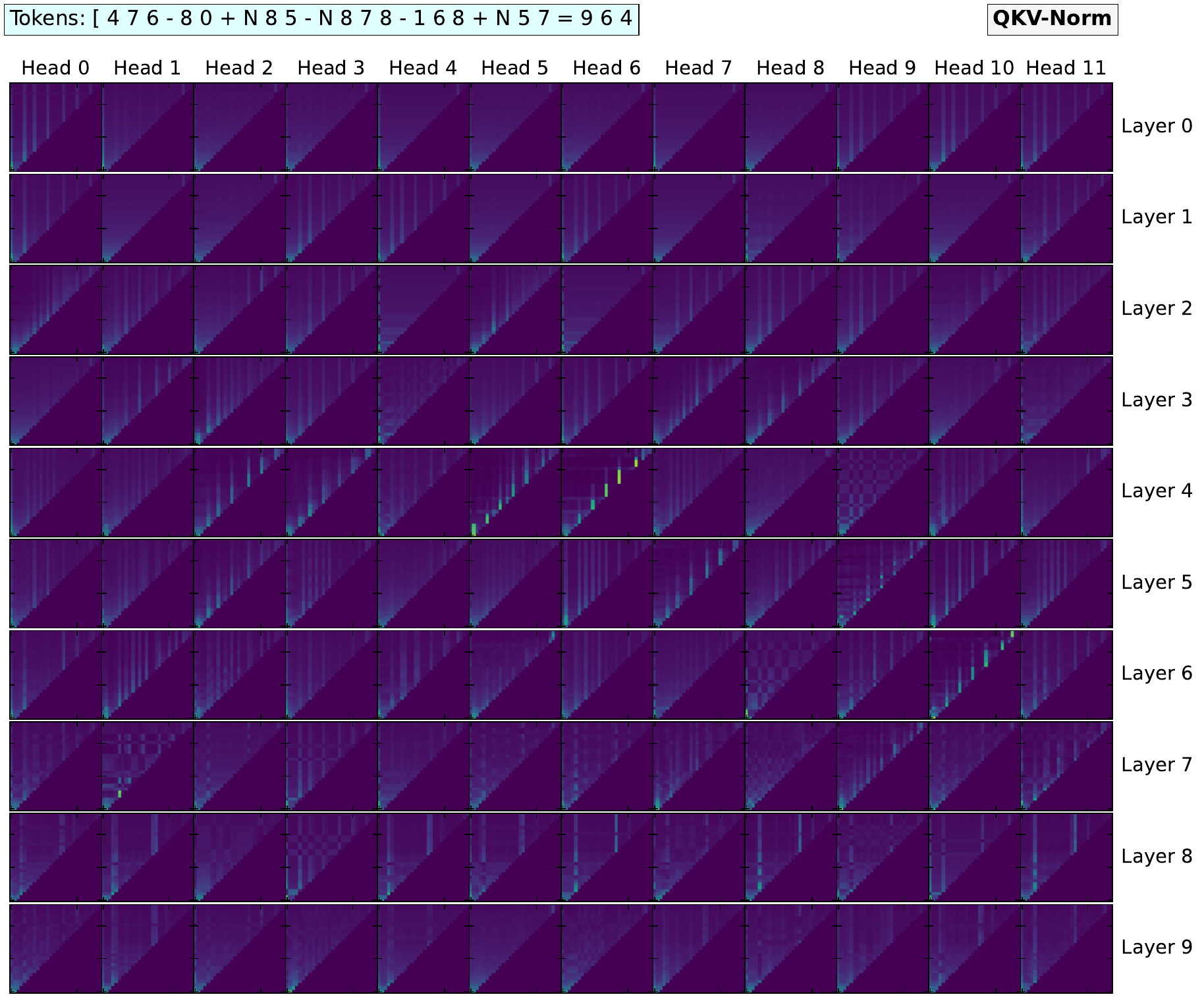}
\caption{Attention maps for a random in-distribution example using the \texttt{Baseline} \texttt{QKV-Norm} model. We observe much less sparsity than in the \texttt{Pre-Norm} model, shown in Figure~\ref{fig: attention map prenorm}. Similar behaviour is reported for \texttt{QK-Norm} in \cite{DBLP:journals/corr/abs-2010-04245}.}
\label{fig: attention map qkvnorm}
\end{figure}

We see that the \texttt{Pre-Norm} distribution has a sharp peak at $1$, indicating a significant use of sparse-attention. By contrast, the \texttt{QKV-Norm} distribution is weighted towards $0$ and has no peak at $1$. To verify this behaviour, Figure~\ref{fig: attention map prenorm} shows an attention heatmap for a randomly chosen datapoint when using the \texttt{Baseline} \texttt{Pre-Norm} model, and Figure~\ref{fig: attention map qkvnorm} shows the same datapoint for \texttt{QKV-Norm}. We observe a significantly less sparse attention matrix for \texttt{QKV-Norm}. Note that \cite{DBLP:journals/corr/abs-2010-04245} also shows a similar visualisation.

%%
%%  Appendix:  Supplementary theorems
%%
\clearpage
\section{Supplementary theorems}
\label{appendix: supplementary theorems}

This appendix contains theorems that support the main results, providing additional context or being pre-requisite for the proofs in appendix~\ref{appendix: proofs}. We use the formulation of section~\ref{sec: formulation}, where $1 \leq \{t,t'\} \leq T$ are indices over tokens, $x\in\mathbb{R}^{N_x}$ is the message receiving embedding, $\{y_t\in\mathbb{R}^{N_y}\}$ are the message senders, $w_t = x^T W_{QK} y_t$, and $a_t = \texttt{softmax}_t w_t$ is the attention distribution

\vspace{0.5cm}

\begin{mdframed}[backgroundcolor=green!5,skipabove=-2pt,skipbelow=0]
\begin{theorem}
	Shifting attention scores $w_t$ by a constant offset does not affect the attention distribution. Therefore attention is fully determined by differences in scores.
 \label{theorem: att shift operator}
\end{theorem}

\textit{Proof.} ~Applying the shift $w_t \xrightarrow[\mathrm{offset~w}]{} w_t + \delta w ~\forall~t$ with fixed $\delta w$, we have
\begin{equation}
\begin{split}
a_t ~&=~ \frac{e^{w_t}}{\sum_{t'} e^{w_{t'}}} \\
&\xrightarrow[\mathrm{offset~w}]{} ~ \frac{e^{\delta w}e^{w_t}}{\sum_{t'} e^{\delta w}e^{w_{t'}}} ~=~ \frac{e^{\delta w}}{e^{\delta w}}\frac{e^{w_t}}{\sum_{t'} e^{w_{t'}}} ~=~ 1 \cdot a_t~=~ a_t
\end{split}
\end{equation}
Alternatively we may write
\begin{equation}
a_t ~=~ \frac{e^{w_t}}{\sum_{t'} e^{w_{t'}}} ~=~ \frac{e^{w_t}}{e^{w_{t'}}\sum_{t'} e^{w_{t'}-w_t}} ~=~ \frac{1}{\sum_{t'} e^{w_{t'}-w_t}}
\end{equation}
where $\left(w_{t'}+\delta t\right)-\left(w_t+\delta t\right) = w_{t'}-w_t$.
\end{mdframed}

\begin{mdframed}[backgroundcolor=green!5]
\begin{theorem}
	Multiplying attention scores by a positive factor changes the inverse-temperature of the attention distribution, modulating its sparsity (low temperature = less entropy = more sparse). Corollary: In the sparse limit, attention is fully determined by the order of $w_t$.
 \label{theorem: att scale operator}
\end{theorem}

\textit{Proof.} ~Applying the scaling $w_t \xrightarrow[\mathrm{scale~w}]{} \kappa w_t ~\forall~ t$ with fixed $\kappa > 0$, we have
\begin{equation}
\begin{split}
\frac{e^{\kappa w_t}}{\sum_{t'} e^{\kappa w_{t'}}} ~~&=~~ \frac{1}{\sum_{t'} e^{\kappa (w_{t'}-w_t)}}   \\
~~&\xrightarrow[\kappa\rightarrow 0]{} ~~~~ \frac{1}{\sum_{t'} e^0} ~=~ \frac{1}{T} ~\forall~ t ~~~~~~~~~~~~~~~~~~~~~\text{[fully isotropic distribution]} \\
~~&\xrightarrow[\kappa\rightarrow \infty]{} ~~ 
\begin{cases}
1 ~~ ~\mathrm{for}~ t = \mathrm{argmax}_{t'} w_{t'} \\
0 ~~ ~~~\forall~ t \neq \mathrm{argmax}_{t'} w_{t'} \\
\end{cases}
~~~\text{[fully sparse distribution]}
\end{split}
\end{equation}
where the \texttt{argmax} operator is fully determined by the order of $w_t$.
\end{mdframed}

\begin{mdframed}[backgroundcolor=green!5]
\begin{theorem}
    In the \texttt{No-Norm} case, the attention distribution $a_t$ is defined by the projection of $y_t$ onto a fixed vector $y_x$ for a given $x$. The length of $y_x$ is an inverse-temperature parameter.
\end{theorem}

\textit{Proof.} ~Write $w_t = x^TW_{QK}y_t = (W_{QK}^Tx)^T y_t \equiv y_x^T y_t$ where $y_x \triangleq W_{QK}^T x\in\mathbb{R}^{N_y}$, which is the dot-product between $y_t$ and a fixed vector $y_x$ on the row space of $W_{QK}$. Then, re-writing in terms of the vector lengths and the enclosing angle $\theta_{y_t} = y_x\wedge y_t$, we have $w = |y_x||y_t|\cos\theta_{y_t}$. The factor $|y_x|$ is identical for all $t$, making it an inverse-temperature.
\end{mdframed}

% \begin{mdframed}[backgroundcolor=green!5]
% \begin{theorem}
%     In the \texttt{No-Norm} case, for parallel heads to attend to a different ordering of tokens, the difference must be driven by information encoded in angles rather than norms. Corollary: parallel circuits cannot attend to different tokens based on information in norms.
% \end{theorem}

% \textit{Proof.} ~Consider the form $w_t = |y_x||y_t|\cos\theta_{y_t}$. The factor $|y_x|$ does not affect the order of $w_t$. The factor $|y_t|$ is identical for different heads. Any difference in order must therefore be encoded in $\cos\theta_{y_t}$.
% \end{mdframed}

\clearpage
\begin{mdframed}[backgroundcolor=green!5]
\begin{theorem}
In the \texttt{No-Norm} case, bias parameters in the construction of query and key vectors are nullified by the \texttt{softmax}, or only contribute terms that may be recovered if $x$ contains a constant direction.
\label{theorem: decomposition}
\end{theorem}

\textit{Proof.} ~Consider a modification to the construction of query and key vectors that uses the affine transformations $q = W_Qx+b_Q$ and $k_t = W_Ky_t+b_K$, with $W_{Q}\in\mathbb{R}^{N_{qkv}\times N_x}$, $W_{K}\in\mathbb{R}^{N_{qkv}\times N_y}$, $W_{QK}\triangleq W_Q^TW_K$, and $b_Q,b_K\in\mathbb{R}^{N_{qkv}}$. The dot-product attention scores are then:
\begin{equation}
\begin{split}
    w_t ~&=~ q^T k_t \\
       &=~ \left(W_Qx + b_Q\right)^T \left(W_Ky_t + b_K\right)  \\
       &=~ x^TW_{QK}y_t ~+~ ({W_Q}^Tb_K)^Tx ~+~ ({W_K}^Tb_Q)^Ty_t ~+~ b_Q^Tb_K \\
    w_t ~+~ const~ &=~ x^TW_{QK}y_t ~+~ ({W_Q}^Tb_K)^Tx ~+~ ({W_K}^Tb_Q)^Ty_t  \\
       &\triangleq~ x^TW_{QK}y_t ~+~ \rho_x^Tx ~+~ \rho_y^Ty_t  ~~~~~~~~~~~~~\rightarrow ~\rho_x^Tx =const ~\mathrm{given}~x\rightarrow \\
       &=~ x^TW_{QK}y_t ~+~ \rho_y^Ty_t  ~~~~~~~~~~~~~~~~~~~~~~~~~~~ \rightarrow ~W_{QK} \triangleq \Omega^T \Lambda \Sigma~\text{via SVD}~\rightarrow  \\
       &=~ x^T\Omega^T\Lambda\Sigma y_t ~+~ \rho_y^Ty_t ~~~~~~~~~~~~~~~~~~~~~~~~~~ \rightarrow ~x' \triangleq \Omega x, ~~y'_t \triangleq \Sigma y_t ~\rightarrow \\
       &=~ {x'}^T\Lambda y'_t ~+~ \rho_y^Ty_t
\end{split}
\end{equation}
After expanding the terms, we find an additive constant $b_Q^Tb_K$, and move this onto the LHS. Theorem~\ref{theorem: att shift operator} states that this has no impact on the output of the \texttt{softmax} operator. We identify $\rho_x\triangleq W_Q^Tb_k$ and $\rho_y\triangleq W_K^Tb_q$ as vectors on the \textbf{row-spaces of $W_Q$ and $W_K$ respectively}, defined as linear maps of the special directions $b_K$ and $b_Q$. Since $x$ is constant for each \texttt{softmax}, $\rho_x^Tx$ is constant, and we absorb it into the LHS. We perform the singular value decomposition $W_{QK} \triangleq \Omega^T \Lambda \Sigma$ where $\{\Omega\in\mathbb{R}^{N_x\times N_x},~\Sigma\in\mathbb{R}^{N_y\times N_y}\}$ are orthonormal matrices and $\Lambda \in \mathbb{R}^{N_x\times N_y}$ is a diagonal matrix of positive-semidefinite singular values with maximum rank $\min(N_x,N_y,N_{qkv})$. Orthonormal matrices apply a basis change to the embedding space using rotations and reflections. We write the transformed embeddings as $x' \triangleq \Omega x$ and $y_t' \triangleq \Sigma y_t$. The dot-product then has two terms:
\begin{enumerate}
    \item ${x'}^T \Lambda y'_t = \sum_i \Lambda_{ii} x'_i y'_{ti}$ sculpts the attention distribution according to \textit{pairwise relationships} between embeddings. We can say that $\{\Omega,\Sigma\}$ align the bases of $x$ and $y_t$, mapping them onto a common orthonormal coordinate system. $\Lambda_{ii}$ then assigns an importance weight to each coordinate $i$, determining the contribution of $x'_iy'_{ti}$.
    %\footnote{``Every direction'' is true when $M=N_x=N_y\equiv N$, otherwise the rank of $Q^TK$ is $min(M,N_x,N_y)$.}
    %\item $\rho_x^Tx$ means ``$(s)$ receives from all senders when $x \parallel \rho_x$'', where $\rho_x$ must be a vector on the column-space of $Q$.
    \item $\rho_y^Ty$ means ``token $t$ sends to all receivers when $y_t \parallel \rho_y$'', where $\rho_y$ must be a vector on the row-space of $W_K$. This may be recovered in the expansion of ${x'}^T\Lambda y'_t$ if there exists a direction $i$ for which $x'_i=const$.
\end{enumerate}
\end{mdframed}

%%
%%  Appendix:  Theorem proofs
%%
\clearpage
\section{Proofs of theorems in the main text}
\label{appendix: proofs}

This appendix provides proofs for the theorems presented in section~\ref{sec: theory: residual structure}-\ref{sec: theory: circuit stability}.

\vspace{0.5cm}

\begin{mdframed}[backgroundcolor=green!5]
\textbf{Theorem~\ref{theorem: structure: no-norm}.} \textit{\texttt{No-Norm}: If two heads with finite non-zero temperature attend to different semantic subspaces, the subspaces must be linearly independent $\mathbb{S}^{N_\alpha}_\alpha \equiv \mathbb{R}^{N_\alpha}$. Corollary: $W_{QK}$ is a low-rank matrix with (left and right) null-spaces that span all non-attended information.}

\textit{Proof.} ~
Let $\theta_A$ and $\theta_B$ be co-ordinates for the subspaces of $x$ attended to by heads A and B respectively, and $\phi$ be all other information. Let $\theta_A\perp\theta_B\perp\phi$ and $x \perp y_t$, where $\perp$ denotes independence. Without loss of generality, write
\begin{equation}
    x(\theta_A,\theta_B,\phi) ~=~ x_A(\theta_A) ~+~ x_B(\theta_B) ~+~ x_{other}(\theta_A,\theta_B,\phi)
\end{equation}
Then write
\begin{equation}
\begin{split}
    w_t^{(A)}(\theta_A) ~&=~ \left(W_{QK}^{(A)} y_t\right)^T x(\theta_A,\theta_B,\phi)  \\
    &=~ \left(W_{QK}^{(A)} y_t\right)^T x_A(\theta_A) ~+~ \left(W_{QK}^{(A)} y_t\right)^T x_B(\theta_B) ~+~ \left(W_{QK}^{(A)} y_t\right)^T x_{other}(\theta_A,\theta_B,\phi) \\
\end{split}
\end{equation}
which requires $\left(W_{QK}^{(A)} y_t\right)^T x_B(\theta_B)=0$ and $\left(W_{QK}^{(A)} y_t\right)^T x_{other}(\theta_A,\theta_B,\phi)=0$, since any cancellation between the two terms must be independent of $\theta_A,\phi$ and so can be absorbed entirely into the function $x_{B}(\theta_B)$. This means that $x_B(\theta_B)$ and $x_{other}(\theta_A,\theta_B,\phi)$ must both be orthogonal to $W_{QK}^{(A)} y_t$, meaning that they reside on the \textit{left null space} of $W_{QK}^{(A)}$, or are projected by ${W_{QK}^{(A)}}^T$onto a null space of $y_t$.

Head A can only attend to $\theta_A$ if $x_A(\theta_A)$ it is not on either of these null spaces, meaning that $x_A(\theta_A)$ is linearly independent of $x_{B}(\theta_B)$ and $x_{other}(\theta_A,\theta_B,\phi)$. Likewise for head B
\begin{equation}
\begin{split}
    w_t^{(B)}(\theta_B) ~&=~ \left(W_{QK}^{(B)} y_t\right)^T x(\theta_A,\theta_B,\phi)  \\
    &=~ \left(W_{QK}^{(B)} y_t\right)^T x_A(\theta_A) ~+~ \left(W_{QK}^{(B)} y_t\right)^T x_B(\theta_B) ~+~ \left(W_{QK}^{(B)} y_t\right)^T x_{other}(\theta_A,\theta_B,\phi) \\
\end{split}
\end{equation}
requires that $x_{B}(\theta_B)$ is linearly independent of both $x_A(\theta_A)$ and $x_{other}(\theta_A,\theta_B,\phi)$. Since $x_{other}$ resides on both null spaces, it is linearly independent of both $x_A(\theta_A)$ and $x_B(\theta_B)$, and may be seen as a third subspace that passes information through to subsequent layers.

We can also write $w_t = \left(W_{QK}^T x\right)^T y_t$, and so the same argument also holds for subspaces on $y_t$. In this case, non-attended subspaces are spanned by the \textit{right null space} of $W_{QK}$.
\end{mdframed}

\clearpage
\begin{mdframed}[backgroundcolor=green!5]
\textbf{Theorem~\ref{theorem: structure: pre-norm}.} \textit{\texttt{Pre-Norm}: Semantic subspaces must be represented as orthogonal spheres $\mathbb{S}^{N_\alpha}\equiv\mathcal{S}^{N_\alpha-1}$ defined using the $L_2$-norm. Corollary: if either orthogonality or constant-norm are violated, semantic subspaces interfere through a multiplicative factor on $w_t$.}

\textit{Proof.} ~Write 
\begin{equation}
x(\theta_A,\theta_B,\phi) ~=~ x_A(\theta_A) ~+~ x_B(\theta_B) ~+~ x_{AB}(\theta_A,\theta_B) ~+~ x_{other}(\theta_A,\theta_B,\phi)
\end{equation}
Then for head A we have
\begin{equation}
    w_t^{(A)}(\theta_A) ~=~ \frac{1}{\left|y_t\right|\left|x(\theta_A,\theta_B,\phi) \right|} {w^*_t}^{(A)}(\theta_A)
\end{equation}
where $w^*_t$ are the attention scores from the \texttt{No-Norm} case, which requires $x_A(\theta_A)$ and $x_B(\theta_B)$ to be linearly independent. Now we additionally require $\left|x(\theta_A,\theta_B,\phi) \right| \perp \theta_B,\phi$, with
\begin{equation}
|x| ~=~ \sqrt{|x_A|^2 ~+~ |x_B ~+~ x_{AB} ~+~ x_{other}|^2 ~+~ 2 x_A^T \left(x_B ~+~ x_{AB} ~+~ x_{other}\right)}
\end{equation}
where we suppress parameter dependence for readability. Since $\sqrt{\cdot}$ is a monotonic function, this can only be satisfied if
\begin{equation}
|x_A|^2 ~+~ |x_B ~+~ x_{AB} ~+~ x_{other}|^2 ~+~ 2 x_A^T \left(x_B ~+~ x_{AB} ~+~ x_{other}\right) ~\perp~ \theta_B,\phi
\end{equation}
Repeating this process for head B gives
\begin{equation}
|x_B|^2 ~+~ |x_A ~+~ x_{AB} ~+~ x_{other}|^2 ~+~ 2 x_B^T \left(x_A ~+~ x_{AB} ~+~ x_{other}\right) ~\perp~ \theta_A,\phi
\end{equation}
Combining and collecting dependencies, we then have
\begin{align}
    |x_A|^2 ~=~ const ~~~&\forall~~~ \theta_A \\
    |x_B|^2 ~=~ const ~~~&\forall~~~ \theta_B \\
    %|x_{AB}|^2 ~+~ 2x_A^Tx_B ~+~ 2x_{AB}^T\left(x_A + x_B\right) ~=~ const ~~~&\forall~~~ \theta_A,\theta_B \\
    \left( x_{AB} ~+~ 2x_A ~+~ 2x_B \right)^T x_{AB} ~+~ 2x_A^Tx_B ~=~ const ~~~&\forall~~~ \theta_A,\theta_B \\
    \left(x_{other} + 2x_A + 2x_B + 2x_{AB}\right)^T x_{other} ~=~ const ~~~&\forall~~~ \theta_A,\theta_B,\phi
\end{align}
We can go one step further, noticing that each individual term carries a different functional dependence, and so must independently be constant\footnote{N.B. If $|x_{AB}|^2 \propto x_A^Tx_B$ then $|x_{AB}|^2=const$ reduces to $x_A^Tx_B=const$, which is already required.}. We then have $\forall~~\mu,\nu\in\{A,B,AB,other\}$
\begin{equation}
    |x_\mu|=const   ~~~~~~\mathrm{and}~~~~~~  x_\mu^Tx_\nu=const 
\end{equation}
The requirements $|x_A(\theta_A)|=const ~\forall~\theta_A$ and $|x_B(\theta_B)|=const ~\forall~\theta_B$ mean that the semantic subspaces have a spherical structure defined by the $L_2$-norm $|\cdot|$.

Now consider the requirement $x_A(\theta_A)^Tx_B(\theta_B)=const$. Say that $\theta_A$ and $\theta_B$ have $N_A$ and $N_B$ degrees of freedom, meaning that $x_A$ and $x_B$ have $N_A-1$ and $N_B-1$ respectively, since they each lose one by confinement to the sphere. Say that the constant is nonzero such that $x_A^Tx_B \neq 0$. This means that there must be some direction $i$ for which $x_{Ai}x_{Bi} \neq0$. If we know all $N_A-1$ coordinates of $x_A$, and all $N_B - 2$ coordinates of $x_B$ except for direction $i$, then we also know the value of $x_{Bi}$, because it is fixed by the constant. However, this would mean that $x_A$ and $x_B$ are not independent, violating the condition $\theta_A \perp \theta_B$. The only way to satisfy independence is if $x_{Ai}x_{Bi}=0~\forall~i$, ensuring that degrees of freedom on $x_A$ and $x_B$ never become entangled. Therefore, to satisfy semantic independence, we must have $x_A(\theta_A)^Tx_B(\theta_B)=0 ~\forall~\theta_A,\theta_B$. This means that the subspaces are not just linearly independent, but orthogonal.

We have shown the proof for semantic subspaces of $x$. As for Theorem~\ref{theorem: structure: no-norm}, the same structure must be true for $y_t$ by symmetry.
\end{mdframed}

\clearpage
\begin{mdframed}[backgroundcolor=green!5]
\textbf{Theorem~\ref{theorem: structure: qkv-norm}.} \textit{\texttt{QKV-Norm}: Semantic subspaces must be linearly separable, reproducing the \texttt{No-Norm} case.}

\textit{Proof.} ~We have
\begin{equation}
    w_t^{(A)}(\theta_A) ~=~ \frac{1}{\left|k_t^{(A)}\right|\left|q^{(A)}\right|} {w^*_t}^{(A)}(\theta_A)
\end{equation}
where $w^*_t$ are the attention scores from the \texttt{No-Norm} case, which requires $x_A(\theta_A)$ and $x_B(\theta_B)$ to be linearly independent. Use
\begin{equation}
x(\theta_A,\theta_B,\phi) ~=~ x_A(\theta_A) ~+~ x_B(\theta_B) ~+~ x_{other}(\theta_A,\theta_B,\phi)
\end{equation}
and 
\begin{equation}
\begin{split}
    q^{(A)}(\theta_A) ~&=~ W_Q^{(A)} x(\theta_A,\theta_B,\phi)  \\
    &=~ W_Q^{(A)} x_A(\theta_A) ~+~ W_Q^{(A)} x_B(\theta_B) ~+~ W_Q^{(A)} x_{other}(\theta_A,\theta_B,\phi) \\
\end{split}
\end{equation}
Since we already have the condition of linearly independent $x_A,x_B$, there must exist a linear projection operator $P_A$ such that $P_A x_A = x_A$. Defining $W_Q^{(A)}=P_A$, we then have
\begin{equation}
    q^{(A)}(\theta_A) ~=~ W_Q^{(A)} x_A(\theta_A) 
\end{equation}
This demonstrates that it is possible to separate linearly independent semantic subspaces on $x$. By symmetry of $w_t^{(A)}(\theta_A)$, the same must be true for $y_t$.
\end{mdframed}

\begin{mdframed}[backgroundcolor=green!5]
\textbf{Theorem~\ref{theorem: stability: general}.} Consider independent infinitesimal perturbations on queries $\epsilon^q \in \mathbb{R}^{N_{qkv}}$, keys $\epsilon^k_t \in \mathbb{R}^{N_{qkv}}$, and messages $\epsilon^m_t \in \mathbb{R}^{N_{qkv}}$. These propagate onto $\Delta x = \sum_{t}a_tm_t$ as
    \begin{align}
        \epsilon^{\Delta x(q)} ~~&\xrightarrow[\epsilon^q\rightarrow0]{\mathrm{~~~~perturb~q~~~~}}~~ \mathop{\mathbb{E}}_{a_t} \Big[ m_t {\tilde k}_t^T \Big] \epsilon^q ~~~~~~~~~~~~~~~~~  {\tilde k}_t ~\triangleq~ k_t ~- \mathop{\mathbb{E}}_{a_t} \Big[ k_t \Big] 
        \label{eq: stability: general q}\\
        \epsilon^{\Delta x(k)} ~~&\xrightarrow[\epsilon^k_t\rightarrow0]{\mathrm{~~~~perturb~k~~~~}}~~ \mathop{\mathbb{E}}_{a_t} \Big[ {\tilde m}_t {\epsilon^k_t}^T \Big] q ~~~~~~~~~~~~~~~~~  {\tilde m}_t ~\triangleq~ m_t ~- \mathop{\mathbb{E}}_{a_t} \Big[ m_t \Big]
        \label{eq: stability: general k} \\
        \epsilon^{\Delta x(m)} ~~&\xrightarrow[\epsilon^m_t\rightarrow0]{\mathrm{~~~~perturb~m~~~~}}~~ \mathop{\mathbb{E}}_{a_t} \Big[ \epsilon^m_t \Big]
        \label{eq: stability: general m}
    \end{align}
    where ${\tilde z}_t$ is the value of $z_t$ measured from the attention-weighted centroid $\mathbb{E}_{a_t}[z_t] = \sum_t a_t z_t$.

\textit{Proof.} ~Consider $q\rightarrow q+\epsilon^q$ where $\epsilon^q$ are infinitesimal perturbations on $q$. Then $\Delta x \rightarrow \Delta x + \epsilon^{\Delta x(q)}$ where by Taylor expansion we find
\begin{equation}
    \epsilon^{\Delta x(q)} ~=~ \frac{\partial \Delta x}{\partial q}\epsilon^q ~+~ \mathcal{O}\left({\epsilon^q}^2\right)
\end{equation}
where the leading term is a matrix $\frac{\partial \Delta x}{\partial q}$ acting on a vector $\epsilon^q$. Differentiating gives
\begin{equation}
    \frac{\partial\Delta x}{\partial q} ~=~ \sum_{ij} m_i \frac{\partial a_i}{\partial w_j} \frac{\partial w_j}{\partial q}
\end{equation}
with $a_i = \texttt{softmax}_i(w_i)$ and $w_i=k_i^Tq$, and we are using $i,j,k$ etc to index over tokens instead of $t,t',t''$ etc, because this is more readable when we have many summations. Then

\textcolor{Maroon}{\textit{[continued in next box...]}}
\end{mdframed}

\clearpage
\begin{mdframed}[backgroundcolor=green!5]
\textcolor{Maroon}{\textit{[...continuing from previous box]}}

\begin{equation}
\begin{split}
    \frac{\partial a_i}{\partial w_j} ~&=~ \frac{\partial}{\partial w_j} ~ \frac{e^{w_i}}{\sum_k e^{w_k}} \\
    &=~ \frac{\delta_{ij}e^{w_i}}{\sum_k e^{w_k}} ~+~ e^{w_i}\left(-\frac{e^{w_j}}{\left(\sum_ke^{w_k}\right)^2}\right) \\
    &=~ \frac{e^{w_i}}{\sum_k e^{w_k}}\left( 1 ~-~ \frac{e^{w_j}}{\sum_le^{w_l}}\right) \\
    &=~ a_i\left(\delta_{ij} ~-~ a_j\right) \\
\end{split}
\end{equation}
and $\frac{\partial w_i}{\partial q} = k_i^T$, where we retain the transpose to indicate that this is an element of the dual vector space (i.e. covector). Inserting these results into our expression for $\epsilon^{\Delta x(q)}$ gives
\begin{equation}
\begin{split}
    \epsilon^{\Delta x(q)} ~&=~ \sum_{ij} m_i a_i\left(\delta_{ij} ~-~ a_j\right) k_j^T \epsilon^q \\
    &=~ \sum_{i} m_i a_i \left(k_i ~-~ \sum_j a_jk_j \right)^T \epsilon^q \\
    &=~ \sum_{i} m_i a_i {\tilde k}_i^T \epsilon^q \\
    &=~ \mathop{\mathbb{E}}_{a_i} \Big[m_i {\tilde k}_i^T \Big] \epsilon^q \\
\end{split}
\end{equation}
This is the result for Eq.~\ref{eq: stability: general q}. Repeating the process for perturbations on $k_i$, we have
\begin{equation}
    \epsilon^{\Delta x(k)} ~=~ \sum_i\frac{\partial \Delta x}{\partial k_i}\epsilon^k_i ~+~ \mathcal{O}\left({\epsilon^k}^2\right)
\end{equation}
and
\begin{equation}
\begin{split}
    \frac{\partial\Delta x}{\partial k_i} ~&=~ \sum_{jk} m_j \frac{\partial a_j}{\partial w_k} \frac{\partial w_k}{\partial k_i} \\
    &=~ \sum_{jk} m_j a_j \left(\delta_{jk} ~-~ a_k\right) \delta_{ki} q^T \\
    &=~ \sum_{j} m_j a_j \left(\delta_{ji} ~-~ a_i\right) q^T \\
    &=~ a_i {\tilde m}_i q^T
\end{split}
\end{equation}
Therefore
\begin{equation}
    \epsilon^{\Delta x(k)} ~=~ \sum_i a_i {\tilde m}_i q^T \epsilon^k_i ~=~ \mathop{\mathbb{E}}_{a_i} \Big[{\tilde m}_i {\epsilon^k_i}^T \Big] q
\end{equation}
which is the result for Eq.~\ref{eq: stability: general k}. Finally,
\begin{equation}
\begin{split}
    \epsilon^{\Delta x(m)} ~&=~ \sum_i \frac{\partial \Delta x}{\partial m_i}\epsilon^m_i \\
    &=~ \sum_{i} a_i \epsilon^m_i \\
    &=~ \mathop{\mathbb{E}}_{a_i} \Big[ \epsilon^m_i \Big]
\end{split}
\end{equation}
using $\frac{\partial\Delta x}{\partial m_i} = \frac{\partial}{\partial m_i}\sum_j a_j m_j = \sum_j a_j \delta_{ij} = a_i$. This is the result for Eq.~\ref{eq: stability: general m}.
\end{mdframed}

\begin{mdframed}[backgroundcolor=green!5]
\textbf{Theorem~\ref{theorem: stability: sparse}.} For sparse attention:
    \begin{equation}
        \epsilon^{\Delta x(q)} \xrightarrow[\epsilon^q\rightarrow0]{\mathrm{~~perturb~q~~}} 0   ~~~~~~~~~~
        \epsilon^{\Delta x(k)} \xrightarrow[\epsilon^k_t\rightarrow0]{\mathrm{~~perturb~k~~}} 0   ~~~~~~~~~~
        \epsilon^{\Delta x(m)} \xrightarrow[\epsilon^m_t\rightarrow0]{\mathrm{~~perturb~m~~}} \epsilon^m_{t^*}
    \end{equation}
    i.e. the message is stable with respect to small interference in the queries and keys. Interference in the selected value is linearly transferred onto the message.

\textit{Proof.} ~For sparse attention we have $a_t = \delta_{tt^*}$ for some $t^*$. For perturbations of $q$, the RHS of Eq.~\ref{eq: stability: general q} becomes
\begin{equation}
\begin{split}
    \mathop{\mathbb{E}}_{a_t} \Big[ m_t {\tilde k}_t^T \Big] \epsilon^q ~&=~ \sum_{t} a_t m_t {\tilde k}_t^T \epsilon^q \\
    &=~ \sum_{t} \delta_{tt^*} m_t {\tilde k}_t^T \epsilon^q \\
    &=~ m_{t^*} {\tilde k}_{t^*}^T \epsilon^q \\
    &=~ 0 \\
\end{split}
\end{equation}
where the final step is because ${\tilde k}_{t^*} = k_{t^*} - \mathbb{E}_{a_t}[k_t] = k_{t^*} - \sum_t \delta_{tt^*} k_t = k_{t^*}-k_{t^*} = 0$. For perturbations of $k_t$, the RHS of Eq.~\ref{eq: stability: general k} evaluates to $0$ because 
\begin{equation}
\begin{split}
    \mathop{\mathbb{E}}_{a_t} \Big[ {\tilde m}_t {\epsilon^k_t}^T \Big] q ~&=~ \sum_t a_t {\tilde m}_t q^T \epsilon^k_t \\
    &=~ \sum_t \delta_{tt^*} {\tilde m}_t q^T \epsilon^k_t \\
    &=~ {\tilde m}_{t^*} q^T \epsilon^k_{t^*} \\
    &=~ 0 \\
\end{split}
\end{equation}
where the final step is because ${\tilde m}_{t^*} = m_{t^*} - \sum_t \delta_{tt^*} m_t = m_{t^*}-m_{t^*} = 0$. For perturbations of $m_t$, the RHS of Eq.~\ref{eq: stability: general m} evaluates to
\begin{equation}
    \mathop{\mathbb{E}}_{a_t} \Big[ \epsilon^m_t \Big] ~=~ \sum_{t} a_t \epsilon^m_t ~=~ \sum_{t} \delta_{tt^*} \epsilon^m_t ~=~ \epsilon^m_{t^*}
\end{equation}
\end{mdframed}

\begin{mdframed}[backgroundcolor=green!5]
\textbf{Theorem~\ref{theorem: stability: isotropic}.}
For isotropic attention:
\begin{equation}
    \epsilon^{\Delta x(q)} \xrightarrow[\epsilon^q\rightarrow0]{\mathrm{perturb~q}} \langle m_t {\tilde k}_t^T \rangle_t \epsilon^q ~~~~~~~~
    %\epsilon^{\Delta x(q)} \xrightarrow[\epsilon^q\rightarrow0]{\mathrm{~perturb~q~}} 0   ~~~~~~~~~
    \epsilon^{\Delta x(k)} \xrightarrow[\epsilon^k_t\rightarrow0]{\mathrm{perturb~k}} \langle {\tilde m}_t {\epsilon^k_t}^T \rangle_t ~q   ~~~~~~~~
    \epsilon^{\Delta x(m)} \xrightarrow[\epsilon^m_t\rightarrow0]{\mathrm{perturb~m}} \langle \epsilon^m_t \rangle_t
\end{equation}
% \begin{align}
%     \epsilon^{\Delta x(q)} ~&\xrightarrow[\epsilon^q\rightarrow0]{\mathrm{~~perturb~q~~}}~ 0 \\
%     \epsilon^{\Delta x(k)} ~&\xrightarrow[\epsilon^k_t\rightarrow0]{\mathrm{~~perturb~k~~}}~ \langle {\tilde v}_t {\epsilon^k_t}^T \rangle_t ~q  \\
%     \epsilon^{\Delta x(v)} ~&\xrightarrow[\epsilon^v_t\rightarrow0]{\mathrm{~~perturb~v~~}}~ \langle \epsilon^v_t \rangle_t
% \end{alig
N.B. isotropy requires $k_t=const$ or $q=0$. Lemma 1: the update is stable to noisy $q$ when $k_t=const$, or when $m_t \perp k_t$ (c.f. keys and messages from independent subspaces). Lemma 2: the update is stable to noisy $k_t$ when $q=0$, or when $m_t \perp \epsilon_t^k$. Lemma 3: the update is stable to noisy $m_t$ when $\langle \epsilon^m_t \rangle_t=0$. Other cases propagate linearly.

\textit{Proof.} ~For isotropic attention we have $a_t = \frac{1}{T}$. For perturbations of $q$, the RHS of Eq.~\ref{eq: stability: general q} is
\begin{equation}
\begin{split}
    \mathop{\mathbb{E}}_{a_t} \Big[ m_t {\tilde k}_t^T \Big] \epsilon^q ~&=~ \sum_{t} a_t m_t {\tilde k}_t^T \epsilon^q \\
    &=~ \frac{1}{T} \sum_{t=1}^T m_t {\tilde k}_t^T \epsilon^q \\
    &=~ \langle m_t {\tilde k}_t^T \rangle_t \epsilon^q \\
\end{split}
\end{equation}
For lemma 1, we note that $k_t=const$ implies ${\tilde k}_t=0$, and if $m_t \perp k_t$ then $\langle m_t {\tilde k}_t^T \rangle_t = \langle m_t k_t \rangle_t - \langle m_t \rangle_t \langle k_t\rangle_t = Cov(m_t,k_t) = 0$.

\textcolor{Maroon}{\textit{[continued in next box...]}}
\end{mdframed}

\clearpage
\begin{mdframed}[backgroundcolor=green!5]
\textcolor{Maroon}{\textit{[...continuing from previous box]}}

For perturbations of $k_t$, the RHS of Eq.~\ref{eq: stability: general k} is
\begin{equation}
\begin{split}
    \mathop{\mathbb{E}}_{a_t} \Big[ {\tilde m}_t {\epsilon^k_t}^T \Big] q ~&=~ \frac{1}{T} \sum_{t=1}^T {\tilde m}_t {\epsilon^k_t}^T q \\
    &=~ \langle {\tilde m}_t {\epsilon^k_t}^T \rangle_t q \\
\end{split}
\end{equation}
For lemma 2, this expression evaluates to $0$ if $q=0$, and if $m_t \perp \epsilon_t^k$ then $\langle {\tilde m}_t {\epsilon^k_t}^T \rangle_t = \langle m_t {\epsilon^k_t}^T \rangle_t - \langle m_t \rangle_t \langle {\epsilon^k_t}^T\rangle_t = Cov(m_t,{\epsilon^k_t}^T) = 0$.

For perturbations of $m_t$, the RHS of Eq.~\ref{eq: stability: general m} evaluates to
\begin{equation}
    \mathop{\mathbb{E}}_{a_t} \Big[ \epsilon^m_t \Big] ~=~ \frac{1}{T}\sum_{t=1}^T \epsilon^m_t ~=~ \langle \epsilon^m_t \rangle_t
\end{equation}
\end{mdframed}

\begin{mdframed}[backgroundcolor=green!5]
\textbf{Theorem~\ref{theorem: multiplicative stability: sparse}. } ~Sensitivity of sparse attention to multiplicative perturbations $\epsilon^q = \kappa^q q$ and $\epsilon^k = \kappa^k_t k_t$ with $\kappa^q,\kappa^k_t\ll1$. Circuit collapse occurs when $\exists~ t \neq t^*$ for which:
\begin{equation}
    \frac{w_{t^*}}{w_t} ~\begin{cases} ~<~ \lambda_w & \mathrm{if}~ w_t \left(1 + \kappa^q + \kappa^k_{t^*}\right) > 0 \\
    ~>~ \lambda_w & \mathrm{otherwise} \\ \end{cases}
    ~~~~~~~~~~~~~ \lambda_w ~\triangleq~ \frac{1 + \kappa^q + \kappa^k_t}{1 + \kappa^q + \kappa^k_{t^*}}
\end{equation}
where temperature cancels in the fraction. \textbf{Attention is fully stable above the critical transition point $\lambda_w$} (c.f. $w_t \left(1 + \kappa^q + \kappa^k_{t^*}\right) > 0$). We see that query perturbations alone are insufficient, as they result in $\lambda_w=1$. Lemma: consider the special case when all keys have similar length $k_t \approx const$, the attended token has $\theta_{t^*}\approx0$, the keys are far-from-orthogonal s.t. $\theta_t \ll 1$, and $\kappa^q\approx0$. Using $w_t \triangleq |q| |k_t| \cos\theta_t$, circuit collapse occurs when $\exists~ t \neq t^*$ for which:
\begin{equation}
        \frac{1}{2}\theta_t^2 ~\lesssim~ \kappa^k_t - \kappa^k_{t^*}   ~~~~~~~~~~~ \mathrm{if}~ w_t \left(1  + \kappa^k_{t^*}\right) > 0 ~\text{, otherwise reverse}
\label{eq: app: sparse circuit collapse result}
\end{equation}
i.e. stability requires either well-separated keys s.t. $\theta_t \gg 0$, or small perturbations $\kappa_t-\kappa^*_t \ll 1$.

\textit{Proof.} ~ Apply $q\rightarrow q+\epsilon^q$ and $k_t\rightarrow k_t+\epsilon_t^k$ to $w_t = q^Tk_t$, then we have $w_t \rightarrow w_t + \epsilon_w$ such that $\epsilon^w_t = q^T\epsilon_t^k + {\epsilon^q}^Tk_t + {\epsilon^q}^T\epsilon_t^k$. For multiplicative perturbations we have  $\epsilon^q = \kappa^q q$ and $\epsilon^k = \kappa^k_t k_t$, and so $\epsilon^w_t = \kappa^k_t q^Tk_t + \kappa^q q^Tk_t + \kappa^k_t\kappa^qq^Tk_t$. Each term recovers a factor of $w_t=q^Tk_t$, which we factor out to give $\epsilon^w_t = \left(\kappa^q  + \kappa^k_t + \kappa^k_t\kappa^q\right)w_t$. The final term is subleading in the limit of small perturbations, and so
\begin{equation}
    \epsilon^w_t ~\xrightarrow[~\kappa^q,\kappa^k_t\rightarrow0~]{}~ \left(\kappa^q  ~+~ \kappa^k_t\right)w_t ~+~ \mathcal{O}\left(\kappa^q\kappa^k_t\right)
\end{equation}
Circuit collapse occurs when $w_{t^*} - w_t < \epsilon^w_t - \epsilon^w_{t^*}$ for some $t$. Substituting our limit for $\epsilon^w_t$ gives
\begin{equation}
    w_{t^*} - w_t ~<~ \left(\kappa^q  ~+~ \kappa^k_t\right)w_t - \left(\kappa^q  ~+~ \kappa^k_{t^*}\right)w_{t^*}
\end{equation}
and collecting terms gives
\begin{equation}
    \left(1 ~+~ \kappa^q ~+~ \kappa^k_{t^*}\right) w_{t^*} ~<~ \left(1 ~+~ \kappa^q ~+~ \kappa^k_t\right)w_t
\end{equation}
We then divide each side by $w_t (1 + \kappa^q + \kappa^k_{t^*})$, taking care to reverse the sign of the inequality when this factor is negative, to give

\textcolor{Maroon}{\textit{[continued in next box...]}}
\end{mdframed}

\clearpage
\begin{mdframed}[backgroundcolor=green!5]
\textcolor{Maroon}{\textit{[...continuing from previous box]}}

\begin{equation}
    \frac{w_{t^*}}{w_t} ~\begin{cases} ~<~ \lambda_w & \mathrm{if}~ w_t \left(1 + \kappa^q + \kappa^k_{t^*}\right) > 0 \\
    ~>~ \lambda_w & \mathrm{otherwise} \\ \end{cases}
    ~~~~~~~~~~~~~ \lambda_w ~\triangleq~ \frac{1 + \kappa^q + \kappa^k_t}{1 + \kappa^q + \kappa^k_{t^*}}
\end{equation}
which is the first expression in the theorem.  We note that any temperature parameter cancels in the fraction, which means that the attention head cannot become more stable by reducing its temperature to become more sparse. $\lambda_w$ has the limits
\begin{equation}
    \lambda_w ~\xrightarrow[\kappa^q\rightarrow0]{~~\mathrm{keys~only}~~}~ \frac{1+\kappa^k_t}{1+\kappa^k_{t^*}}
    ~~~~~~~~~~~~~~~~~
    \lambda_w ~\xrightarrow[\kappa^k_t,\kappa^k_{t^*}\rightarrow0]{~~\mathrm{query~only}~~}~ \frac{1 + \kappa_q}{1 + \kappa_q} = 1
\end{equation}
meaning that query perturbations alone are insufficient, contributing only when they co-occur with perturbations on the keys. Write $w_t = |q| |k_t| \cos\theta_t$ with $\theta_t = q \wedge k_t$, and the approximation of identical key norms $k_{t^*}=k_t\equiv k$ turns this into $w_t = |q| |k| \cos\theta_t$. Then
\begin{equation}
    \frac{w_{t^*}}{w_t} ~=~ \frac{|q| |k| \cos\theta_{t^*}}{|q| |k| \cos\theta_t} ~=~ \frac{\cos\theta_{t^*}}{\cos\theta_t}
\end{equation}
Then $\theta_{t^*}=0$ means that $\cos\theta_{t^*} = \cos0=1$, and so $\frac{\cos\theta_{t^*}}{\cos\theta_t} = \frac{1}{\cos\theta_t} = \sec \theta_t$. We perform a Taylor expansion in $\theta_t$ to obtain
\begin{equation}
    \frac{w_{t^*}}{w_t} ~\approx~ \sec\theta_t ~\approx~ 1 ~+~ \frac{1}{2}\theta_t^2 ~+~\mathcal{O}\left(\theta_t^4\right)
\end{equation}
which is valid when $\theta_t \ll 1$. This is true for any $t\neq t^*$ for which $k_t$ is far from orthogonal with $k_{t^*}$. Substituting this into our circuit collapse condition, we have
\begin{equation}
    1 ~+~ \frac{1}{2}\theta_t^2 ~<~ \frac{1 + \kappa^k_t}{1 + \kappa^k_{t^*}} ~~~~~~~~~~~~~~ \mathrm{if}~ w_t \left(1 + \kappa^k_{t^*}\right) > 0 
\end{equation}
where we consider the case of $\kappa_q\approx0$ for readability. Re-arranging gives
\begin{equation}
    \frac{1}{2}\theta_t^2 ~\lesssim~  \kappa^k_t - \kappa^k_{t^*} ~~~~~~~~~~~~~~~~\text{Circuit~collapse~when~}k_t~\text{similar}
\label{eq: app: sparse circuit collapse result duplicate}
\end{equation}
if $w_t(1 + \kappa^k_{t^*}) > 0$, and we reverse the inequality otherwise. We have approximated the denominator on the RHS as $1 + \kappa^k_{t^*} \approx 1$ for $\kappa^k_{t^*}\rightarrow0$.

When $\theta_t \ll 1$, the LHS of Eq.~\ref{eq: app: sparse circuit collapse result duplicate} is small. This means that the attention head can tolerate only very small perturbations $\{\kappa^k_t,\kappa^k_{t^*}\}$. Therefore semantic subspaces must either have a highly orthogonal substructure s.t. $\theta_t \gtrsim 1 ~\forall~t\neq t^*$, or be orthogonal s.t. $\kappa_t\ll1 ~\forall~ t$.

\end{mdframed}

\clearpage
\begin{mdframed}[backgroundcolor=green!5]
\textbf{Theorem. ~\ref{theorem: multiplicative stability: isotropic}}. ~Sensitivity of isotropic attention to multiplicative perturbations. Say $\epsilon^k = \kappa^k_t k_t$ with $\kappa^k_t\ll1$ where $\{\kappa_t\}$ have comparable amplitudes. Then
\begin{equation}
\epsilon^{\Delta x(k)} %~\xrightarrow[\epsilon^k_t\rightarrow0]{\mathrm{~~perturb~k~~}}%~ w~ \langle {\tilde v}_t \kappa^k_t \rangle_t 
~\approx~
    \begin{cases}
    0 ~&~ \text{if~$\kappa_t$~independent~of~${\tilde m}_t$,~by~symmetry} \\
    0 ~&~ \text{if~$\kappa_t\equiv\kappa$~for~constant~$\kappa$} \\
    0 ~&~ \text{if~$q=0$} \\
    w \langle {\tilde m}_t \kappa^k_t \rangle_t  ~&~ \text{otherwise}
    \end{cases}
\end{equation}
%Stability is driven by the central limit theorem, replacing isotropic perturbations with their mean-field limit. Variance creates instability that increases with the number of random variables.

\textit{Proof.} ~We begin with the following result from Theorem~\ref{theorem: stability: isotropic}:
\begin{equation}
\epsilon^{\Delta x(k)} ~\xrightarrow[\epsilon^k_t\rightarrow0]{\mathrm{~~perturb~k~~}}~ \langle {\tilde m}_t {\epsilon^k_t}^T \rangle_t ~q
\end{equation}
Substituting $\epsilon^k = \kappa^k_t k_t$ and taking $q$ inside the brackets gives
\begin{equation}
\langle{{\tilde m}_t  \epsilon^k_t}^T \rangle_t ~q ~=~ 
\langle {\tilde m}_t \kappa_t {k_t}^T \rangle_t q ~=~
 ~ \langle {\tilde m}_t \kappa_t w_t \rangle_t
\end{equation}
We then notice that isotropic attention requires that $w_t$ is a constant, which we call $w$. Then
\begin{equation}
\epsilon^{\Delta x(k)} ~\approx~ w \langle {\tilde m}_t \kappa_t \rangle_t
\end{equation}
is our general result. We then note three special cases, each resulting in $\epsilon^{\Delta x(k)}=0$:
\begin{enumerate}
    \item If $\kappa_t \perp {\tilde m}_t$ then $\langle {\tilde m}_t \kappa_t \rangle_t = \langle m_t \kappa_t \rangle_t - \langle m_t \rangle_t \langle \kappa_t \rangle_t = Cov(m_t,\kappa_t) = 0$. This is case when interference $\kappa_t^k$ on the keys is not dominated by the same semantic subspace as the message $m_t$.
    \item If all keys are perturbed by the same factor $\kappa_t\equiv\kappa$, then $\langle {\tilde m}_t \kappa_t \rangle_t = \kappa \langle {\tilde m}_t \rangle_t =0$ because $\langle {\tilde m}_t \rangle_t=0$.
    \item Isotropic attention can be achieved by either $q=0$ or $k_t=const$. If the case is $q=0$ then this implies $w=0$ also.
\end{enumerate}
\end{mdframed}

\end{document}